\pdfoutput=1

\documentclass[11pt]{article}

\usepackage[]{acl}

\usepackage{times}
\usepackage{latexsym}
\usepackage{graphicx}
\usepackage{svg}
\usepackage{amsmath}
\usepackage{tcolorbox}\tcbuselibrary{skins}
\usepackage{subcaption}
\usepackage{multirow}
\usepackage{array}
\usepackage{makecell}
\usepackage{booktabs}
\usepackage{float}
\usepackage{soul}

\tcbset{daprompt/.style={
    enhanced,
    size=fbox,
    boxrule=2pt,
    arc=2mm,
    auto outer arc,
    left=1pt,
    right=1pt,
    top=1pt,
    bottom=1pt,
    fontupper=\fon{cmtt}, 
    colback=CB_pear!10,
    colframe=CB_pear!25,
    coltitle=CB_pear!20!black, 
}}

\tcbset{exampleprompt/.style={
    enhanced,
    size=fbox,
    boxrule=3pt,
    arc=2mm,
    auto outer arc,
    left=1pt,
    right=1pt,
    top=1pt,
    bottom=1pt,
    fontupper=\fon{cmtt}, 
    colback=CB_lightCyan!15,
    colframe=CB_lightCyan!30,
    coltitle=CB_lightCyan!25!black, 
}}

\tcbset{suggestionprompt/.style={
    enhanced,
    size=fbox,
    boxrule=3pt,
    arc=2mm,
    auto outer arc,
    left=1pt,
    right=1pt,
    top=1pt,
    bottom=1pt,
    colback=CB_pink!15,
    colframe=CB_pink!30,
    coltitle=CB_pink!25!black, 
    fontupper=\fon{cmtt}, 
}}

\tcbset{modelresponse/.style={
    enhanced,
    size=fbox,
    boxrule=3pt,
    arc=2mm,
    auto outer arc,
    left=1pt,
    right=1pt,
    top=1pt,
    bottom=1pt,
    fontupper=\fon{cmtt}, 
    colback=CB_gray!10,
    colframe=CB_gray!25,
    coltitle=CB_gray!25!black, 
}}

\tcbset{humanresponse/.style={
    enhanced,
    size=fbox,
    boxrule=3pt,
    arc=2mm,
    auto outer arc,
    left=1pt,
    right=1pt,
    top=1pt,
    bottom=1pt,
    fontupper=\fon{cmtt}, 
    colback=CB_orange!10,
    colframe=CB_orange!25,
    coltitle=CB_orange!20!black, 
}}

\def\numCLASSSegments{$18$}
\def\numMQISegments{$18$}
\def\numSuggestionSegments{$18$}
\def\totalCLASSitems{$216$}
\def\totalMQIitems{$288$}
\def\totalSuggestions{$36$}

\def\classPositiveClimate{CLPC}
\def\classBehavioralManagement{CLBM}
\def\classInstructionalDialogue{CLINSTD}

\def\mqiExplanations{EXPL}
\def\mqiRemediation{REMED}
\def\mqiErrors{LANGIMP}
\def\mqiStudentMath{SMQR} 

\def\directAnswer{\textsf{DA}}
\def\reasoningAnswer{\textsf{RA}}
\def\directAnswerDescription{\textsf{DA}$^+$}

\newcommand{\citepinteralia}[1]{\citep[][{\textit{inter alia}}]{#1}}

\definecolor{CB_pear}{HTML}{BBCC33}
\definecolor{CB_pink}{HTML}{FFAABB}
\definecolor{CB_lightCyan}{HTML}{99DDFF}
\definecolor{CB_gray}{HTML}{DDDDDD}
\definecolor{CB_orange}{HTML}{EE8866}

\newcommand{\fon}[1]{\fontfamily{#1}\selectfont} 

\usepackage[T1]{fontenc}

\usepackage[utf8]{inputenc}

\usepackage{microtype}
\usepackage{enumitem}

\title{Is ChatGPT a Good Teacher Coach?\\ Measuring Zero-Shot Performance For Scoring and \\Providing Actionable Insights on Classroom Instruction}

\author{Rose Wang\\
    \texttt{rewang@cs.stanford.edu} \\
    Stanford University
    \And
  Dorottya Demszky \\
  \texttt{ddemszky@stanford.edu} \\
  Stanford University
  }

\begin{document}
\maketitle
\begin{abstract}

Coaching, which involves classroom observation and expert feedback, is a widespread and fundamental part of teacher training. However, the majority of teachers do not have access to consistent, high quality coaching due to limited resources and access to expertise. We explore whether generative AI could become a cost-effective complement to expert feedback by serving as an automated teacher coach. In doing so, we propose three teacher coaching tasks for generative AI: (A) scoring transcript segments based on classroom observation instruments, (B)
identifying highlights and missed opportunities for good instructional strategies, and (C) providing actionable suggestions for eliciting more student reasoning. 
We recruit expert math teachers to evaluate the zero-shot performance of ChatGPT on each of these tasks for elementary math classroom transcripts. Our results reveal that ChatGPT generates responses that are relevant to improving instruction, but they are often not novel or insightful.  For example, $82$\% of the model's suggestions point to places in the transcript where the teacher is already implementing that suggestion. Our work highlights the challenges of producing insightful, novel and truthful feedback for teachers while paving the way for future research to address these obstacles and improve the capacity of generative AI to coach teachers.\footnote{The code and model outputs are open-sourced here: \url{https://github.com/rosewang2008/zero-shot-teacher-feedback}.}

\end{abstract}

\section{Introduction}

Classroom observation, coupled with coaching, is the cornerstone of teacher education and professional development internationally \citep{adelman2003guide,wragg2011introduction, martinez2016classroom, desimone2017instructional}. In the United States, teachers typically receive feedback from school administrators or instructional coaches, who assess teachers based on predetermined criteria and rubrics. These structured evaluations often involve pre- and post-observation conferences, where the observer and teacher discuss teaching strategies and reflect on the observed instruction.

Despite its widespread adoption, classroom observation lacks consistency across schools and different learning contexts due to time and resource constraints, human subjectivity, and varying levels of expertise among observers \citep{KraftBlazarHogan2018,kelly2020using}. Frequency and quality of feedback can vary significantly from one school or learning context to another, resulting in disparities in teacher development opportunities and, consequently, student outcomes.

Prior work has sought to complement the limitations of manual classroom observation by leveraging natural language processing (NLP) to provide teachers with scalable, automated feedback on instructional practice \citep{demszky2021feedback,SureshJacobsLaiTanWardMartinSumner2021}. 
These approaches offer low-level statistics of instruction, such as the frequency of teaching strategies employed in the classroom--- different from the high-level, actionable feedback provided during coaching practice. Receiving high-level, actionable feedback automatically could be easier for teachers to interpret than low level statistics, and such feedback also aligns more closely with existing forms of coaching.

\begin{figure*}[t]
    \centering
    \includegraphics[width=\linewidth]{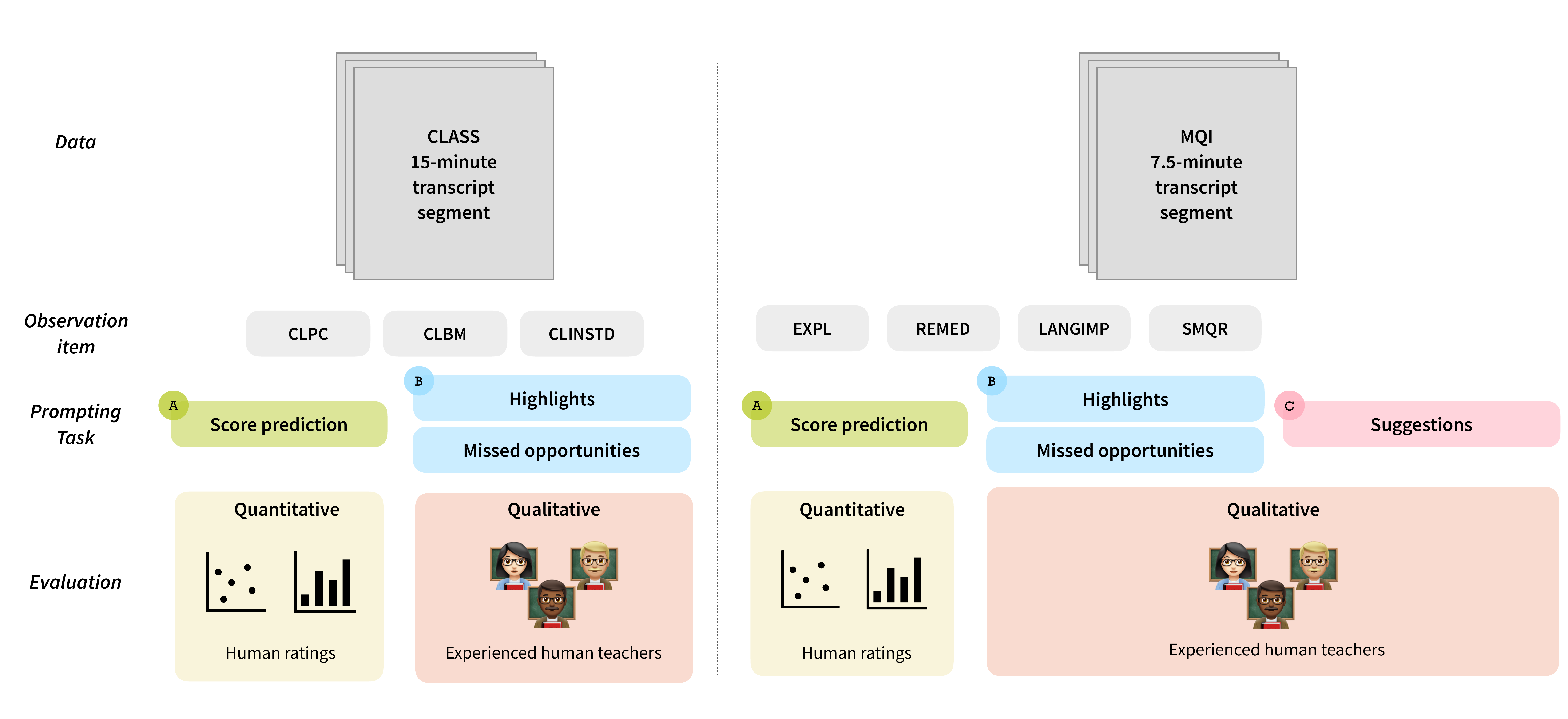}    
    \caption{Setup for the automated feedback task. Our work proposes three teacher coaching tasks. Task A is to score a transcript segment for items derived from classroom observation instruments; for instance, 
    \classPositiveClimate, \classBehavioralManagement, and \classInstructionalDialogue{} are CLASS observation items, and \mqiExplanations, \mqiRemediation, \mqiErrors, \mqiStudentMath{} are MQI observation items.
    Task B is to identify highlights and missed opportunities for good instructional strategies. 
    Task C is to provide actionable suggestions for eliciting more student reasoning. 
    \label{fig:main_figure}}
\end{figure*}

Recent advances in NLP have resulted in models like ChatGPT that have remarkable few-shot and zero-shot abilities.
ChatGPT has been applied to various NLP tasks relevant to education, such as essay writing \citep{basic2023better} or assisting on mathematics problems \citep{pardos2023learning}, and providing essay feedback to students \citep{dai2023can}.
A survey conducted by the Walton Family Foundation shows that 40\% of teachers use ChatGPT on a weekly basis for tasks such as lesson planning and building background knowledge for lessons \citep{impact_foundation_chatgpt_nodate}. 
Given ChatGPT's potential and teachers' growing familiarity with it, we are interested in the following research question:
Can ChatGPT help instructional coaches and teachers by providing effective feedback, like generating classroom observation rubric scores and helpful pedagogical suggestions?

To answer this question, we propose the following teacher coaching tasks for generative AI.

\begin{enumerate}[label=\textbf{Task \Alph*.}, leftmargin=4em]
    \item \textit{Score} a transcript segment for items derived from classroom observation instruments
    \item   \textit{Identify highlights and missed opportunities} for good instructional strategies
    \item  \textit{Provide actionable suggestions} for eliciting more student reasoning
\end{enumerate}

We evaluate the performance of ChatGPT with zero-shot prompting on each of these tasks via the process in Figure~\ref{fig:main_figure}. 
We use the NCTE dataset \citep{demszky2022ncte}, a large dataset of elementary math classroom transcripts.
The data is annotated by experts with two observation protocols: the Classroom Assessment Scoring System (CLASS) \citep{pianta2008classroom} and Mathematical Quality Instruction (MQI) \citep{hill2008mathematical} instruments. 
We prompt ChatGPT to score segments from these transcripts (Task A) and to identify highlights and missed opportunities (Task B) with respect to items derived from CLASS and MQI. 
Finally, we prompt the model to generate suggestions to the teacher for eliciting more student mathematical reasoning in the classroom (Task C). 
We evaluate ChatGPT by comparing the model's numerical predictions to raters' scores in the NCTE data (Task A).
We also recruit math teachers to rate the ChatGPT's responses along multiple helpfulness criteria (Tasks B \& C).

We find that ChatGPT has significant room for improvement in all three tasks, but still holds promise for providing scalable high-quality feedback. 
On predicting scores, ChatGPT has low correlation with human ratings across all observation items even with added rubric information and reasoning.
On identifying highlights and missed opportunities, ChatGPT generates responses that are often not insightful ($50$-$70$\%) or relevant ($35$-$50$\%) to what is being asked for by both instruments. 
Finally, the majority of suggestions generated by ChatGPT ($82$\%) describe what the teacher already does in the transcript. 
Nonetheless, the model does generate a majority of suggestions that are actionable and faithfully interpret the teaching context.
We believe that with further development, ChatGPT can become a valuable tool for instructional coaches and teachers.
Our work highlights an exciting area for future research to improve on the current limitations of automated feedback systems.

In sum, we make the following contributions: we (1) propose three teacher coaching tasks for generative AI, (2) recruit expert teachers to evaluate ChatGPT's zero-shot performance on these tasks given elementary math classroom transcripts, (3) demonstrate that ChatGPT is useful in some aspects but still has a lot of room for improvement, and finally (4) highlight directions for future directions towards providing useful feedback to teachers.

\section{Related Work}

\paragraph{Automated feedback to educators.} 
Prior works on automated feedback tools provide analytics on student engagement and progress \citep[][among others]{su2014developing,schwarz2018orchestrating,AslanAlyuzTanrioverMeteOkurDMelloArslanEsme2019,bonneton2020can,AlrajhiAlamriPereiraCristea2021}. 
These tools enable teachers to monitor student learning and intervene as needed.
Recent NLP advances are able to provide teachers feedback on their classroom discourse, promoting self-reflection and instructional development \citep{SameiOlneyKellyNystrandDMelloSBlanchardSunGlausGraesser2014,DonnellyBlanchardOlneyKellyNystrandDMello2017,KellyOlneyDonnellyNystrandDMello2018,JensenDaleDonnellyStoneKellyGodleyDMello2020}.
For example, \citet{SureshJacobsLaiTanWardMartinSumner2021} provides feedback to teachers on their teaching moves, such as how frequently the teacher revoices a student's idea or how frequently the teacher asks students to reason aloud.
\citet{jacobs2022promoting} provides evidence that K-12 math teachers receive this kind of feedback positively. 
A similar tool, M-Powering Teachers, provides feedback to teachers on their uptake of student ideas and demonstrates effectiveness in the 1-on-1 learning setting \citep{demszky2023mpowering} and online group instruction \citet{demszky2022can}. 
Altogether, these findings show a positive impact of cost-effective automated tools.
They prompt further investigations into what other types of automated feedback are effective. 
Our work constitutes one exploration in this area.

\paragraph{Testing zero-shot capabilities of ChatGPT.}
Recent works have measured the capabilities of ChatGPT for annotation on established datasets and benchmarks \citep{kuzman2023chatgpt, he2023annollm,gilardi2023chatgpt,dai2023can}. 
For example, in a non-education setting, \citet{gilardi2023chatgpt} evaluates the zero-shot ability of ChatGPT to classify tweets.
\citet{dai2023can} is a recent education work that investigates ChatGPT's zero-shot ability to provide feedback to students on business project proposals.
However, their study only utilizes a single broad prompt to solicit feedback and they do not evaluate for common model issues like hallucination \cite{Ji_2023}.
Our work proposes three concrete tasks to generate different forms of feedback for teachers, and our evaluation targets common qualitative issues in model generations. 
For other recent applications of ChatGPT, we refer the reader to \citet{liu2023summary}.

\section{Data \label{sec:data}}

We use the National Center for Teacher Effectiveness (NCTE) Transcript dataset \citep{demszky2022ncte} in this work, which is the largest publicly available dataset of U.S. classroom transcripts linked with classroom observation scores.
The dataset consists of 1,660 45-60 minute long 4th and 5th grade elementary mathematics observations collected by the NCTE between 2010-2013. The transcripts are anonymized and represent data from 317 teachers across 4 school districts that serve largely historically marginalized students.

Transcripts are derived from video recordings, which were scored by expert raters using two instruments at the time of the NCTE data collection:
the Classroom Assessment Scoring System (CLASS) \citep{pianta2008classroom} and Mathematical Quality Instruction (MQI) \citep{hill2008mathematical} instruments. We evaluate ChatGPT's ability to predict scores for both instruments, as described below.

\paragraph{The CLASS instrument.} CLASS is an observational instrument that assesses classroom quality in PK-12 classrooms along three main dimensions: \textit{Emotional Support}, \textit{Classroom Organization} and \textit{Instructional Support}. Each of these dimensions is measured by multiple observation items; we choose one item from each dimension to provide a proof-of-concept. For \textit{Emotional Support}, we focus on the \textsc{Positive Climate} (\classPositiveClimate) item, which measures the enjoyment and emotional connection that teachers have with students and that students have with their peers. For \textit{Classroom Organization}, we focus on the \textsc{Behavior Management} (\classBehavioralManagement) item which measures how well the teachers encourage positive behaviors and monitor, prevent and redirect misbehavior. Finally, for \textit{Instructional Support}, we focus on the \textsc{Instructional Dialogue} (\classInstructionalDialogue) dimension which measures how the teacher uses structured, cumulative questioning and discussion to guide and prompt students' understanding of content. Each item is scored on a scale of 1-7 where 1 is low and 7 is high. All items are scored on a 15-minute transcript segment, which is typically about a third or fourth of the full classroom duration.

\paragraph{The MQI instrument.} The MQI observation instrument assesses the mathematical quality of instruction, characterizing the rigor and richness of the mathematics in the lesson, along four dimensions: \textit{Richness of the Mathematics}, \textit{Working with Students and Mathematics}, \textit{Errors and Imprecision}, and \textit{Student Participation in Meaning-Making and Reasoning}.
Similar to CLASS, each of these dimensions is measured by several observation items and we select one from each. For \textit{Richness of the Mathematics}, we focus on the \textsc{Explanations} (\mqiExplanations) dimension which evaluates the quality of the teacher's mathematical explanations. For \textit{Working with Students and Mathematics}, we focus on the \textsc{Remediation of Student Errors and Difficulties} (\mqiRemediation) which measures how well the teacher remediates student errors and difficulties. For \textit{Errors and Imprecision}, we focus on the \textsc{Imprecision in Language or Notation} (\mqiErrors) dimension which measures the teacher's lack of precision in mathematical language or notation. 
Finally, for \textit{Student Participation in Meaning-Making and Reasoning}, we focus on the \textsc{Student Mathematical Questioning and Reasoning} (\mqiStudentMath) dimension which measures how well students engage in mathematical thinking.
These items are scored on scale of 1-3 where 1 is low and 3 is high. They are scored on a 7.5 minute transcript segment, which is typically a seventh or eighth of the full classroom duration.

\subsection{Pre-processing}
\label{ssec:preprocessing}

\paragraph{Transcript selection.}
Due to classroom noise and far-field audio, student talk often contains inaudible talk marked as ``[inaudible]''.
In preliminary experiments, we notice that ChatGPT often overinterprets classroom events when ``[inaudible]'' is present in the student's transcription. 
For example, the model misinterprets the transcription line ``student: [inaudible]'' as `` A student's response is inaudible, which may make them feel ignored or unimportant.'' or the line ``Fudge, banana, vanilla, strawberry, banana, vanilla, banana, [inaudible]. [...]'' as the teacher allowing students to talk over each other and interrupt the lesson.
To reduce the occurrences of the model overinterpreting the classroom events and best evaluate the model's ability to provide feedback, we only consider transcripts where less than 10\% of the student contributions includes an ``[inaudible]'' marker. 
Because these transcripts are very long and it would be costly to evaluate ChatGPT on all of the transcripts, we randomly pick 10 for the CLASS instrument and 10 for the MQI instrument to use.

\paragraph{Transcript segmentation.} The CLASS observation instrument applies to 15-minute segments and MQI to 7.5-minute segments. Each transcript has an annotation of the total number of CLASS segments and MQI segments. We split each transcript into segments by grouping utterances into equal-sized bins. For example, if a transcript has 3 CLASS segments and 300 utterances, we each segment will have 100 utterances each.

\paragraph{Segment formatting.}
In the \textit{quantitative} Task A experiments, every utterance in the  transcript segment is formatted as: ``<speaker>: <utterance>''. 
<speaker> is either the teacher or a student and <utterance> is the speaker's utterance.
In our \textit{qualitative} Task B and C experiments, we mark every utterance with a number. The utterance is formatted as: ``<utterance number>. <speaker>: <utterance>''.
We use utterance numbers in the qualitative experiments because our prompts ask the model to identify utterances when providing specific feedback. In contrast, the quantitative experiments evaluate the entire transcript segment holistically.

\section{Methods \label{sec:methods}}
We use the \texttt{gpt-3.5-turbo} model through the OpenAI API, the model that powers ChatGPT. 
We decode with temperature 0.
We employ zero-shot prompting in our study for three reasons. First, transcript segments are long, and the length of annotated example segments would exceed the maximum input size. Second, zero-shot prompting mimics most closely the current ways in which teachers interact with ChatGPT. Third, we are interested in evaluating ChatGPT's capabilities off-the-shelf, without additional tuning.

\subsection{Prompting}
We provide an overview of prompting methods. 
Appendix~\ref{app:prompts} contains all the prompts used in this work and information about how they are sourced. 

\paragraph{Task A: Scoring transcripts.} We zero-shot prompt ChatGPT to predict observation scores according to the CLASS and MQI rubrics. We employ three prompting techniques: (1) prompting to directly predict a score with 1-2 sentence summary of the item (\textit{direct answer}, \directAnswer) -- see example for \classBehavioralManagement{} in Figure~\ref{fig:example_task_a_output}, (2) same as \directAnswer{} but with additional one-sentence descriptions for low/mid/high ratings (\textit{direct answer with description}, \directAnswerDescription) and (3) same as \directAnswer{}, with asking the model to provide reasoning before predicting a score
 (\textit{reasoning then answer}, \reasoningAnswer).
\reasoningAnswer{} follows recent literature on LLM prompting with reasoning where models benefit from added reasoning on mathematical domains \citepinteralia{wei2022chain}. The item descriptions all derived from the original observation manuals, condensed to fit the context window of the model while accounting for space taken up by the transcript segment.
For all the prompts, the model correctly outputs integer values within each observation instrument's score range.

\paragraph{Task B: Identify highlights and missed opportunities.} We zero-shot prompt ChatGPT to identify and elaborate on highlights and missed opportunities for CLASS and MQI items. Specifically, we prompt ChatGPT to identify 5 good and bad examples (i.e. missed opportunities or poor execution) of each dimension.
The prompt includes numbered transcript sentences and asks the model to indicate the line number, before explaining the example.
See Figure~\ref{fig:example_task_b_output} for an example of the prompt and model outputs. 

\begin{figure}[h!]
    \centering 
    \small
    \begin{tcolorbox}[
    exampleprompt,
    title={\textbf{Example of Task B}},
    ]
    Consider the following classroom transcript.\\\\
    Transcript:\\
    1. student: Because --\\
    2. teacher: Nice and loud so we can hear you.\\
    3. student: I did 12 times 3 equals 36.\\
    4. teacher: Okay, 12 times 3 equals 36.\\
    5. student: Times 2, 36 times 2.\\
    6. teacher: Then 36 times 2 equals 72.  And that's your answer, right? Seventy-two inches.  Where'd you get the 12 from?\\
    \textbf{[...]}\\
    26. teacher: We're doing something different now.  I need everyone's attention up front.  Everyone should be looking up here. Paying attention to me.  Right now the question is right here.  How many inches are in one yard?\\
    \textbf{[...]}\\
    
    Please do the following.\\
    1. Provide up to 5 good examples of the teacher's behavior management. Behavior management encompasses the teacher's use of effective methods to encourage desirable behavior and prevent and re-direct misbehavior.\\
    2. Provide up to 5 bad examples (eg. missed opportunities or poor execution) of the teacher's behavior management.\\
    \textbf{[...]}
    \end{tcolorbox}
    \begin{tcolorbox}[
    modelresponse,
    title={\textbf{Model response}},
    ]
    \small
    Good examples: \\
    1. Line number: 2, Segment: ``Nice and loud so we can hear you.'', Reason: The teacher encourages the student to speak up and be heard. \\ 
    \textbf{[...]} \\
    Bad examples: \\ 
    \textbf{[...]} \\
    2. Line number: 26, Segment: ``We're doing something different now.'', Reason: The teacher abruptly changes the topic without providing any context or explanation.\\
    \textbf{[...]}
    \end{tcolorbox}
    \caption{
    The top box shows an example of task B for identifying highlights and missed opportunities on the teacher's behavior management (\classBehavioralManagement). 
    The bottom box shows the model's output. 
     For space reasons, the prompt and generations are abbreviated in this example, indicated by \textbf{[...]}.
    \label{fig:example_task_b_output}}
\end{figure}

\paragraph{Task C: Provide actionable suggestions for eliciting student reasoning.} We zero-shot prompt ChatGPT to provide actionable suggestions to the teacher for eliciting more student mathematical reasoning in the classroom. This task is open-ended: We simply ask the model to provide 5 suggestions for the teacher for eliciting more student reasoning in the classroom. We provide the model with a definition of student reasoning. The prompt includes numbered transcript sentences and asks the model to indicate the line number, before explaining the example. We use the MQI transcript segments because they are shorter and more likely to fit within the model's context window. 
See Figure~\ref{fig:example_task_c_output} for an example of the prompt and model outputs.

\begin{figure}[h!]
    \centering 
    \small
    \begin{tcolorbox}[
    suggestionprompt,
    title={\textbf{Example of Task C}},
    ]
    Consider the following classroom transcript.\\\\
    Transcript:\\
    1. student: Divisor.\\
    2. teacher: Divided.  It means the numerator divided by the denominator. That's what that line means.  We also know that when we see, when something's in division it also looks like this. That means divided by, but here it's just the line.  Then we see numbers right across the top here.  Let's read these numbers with me everybody.  Can everybody see?  What are they? 12.\\
    3. teacher: Good. Then along the sides we see some more numbers.  We've been forming one, two, three, four, five, six, seven, eight, nine, 10, 11, 12.  So the numbers across the top are called the what? Numerators.  The numbers down the side are called what? Denominators.  So this is almost going to be like a division table. \textbf{[...]}\\
    
    The transcript contains many short student responses. Please provide 5 suggestions for the teacher on how the teacher could elicit more student reasoning in the classroom. Student reasoning is counted broadly as students asking questions, engaging in mathematical discourse with their teacher or peers, and providing explanations such as justifying their answers. \\
    \textbf{[...]}\\
    
    Advice to the teacher:
    \end{tcolorbox}
    \begin{tcolorbox}[
    modelresponse,
    title={\textbf{Example of model output}},
    ]
    \small 
    1. Line number: 2, Segment: ``Let's read these numbers with me everybody. Can everybody see? What are they? 12.'', Suggestion: Instead of simply asking students to read the numbers, ask them to explain how they know the numbers are 12 and to provide examples of other numbers they see in the table.\\
    \textbf{[...]}
    \end{tcolorbox}
    \caption{Example on task C for providing suggestions on eliciting more student mathematical reasoning in the classroom. The model's output is italicized. For space reasons, the prompt and generations are abbreviated in this example, indicated by \textbf{[...]}.
    \label{fig:example_task_c_output}}
\end{figure}

\subsection{Validation \label{sec:validation_methods}}

We describe the analytical methods we use to answer each of the research questions. 

\paragraph{Task A.} The NCTE transcript dataset contains CLASS and MQI scores from human annotators. We compare ChatGPT's predictions against the human annotator scores. We randomly pick 100 transcript segments from the processed dataset (rf. Section~\ref{ssec:preprocessing}) for evaluation. We compute Spearman correlation to measure how well the model's predictions correspond to the human ratings. 
We also compare the distribution of human ratings vs model ratings via a histogram, to understand how well ChatGPT is calibrated for this task.

\paragraph{Task B.} We randomly pick 10 transcript segments and prompt the model to identify highlights and missed opportunities per observation item in CLASS and MQI.
We randomly select two highlights and two missed opportunities to be evaluated.

This results in \totalCLASSitems{} CLASS examples ($ =$ \numCLASSSegments{} segments $\times 3$ CLASS codes $\times$ ($2$ highlights $+2$ missed opportunities)) and \totalMQIitems{} MQI examples ($=$ \numMQISegments{} segments $\times 4$ MQI codes $\times $ ($2$ highlights $+2$ missed opportunities)).
We recruit two math teachers to evaluate the model's outputs: one of the teachers has decades of experience as an instructional coach, and the other has 6 years of math teaching experience in title 1 public schools. Examples were split evenly between the teachers.

Teacher are asked to rate each example along three criteria, which we identify based on preliminary experiments (e.g. observed hallucination) and by consulting the teachers.
\begin{enumerate}[leftmargin=1em]
    \item \textit{Relevance}: Is the model's response relevant to the CLASS or MQI item of interest?
    \item \textit{Faithfulness}: Does the model's response have an accurate interpretation of the events that occur in the classroom transcript? 
    \item \textit{Insightfulness}: Does the model's response reveal insights beyond a literal restatement of what happens in the transcript?
\end{enumerate} 

Each criteria is evaluated on a 3-point scale (yes, somewhat, no) with optional comments. For more details on the experimental setup and interrater comparison, please refer to Appendix~\ref{app:human_experiments}.

\paragraph{Task C.} 
We evaluate this task similarly to Task B, except for slight changes in the criteria. We prompt the model using the 
18 transcript segments from Task B to generate suggestions for eliciting more student reasoning. We randomly sample 2 suggestions per segment, resulting in $36$ examples. Examples were split evenly between annotators. We use the following evaluation criteria:

\begin{enumerate}[leftmargin=1em]
    \item \textit{Relevance}: Is the model's response relevant to eliciting more student reasoning?
    \item \textit{Faithfulness}: Does the model's response have the right interpretation of the events that occur in the classroom transcript?
    \item \textit{Actionability}: Is the model's suggestion something that the teacher can easily translate into practice for improving their teaching or encouraging student mathematical reasoning?
    \item \textit{Novelty}: Is the model suggesting something that the teacher already does or is it a novel suggestion? Note that the experimental interface asks about ``redundancy''; we reverse the rating here for consistency across criteria (higher= better).
\end{enumerate} 

Similar to the previous section, we ask the teachers to evaluate on a 3-point scale (yes, somewhat, no) with optional comments.

\section{Results \& Discussion}

\begin{figure*}[h!]
    \centering
    \begin{subfigure}{0.49\linewidth}
        \includegraphics[width=\linewidth]{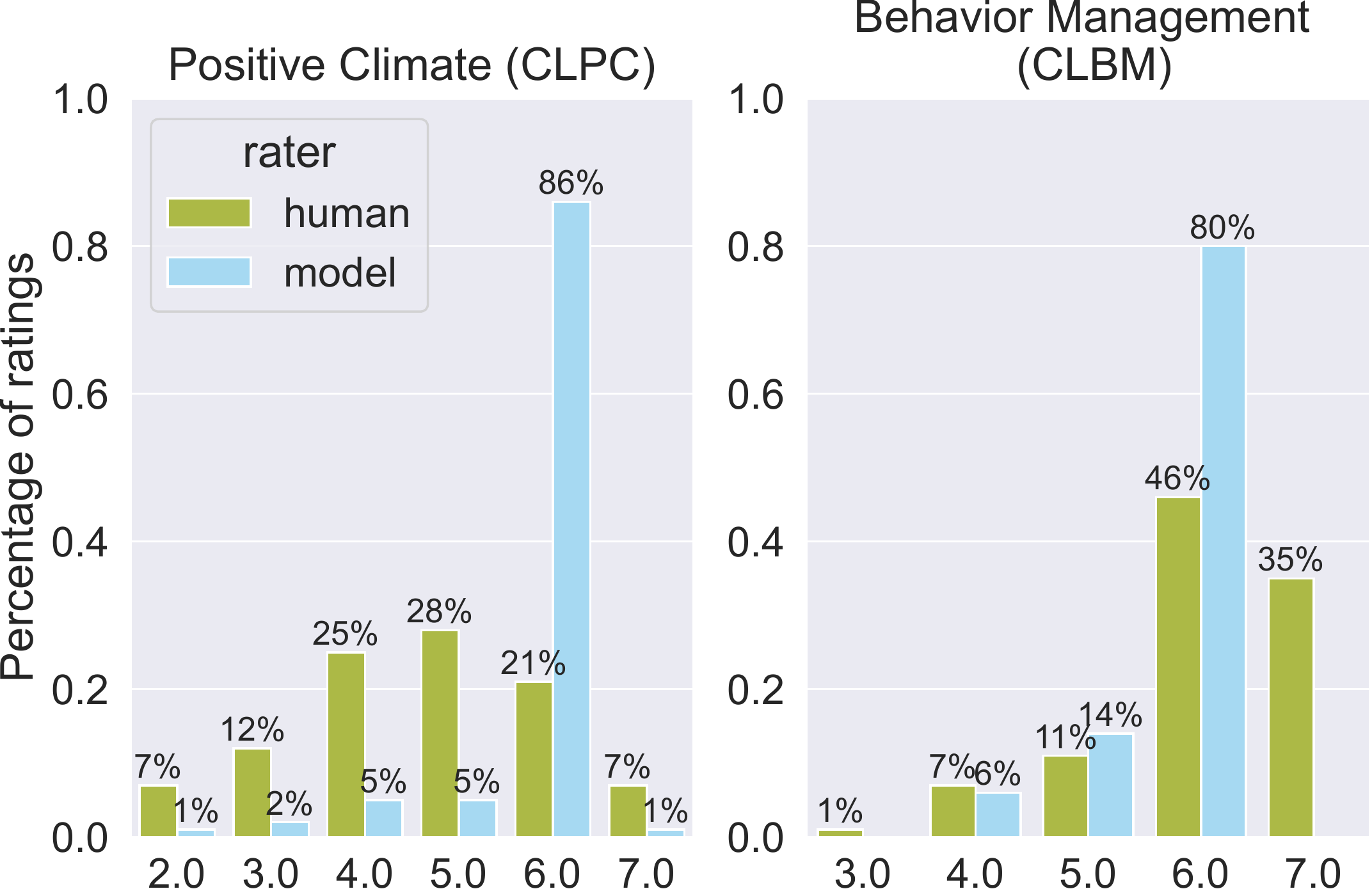}
        \caption{}
    \end{subfigure}
    \begin{subfigure}{0.49\linewidth}
        \includegraphics[width=\linewidth]{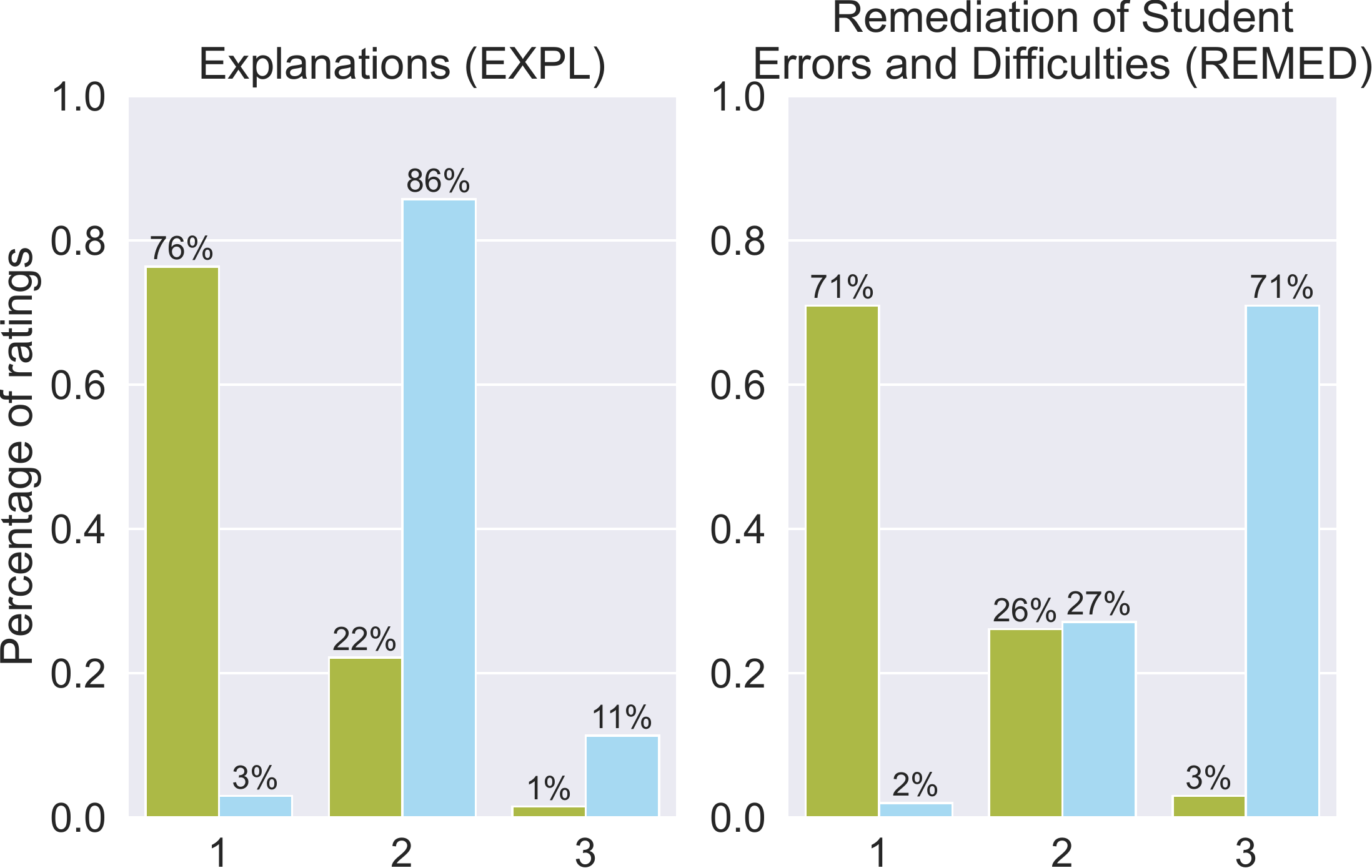}
        \caption{}
    \end{subfigure}
    \vspace{-0.5em}
    \caption{
    \textbf{Human and model distribution over scores for CLASS and MQI (Task A).} 
    The model scores are collected using \directAnswer{} prompting on (a) \classPositiveClimate{} and \classBehavioralManagement, and (b) \mqiExplanations{} and \mqiStudentMath.
    }
    \label{fig:histogram_scores}
\end{figure*}

\begin{table}[h]
    \centering
    \small
    \begin{tabular}{c c c c}
    \toprule
     & \classPositiveClimate{} &  \classBehavioralManagement{} & \classInstructionalDialogue{} \\ 
    \midrule
    \directAnswer{} & $0.00$ & $0.35$ & $-0.01$ \\ 
    \directAnswerDescription{} & $0.04$ & $0.23$ & $0.07$ \\ 
    \reasoningAnswer{} & $-0.06$ & $0.07$ & $-0.05$ \\
    \bottomrule
    \vspace{0.1em}
    \end{tabular}
    \begin{tabular}{c c c c c}
    \toprule
     & \mqiExplanations &  \mqiRemediation & \mqiErrors & \mqiStudentMath \\ 
    \midrule
    \directAnswer{} & $0.02$ & $0.05$ & $0.00$ & $0.17$  \\ 
    \directAnswerDescription{} & $0.12$ & $0.06$ & $0.02$ & $0.17$ \\ 
    \reasoningAnswer{} & $-0.11$ & $-0.06$ & $0.04$ & $0.06$ \\ 
    \bottomrule
    \end{tabular}
    \caption{
    The Spearman correlation values between the human scores and model predictions on the CLASS dimensions (top table) and MQI dimensions (bottom table). 
    The columns represent the different dimensions and the rows represent the different prompting methods discussed in Section~\ref{sec:methods}.
    }
    \label{tab:correlation_scores_spearman}
\end{table}

\paragraph{Task A: Scoring transcripts.} ChatGPT performs poorly at scoring transcripts both for MQI and CLASS items. 
Table~\ref{tab:correlation_scores_spearman} reports the Spearman correlation values, and Figure~\ref{fig:histogram_scores} reports the score distributions.
Appendix~\ref{app:additional_results} contains additional plots, including a comparison of the human vs. model score distributions.

As for CLASS, two findings are consistent across our prompting methods. 
First, the the model tends to predict higher values on all CLASS dimensions than human ratings and it performs best on \classBehavioralManagement{}. 
We hypothesize that \classBehavioralManagement{} may be easier to predict because (i) it is the only item whose distribution is skewed towards higher values and (ii) because scoring behavior management requires the least pedagogical expertise. 
Interestingly, adding more information to the prompt like per-score descriptions (\directAnswerDescription) or allowing for reasoning (\reasoningAnswer) did not improve the correlation score---in some cases making the score worse, such as for \classBehavioralManagement.

As for MQI, for all dimensions but \mqiRemediation{} the model tends to predict the middle score (2 out of 3); this observation is consistent across all prompting methods. Another interpretation of this finding, consistent with the CLASS results (which is on a 7 point scale), is that the model tends to predict the second to highest rating. We do not have sufficient data to disentangle these two interpretations.

For \mqiRemediation{}, the model generally predicts the highest rating (Figure~\ref{fig:histogram_scores}). 
Similar to the observations made in CLASS, adding more information or reasoning does not help the model. 
The model seems to pick up on \mqiStudentMath{} better than the other items, but its correlation decreases with both added information and reasoning.

Altogether, the models' tendency to predict the same scores for the same MQI or CLASS item suggest that the predicted scores are a function of the dimension description and not of the transcript evidence or the prompting methodology.

\begin{figure*}[t]
    \centering
    \begin{subfigure}{0.32\linewidth}
        \includegraphics[width=\linewidth]{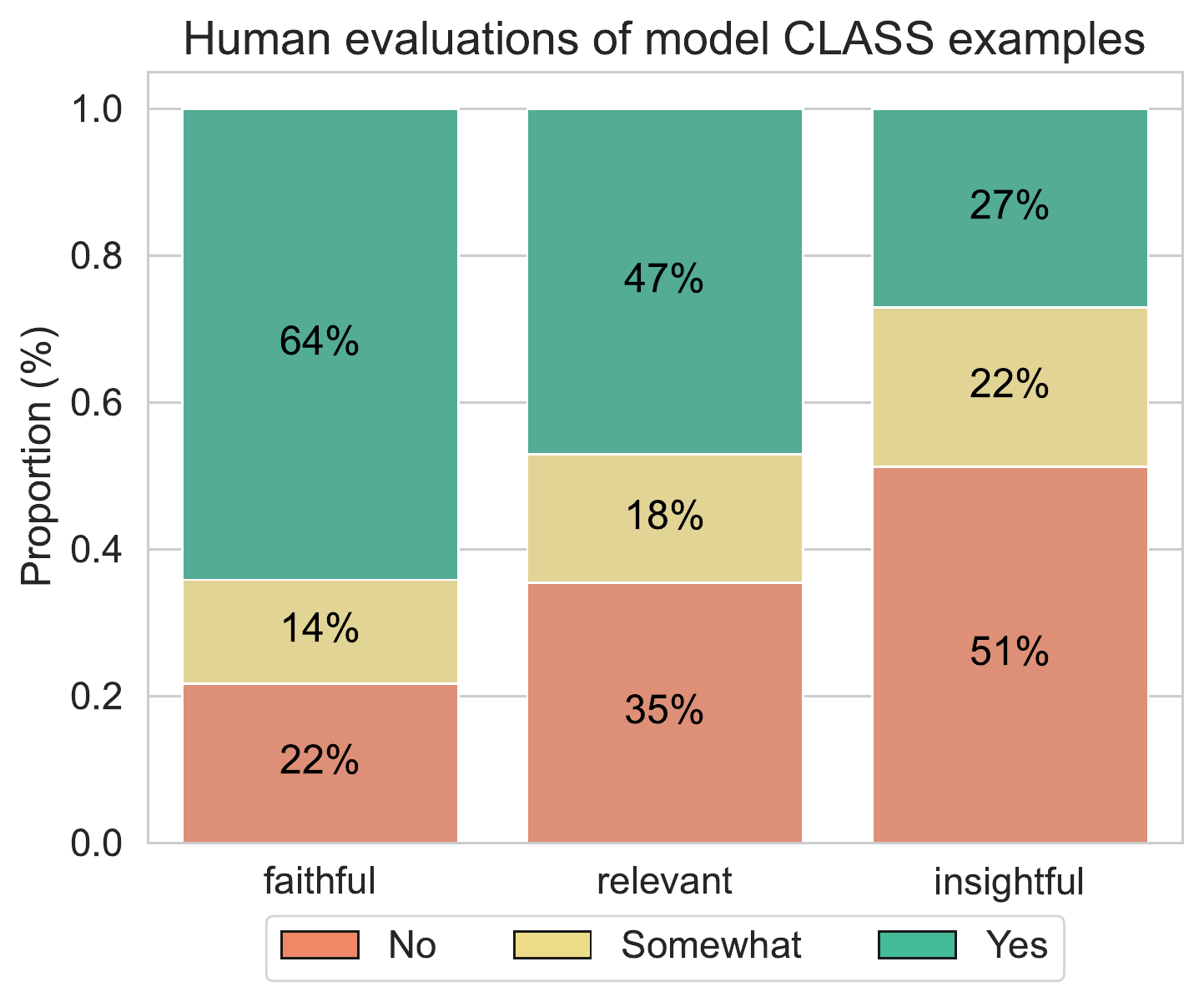}
        \caption{\label{fig:barplot_class_examples}}
    \end{subfigure}
    \begin{subfigure}{0.32\linewidth}
        \includegraphics[width=\linewidth]{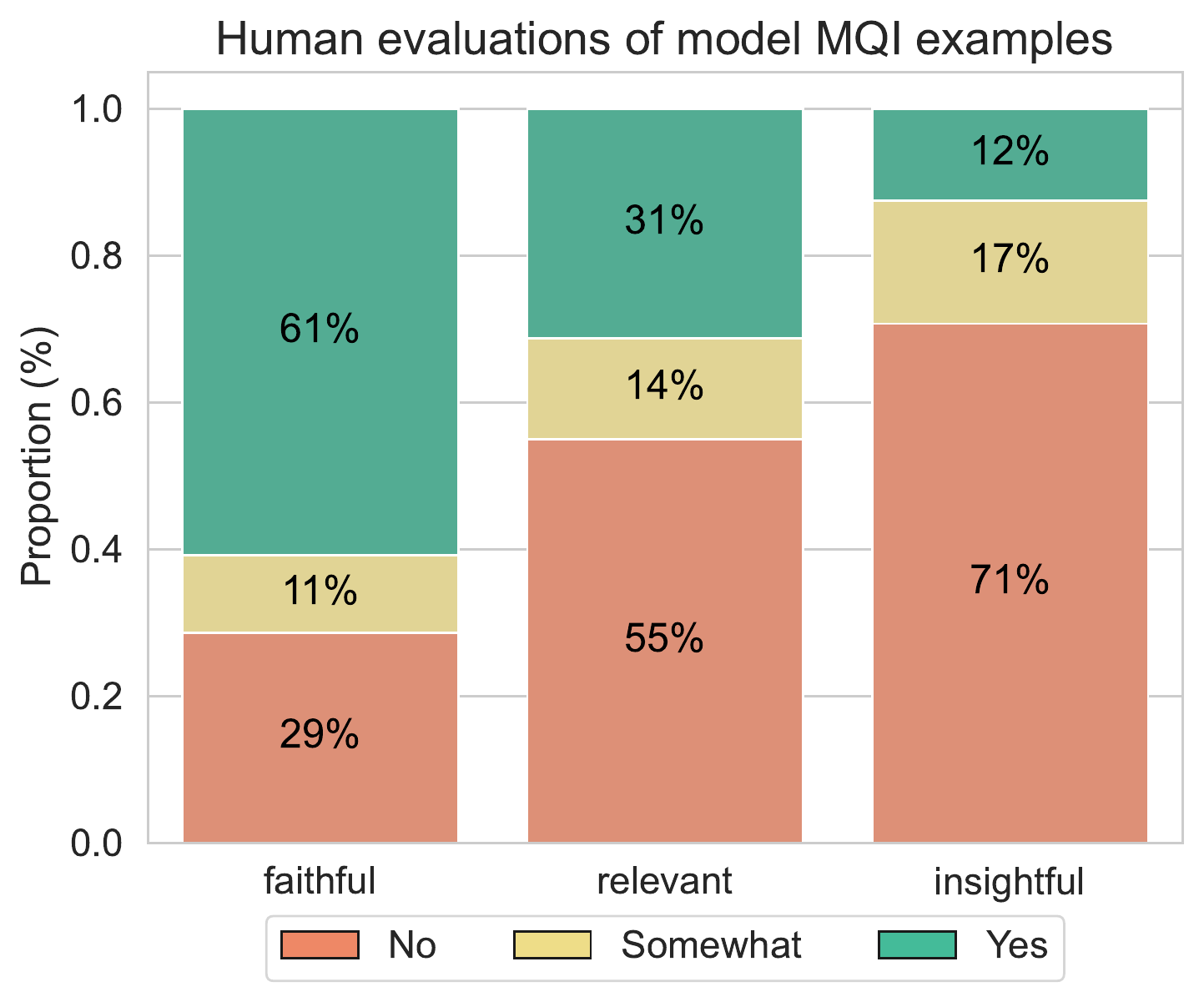}
        \caption{\label{fig:barplot_mqi_examples}}
    \end{subfigure}
    \begin{subfigure}{0.32\linewidth}
        \includegraphics[width=\linewidth]{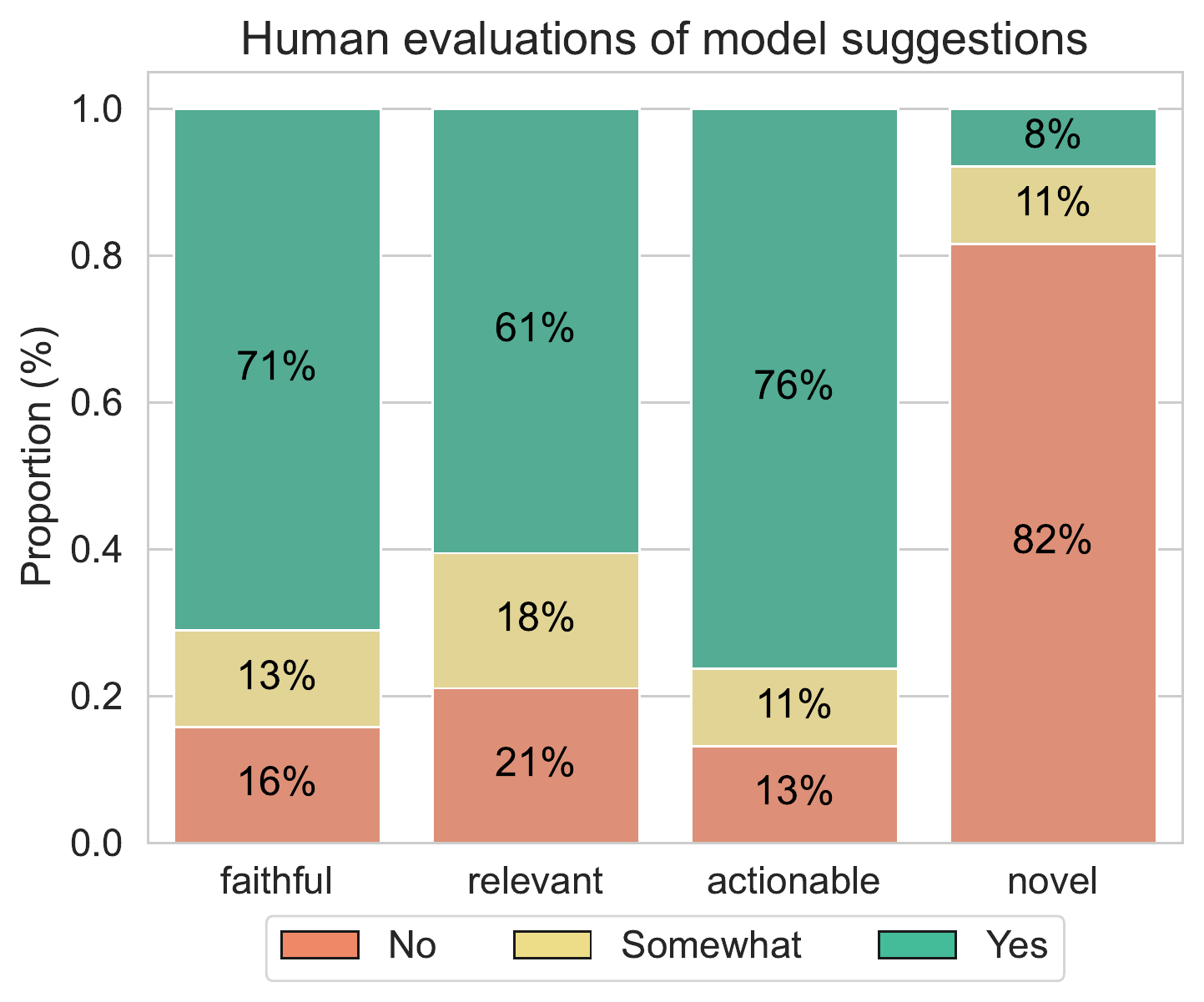}
        \caption{\label{fig:barplot_suggestions}}
    \end{subfigure}
    \vspace{-0.5em}
    \caption{
    Math teachers' evaluations for (a) highlights and missed opportunities (Task B) on CLASS items, (b) highlights and missed opportunities (Task B) on MQI items and (c) suggestions for eliciting more student reasoning (Task C).
    }
    \label{fig:stacked_barplots}
\end{figure*}

\paragraph{Task B: Identify highlights and missed opportunities.}

Figure~\ref{fig:barplot_class_examples} summarizes the ratings on model responses for the CLASS instrument, and Figure~\ref{fig:barplot_mqi_examples} for the MQI instrument.
Teachers generally did not find the model responses insightful or relevant to what was being asked for both instruments.
Hallucination, as rated by \textit{faithfulness}, is not the most problematic dimension out of the three. 
Nonetheless, it appears in a nontrivial amount of the model responses---around $20$-$30\%$ of the model responses are marked with being unfaithful in interpreting the classroom transcript.

Interestingly, the MQI results are worse than the CLASS results across all evaluation dimensions. 
Concretely, the ``No'' proportions increase on every dimension from CLASS$\rightarrow$MQI: Low scores on \textit{faithful} increase $22\rightarrow29\%$ ($+7$), \textit{relevant} $35\rightarrow55\%$ ($+20$), and \textit{insightful} $51\rightarrow71\%$ ($+20$). 
This suggests that the model performs relatively worse on interpreting and evaluating technical aspects of math instruction quality. 
Appendix~\ref{app:additional_results} contains additional plots, including the Cohen's kappa between raters. 

\paragraph{Task C: Provide actionable suggestions for eliciting student reasoning.}

Figure~\ref{fig:barplot_suggestions} summarizes the ratings on the model suggestions.
The most noticeable observation is that the model tends to produce redundant suggestions (opposite of \textit{novelty}), repeating what the teacher already does in the transcript $82\%$ of the time. 
Nonetheless, most model responses were rated to be \textit{faithful} to the transcript context, \textit{relevant} to eliciting more student reasoning, and \textit{actionable} for the teacher to implement. 

The results for Task B and C may be explained by the fact that ChatGPT was unlikely to see examples of instructional feedback, let alone examples of teacher coaching during its training, given the scarcity of publicly available data in this area. Thus, it has only learned to reproduce patterns already observed in the text, and not to produce out-of-the-box expert suggestions.

\section{Limitations}
This section discusses the limitations related to the evaluation process and potential ethical considerations associated with the use of ChatGPT or similar language models in educational settings.

\paragraph{Human evaluation}
Our evaluation is conducted with a limited sample size of two teachers. 
Future work should aim to include a larger and diverse sample of teachers to capture a wider range of perspectives.
This would help tease apart the potential teacher biases from generalizable claims about the feedback quality. 

\paragraph{Ethical considerations}
The use of language models like ChatGPT in educational contexts warrants careful examination. 
For example, because the model relies on transcribed speech and is trained on primarily English, it might misinterpret the transcriptions of teachers or students who do not speak English fluently. 
Additionally, deploying language models in education settings raises concerns regarding privacy and data security. 
For example, the raw classroom transcripts should not be directly fed into the model to provide feedback as it may contain personally identifiable information about students. 
Guardrails should be set to prevent classroom data from being sent directly to external companies.

\section{Avenues for Future Work}

As evidenced from our work, generating good feedback for teaching is \textit{challenging} and ChatGPT has significant room for improvement in this area. 
This section discusses potential future directions to overcome these obstacles.

\paragraph{Reducing hallucination.} 
Our results show that ChatGPT does generate a non-trivial amount of misleading responses as measured by our faithfulness dimension ($15$-$30$\% of the time). 
This observation is documented in the LLM literature as model hallucination \citep{Ji_2023}.
In domains that leverage references or citations such as in fact-checking, remedies include retrieving sources and checking the claims made by the model \citepinteralia{nakano2022webgpt,menick2022teaching}. 
In the domain of teacher feedback, however, it is not obvious what the ``true'' interpretation is, as even human observers may disagree slightly with respect to the teachers' intentions or actions. Future work could decrease hallucination in these higher inference domains, e.g. by forcing the model to be conservative with respect to making inferences.

\paragraph{Involving coaches and educators in model tuning.}

Our results show that ChatGPT struggles to generate insightful and novel feedback for teachers; understandably, since such feedback is not present in its training data. 
Involving coaches and educators in the reinforcement learning stage of model fine-tuning \citep{christiano2017deep} could be an effective way to improve the models' performance for teacher coaching. 
One less costly alternative is to engineer the model's prompt collaboratively with teachers and coaches. 
However, we are sceptical about the effectiveness of prompt engineering for teacher feedback, as it does not address model's lack of exposure to teacher coaching examples during training.

\paragraph{Tailoring feedback to a teacher's needs and expanding to other subjects.} What counts as helpful feedback may be different for each teacher, and look different in other subjects, eg. History and English. 
Even for the same teacher, what they \emph{self-report} to be helpful may be different from what what has a positive \emph{impact} on their practice. 
An effective coach takes this into account, and is able to dynamically adapt the feedback based on the teacher's needs and based on what they observe to be effective for that teacher \citep{thomas2015growth,kraft2018taking}. 
Improving ChatGPT's ability to differentiate feedback based on the teacher's needs, and update the feedback strategy based on teacher's subsequently observed practice would be a valuable direction for future work. 

To adapt our approach beyond mathematics, such as in subjects like History or English, researchers and instructors should collaborate and account for the subject's instructional practices and learning objectives.
This would help identify the relevant dimensions of effective teaching and inform the design of feedback prompts. 
For example, they can build on the subject-specific observation instruments as done in our work.

\paragraph{Integrating automated feedback into human coaching practice.} We envision automated coaching to complement, rather than replace coaching by experts for three reasons. First, as this paper shows, the capabilities of current technology is very far from that of an expert instructional coach. Second, even with improved technology, having an expert in the loop mitigates the risks of misleading or biased model outputs. Finally, even though automated feedback offers several benefits, including flexibility, scalability, privacy, lack of judgment, human interaction is still an important component of coaching and is perceived by teachers as such \citep{hunt2021automated}. Automated coaching could complement human coaching in a \emph{teacher-facing} way, e.g. by directly providing the teacher with feedback on-demand. Such an automated tool can also be \emph{coach-facing}, e.g. by generating diverse range of suggestions that the coach can then choose from based on what they think is most helpful for the teacher they are supporting.

\section{Conclusion}

Our work presents a step towards leveraging generative AI to complement the limitations of manual classroom observation and provide scalable, automated feedback on instructional practice. 
While our results reveal that ChatGPT has room for improvement in generating insightful and novel feedback for teaching, our proposed tasks and evaluation process provide a foundation for future research to address the challenges of teacher coaching using NLP. 
Our work underscores the challenge and importance of generating \textit{helpful} feedback for teacher coaching.
Moving forward, we propose several directions for further research, such as improved prompting methods and reinforcement learning with feedback from coaches. 
Ultimately, we envision a future where generative AI can play a crucial role in supporting effective teacher education and professional development, leading to improved outcomes for students.

\section*{Acknowledgements}
REW is supported by the National Science Foundation Graduate Research Fellowship. 
We thank Jiang Wu and Christine Kuzdzal for their helpful feedback.

\bibliography{anthology,custom}

\begin{thebibliography}{39}
\expandafter\ifx\csname natexlab\endcsname\relax\def\natexlab#1{#1}\fi

\bibitem[{Adelman and Walker(2003)}]{adelman2003guide}
Clement Adelman and Roy Walker. 2003.
\newblock \emph{A guide to classroom observation}.
\newblock Routledge.

\bibitem[{Alrajhi et~al.(2021)Alrajhi, Alamri, Pereira, and
  Cristea}]{AlrajhiAlamriPereiraCristea2021}
L.~Alrajhi, A.~Alamri, F.~D. Pereira, and A.~I. Cristea. 2021.
\newblock Urgency analysis of learners’ comments: An automated intervention
  priority model for mooc.
\newblock In \emph{International Conference on Intelligent Tutoring Systems},
  pages 148--160.

\bibitem[{Aslan et~al.(2019)Aslan, Alyuz, Tanriover, Mete, Okur, D'Mello, and
  Arslan~Esme}]{AslanAlyuzTanrioverMeteOkurDMelloArslanEsme2019}
S.~Aslan, N.~Alyuz, C.~Tanriover, S.~E. Mete, E.~Okur, S.~K. D'Mello, and
  A.~Arslan~Esme. 2019.
\newblock Investigating the impact of a real-time.
\newblock In \emph{multimodal student engagement analytics technology in
  authentic classrooms}, pages 1--12. of the 2019 CHI conference on human
  factors in computing systems.

\bibitem[{Basic et~al.(2023)Basic, Banovac, Kruzic, and
  Jerkovic}]{basic2023better}
Zeljana Basic, Ana Banovac, Ivana Kruzic, and Ivan Jerkovic. 2023.
\newblock \href {http://arxiv.org/abs/2302.04536} {Better by you, better than
  me, chatgpt3 as writing assistance in students essays}.

\bibitem[{Bonneton-Bott{\'e} et~al.(2020)Bonneton-Bott{\'e}, Fleury, Girard,
  Le~Magadou, Cherbonnier, Renault, Anquetil, and Jamet}]{bonneton2020can}
Nathalie Bonneton-Bott{\'e}, Sylvain Fleury, Nathalie Girard, Ma{\"e}lys
  Le~Magadou, Anthony Cherbonnier, Micka{\"e}l Renault, Eric Anquetil, and Eric
  Jamet. 2020.
\newblock Can tablet apps support the learning of handwriting? an investigation
  of learning outcomes in kindergarten classroom.
\newblock \emph{Computers \& Education}, 151:103831.

\bibitem[{Christiano et~al.(2017)Christiano, Leike, Brown, Martic, Legg, and
  Amodei}]{christiano2017deep}
Paul~F Christiano, Jan Leike, Tom Brown, Miljan Martic, Shane Legg, and Dario
  Amodei. 2017.
\newblock Deep reinforcement learning from human preferences.
\newblock \emph{Advances in neural information processing systems}, 30.

\bibitem[{Dai et~al.(2023)Dai, Lin, Jin, Li, Tsai, Gasevic, and
  Chen}]{dai2023can}
Wei Dai, Jionghao Lin, Flora Jin, Tongguang Li, Yi-Shan Tsai, Dragan Gasevic,
  and Guanliang Chen. 2023.
\newblock Can large language models provide feedback to students? a case study
  on chatgpt.

\bibitem[{Demszky and Hill(2022)}]{demszky2022ncte}
Dorottya Demszky and Heather Hill. 2022.
\newblock {The NCTE Transcripts}: A dataset of elementary math classroom
  transcripts.
\newblock \emph{arXiv preprint arXiv:2211.11772}.

\bibitem[{Demszky and Liu(2023)}]{demszky2023mpowering}
Dorottya Demszky and Jing Liu. 2023.
\newblock {M-Powering Teachers}: Natural language processing powered feedback
  improves 1:1 instruction and student outcomes.

\bibitem[{Demszky et~al.(2023{\natexlab{a}})Demszky, Liu, Hill, Jurafsky, and
  Piech}]{demszky2021feedback}
Dorottya Demszky, Jing Liu, Heather Hill, Dan Jurafsky, and Chris Piech.
  2023{\natexlab{a}}.
\newblock Can automated feedback improve teachers’ uptake of student ideas?
  evidence from a randomized controlled trial in a large-scale online.
\newblock \emph{Education Evaluation and Policy Analysis (EEPA)}.

\bibitem[{Demszky et~al.(2023{\natexlab{b}})Demszky, Liu, Hill, Jurafsky, and
  Piech}]{demszky2022can}
Dorottya Demszky, Jing Liu, Heather~C Hill, Dan Jurafsky, and Chris Piech.
  2023{\natexlab{b}}.
\newblock Can automated feedback improve teachers’ uptake of student ideas?
  evidence from a randomized controlled trial in a large-scale online course.
\newblock \emph{Educational Evaluation and Policy Analysis}.

\bibitem[{Desimone and Pak(2017)}]{desimone2017instructional}
Laura~M Desimone and Katie Pak. 2017.
\newblock Instructional coaching as high-quality professional development.
\newblock \emph{Theory into practice}, 56(1):3--12.

\bibitem[{Donnelly et~al.(2017)Donnelly, Blanchard, Olney, Kelly, Nystrand, and
  D’Mello}]{DonnellyBlanchardOlneyKellyNystrandDMello2017}
P.~J. Donnelly, N.~Blanchard, A.~M. Olney, S.~Kelly, M.~Nystrand, and S.~K.
  D’Mello. 2017.
\newblock Words matter: Automatic detection of teacher questions in live
  classroom discourse using linguistics, acoustics and context.
\newblock 218–227. Proceedings of the Seventh International Learning
  Analytics \& Knowledge Conference on - LAK ’17.

\bibitem[{Gilardi et~al.(2023)Gilardi, Alizadeh, and
  Kubli}]{gilardi2023chatgpt}
Fabrizio Gilardi, Meysam Alizadeh, and Maël Kubli. 2023.
\newblock \href {http://arxiv.org/abs/2303.15056} {Chatgpt outperforms
  crowd-workers for text-annotation tasks}.

\bibitem[{He et~al.(2023)He, Lin, Gong, Jin, Zhang, Lin, Jiao, Yiu, Duan, and
  Chen}]{he2023annollm}
Xingwei He, Zhenghao Lin, Yeyun Gong, A-Long Jin, Hang Zhang, Chen Lin, Jian
  Jiao, Siu~Ming Yiu, Nan Duan, and Weizhu Chen. 2023.
\newblock \href {http://arxiv.org/abs/2303.16854} {Annollm: Making large
  language models to be better crowdsourced annotators}.

\bibitem[{Hill et~al.(2008)Hill, Blunk, Charalambous, Lewis, Phelps, Sleep, and
  Ball}]{hill2008mathematical}
Heather~C Hill, Merrie~L Blunk, Charalambos~Y Charalambous, Jennifer~M Lewis,
  Geoffrey~C Phelps, Laurie Sleep, and Deborah~Loewenberg Ball. 2008.
\newblock Mathematical knowledge for teaching and the mathematical quality of
  instruction: An exploratory study.
\newblock \emph{Cognition and instruction}, 26(4):430--511.

\bibitem[{Hunt et~al.(2021)Hunt, Leijen, and van~der
  Schaaf}]{hunt2021automated}
Pihel Hunt, {\"A}li Leijen, and Marieke van~der Schaaf. 2021.
\newblock Automated feedback is nice and human presence makes it better:
  Teachers’ perceptions of feedback by means of an e-portfolio enhanced with
  learning analytics.
\newblock \emph{Education Sciences}, 11(6):278.

\bibitem[{Jacobs et~al.(2022)Jacobs, Scornavacco, Harty, Suresh, Lai, and
  Sumner}]{jacobs2022promoting}
Jennifer Jacobs, Karla Scornavacco, Charis Harty, Abhijit Suresh, Vivian Lai,
  and Tamara Sumner. 2022.
\newblock Promoting rich discussions in mathematics classrooms: Using
  personalized, automated feedback to support reflection and instructional
  change.
\newblock \emph{Teaching and Teacher Education}, 112:103631.

\bibitem[{Jensen et~al.(2020)Jensen, Dale, Donnelly, Stone, Kelly, Godley, and
  D’Mello}]{JensenDaleDonnellyStoneKellyGodleyDMello2020}
E.~Jensen, M.~Dale, P.~J. Donnelly, C.~Stone, S.~Kelly, A.~Godley, and S.~K.
  D’Mello. 2020.
\newblock Toward automated feedback on teacher discourse to enhance teacher
  learning.
\newblock In \emph{Proceedings of the 2020 CHI Conference on Human Factors in
  Computing Systems}. 1–13.

\bibitem[{Ji et~al.(2023)Ji, Lee, Frieske, Yu, Su, Xu, Ishii, Bang, Madotto,
  and Fung}]{Ji_2023}
Ziwei Ji, Nayeon Lee, Rita Frieske, Tiezheng Yu, Dan Su, Yan Xu, Etsuko Ishii,
  Ye~Jin Bang, Andrea Madotto, and Pascale Fung. 2023.
\newblock \href {https://doi.org/10.1145/3571730} {Survey of hallucination in
  natural language generation}.
\newblock \emph{{ACM} Computing Surveys}, 55(12):1--38.

\bibitem[{Kelly et~al.(2018)Kelly, Olney, Donnelly, Nystrand, and
  D’Mello}]{KellyOlneyDonnellyNystrandDMello2018}
S.~Kelly, A.~M. Olney, P.~Donnelly, M.~Nystrand, and S.~K. D’Mello. 2018.
\newblock \href {https://doi.org/10.3102/0013189X18785613} {Automatically
  measuring question authenticity in real-world classrooms}.
\newblock \emph{Educational Researcher}, 47:7.

\bibitem[{Kelly et~al.(2020)Kelly, Bringe, Aucejo, and
  Fruehwirth}]{kelly2020using}
Sean Kelly, Robert Bringe, Esteban Aucejo, and Jane~Cooley Fruehwirth. 2020.
\newblock Using global observation protocols to inform research on teaching
  effectiveness and school improvement: Strengths and emerging limitations.
\newblock \emph{Education Policy Analysis Archives}, 28:62--62.

\bibitem[{Kraft et~al.(2018)Kraft, Blazar, and Hogan}]{KraftBlazarHogan2018}
M.~A. Kraft, D.~Blazar, and D.~Hogan. 2018.
\newblock \href {https://doi.org/10.3102/0034654318759268} {The effect of
  teacher coaching on instruction and achievement: A meta-analysis of the
  causal evidence}.
\newblock \emph{Review of Educational Research}, 88(4):547–588.

\bibitem[{Kraft and Blazar(2018)}]{kraft2018taking}
Matthew~A Kraft and David Blazar. 2018.
\newblock Taking teacher coaching to scale: Can personalized training become
  standard practice?
\newblock \emph{Education Next}, 18(4):68--75.

\bibitem[{Kuzman et~al.(2023)Kuzman, Mozetic, and
  Ljube{\v{s}}ic}]{kuzman2023chatgpt}
Taja Kuzman, Igor Mozetic, and Nikola Ljube{\v{s}}ic. 2023.
\newblock Chatgpt: Beginning of an end of manual linguistic data annotation?
  use case of automatic genre identification.
\newblock \emph{arXiv e-prints}, pages arXiv--2303.

\bibitem[{Liu et~al.(2023)Liu, Han, Ma, Zhang, Yang, Tian, He, Li, He, Liu
  et~al.}]{liu2023summary}
Yiheng Liu, Tianle Han, Siyuan Ma, Jiayue Zhang, Yuanyuan Yang, Jiaming Tian,
  Hao He, Antong Li, Mengshen He, Zhengliang Liu, et~al. 2023.
\newblock Summary of chatgpt/gpt-4 research and perspective towards the future
  of large language models.
\newblock \emph{arXiv preprint arXiv:2304.01852}.

\bibitem[{Martinez et~al.(2016)Martinez, Taut, and
  Schaaf}]{martinez2016classroom}
Felipe Martinez, Sandy Taut, and Kevin Schaaf. 2016.
\newblock Classroom observation for evaluating and improving teaching: An
  international perspective.
\newblock \emph{Studies in Educational Evaluation}, 49:15--29.

\bibitem[{Menick et~al.(2022)Menick, Trebacz, Mikulik, Aslanides, Song,
  Chadwick, Glaese, Young, Campbell-Gillingham, Irving
  et~al.}]{menick2022teaching}
Jacob Menick, Maja Trebacz, Vladimir Mikulik, John Aslanides, Francis Song,
  Martin Chadwick, Mia Glaese, Susannah Young, Lucy Campbell-Gillingham,
  Geoffrey Irving, et~al. 2022.
\newblock Teaching language models to support answers with verified quotes.
\newblock \emph{arXiv preprint arXiv:2203.11147}.

\bibitem[{Nakano et~al.(2022)Nakano, Hilton, Balaji, Wu, Ouyang, Kim, Hesse,
  Jain, Kosaraju, Saunders, Jiang, Cobbe, Eloundou, Krueger, Button, Knight,
  Chess, and Schulman}]{nakano2022webgpt}
Reiichiro Nakano, Jacob Hilton, Suchir Balaji, Jeff Wu, Long Ouyang, Christina
  Kim, Christopher Hesse, Shantanu Jain, Vineet Kosaraju, William Saunders,
  Xu~Jiang, Karl Cobbe, Tyna Eloundou, Gretchen Krueger, Kevin Button, Matthew
  Knight, Benjamin Chess, and John Schulman. 2022.
\newblock \href {http://arxiv.org/abs/2112.09332} {Webgpt: Browser-assisted
  question-answering with human feedback}.

\bibitem[{Pardos and Bhandari(2023)}]{pardos2023learning}
Zachary~A. Pardos and Shreya Bhandari. 2023.
\newblock \href {http://arxiv.org/abs/2302.06871} {Learning gain differences
  between chatgpt and human tutor generated algebra hints}.

\bibitem[{Pianta et~al.(2008)Pianta, La~Paro, and Hamre}]{pianta2008classroom}
Robert~C Pianta, Karen~M La~Paro, and Bridget~K Hamre. 2008.
\newblock \emph{Classroom Assessment Scoring System™: Manual K-3.}
\newblock Paul H Brookes Publishing.

\bibitem[{Samei et~al.(2014)Samei, Olney, Kelly, Nystrand, D’Mello,
  Blanchard, Sun, Glaus, and
  Graesser}]{SameiOlneyKellyNystrandDMelloSBlanchardSunGlausGraesser2014}
B.~Samei, A.~M. Olney, S.~Kelly, M.~Nystrand, S.~D’Mello, N.~Blanchard,
  X.~Sun, M.~Glaus, and A.~Graesser. 2014.
\newblock \href {https://eric.ed.gov/?id=ED566380} {Domain independent
  assessment of dialogic properties of classroom discourse}.

\bibitem[{Schwarz et~al.(2018)Schwarz, Prusak, Swidan, Livny, Gal, and
  Segal}]{schwarz2018orchestrating}
Baruch~B Schwarz, Naomi Prusak, Osama Swidan, Adva Livny, Kobi Gal, and Avi
  Segal. 2018.
\newblock Orchestrating the emergence of conceptual learning: A case study in a
  geometry class.
\newblock \emph{International Journal of Computer-Supported Collaborative
  Learning}, 13:189--211.

\bibitem[{Su et~al.(2014)Su, Hsu, Chen, Huang, and Huang}]{su2014developing}
Yen-Ning Su, Chia-Cheng Hsu, Hsin-Chin Chen, Kuo-Kuang Huang, and Yueh-Min
  Huang. 2014.
\newblock Developing a sensor-based learning concentration detection system.
\newblock \emph{Engineering Computations}, 31(2):216--230.

\bibitem[{Suresh et~al.(2021)Suresh, Jacobs, Lai, Tan, Ward, Martin, and
  Sumner}]{SureshJacobsLaiTanWardMartinSumner2021}
A.~Suresh, J.~Jacobs, V.~Lai, C.~Tan, W.~Ward, J.~H. Martin, and T.~Sumner.
  2021.
\newblock \href {http://arxiv.org/abs/2105.07949} {Using transformers to
  provide teachers with personalized feedback on their classroom discourse: The
  talkmoves application. arxiv}.
\newblock Preprint.

\bibitem[{Thomas et~al.(2015)Thomas, Bell, Spelman, and
  Briody}]{thomas2015growth}
Earl~E Thomas, David~L Bell, Maureen Spelman, and Jennifer Briody. 2015.
\newblock The growth of instructional coaching partner conversations in a
  prek-3rd grade teacher professional development experience.
\newblock \emph{Journal of Adult Education}, 44(2):1--6.

\bibitem[{{Walton Family Foundation}(2023)}]{impact_foundation_chatgpt_nodate}
{Walton Family Foundation}. 2023.
\newblock \href
  {https://www.waltonfamilyfoundation.org/chatgpt-used-by-teachers-more-than-students-new-survey-from-walton-family-foundation-finds}
  {{ChatGPT} {Used} by {Teachers} {More} {Than} {Students}, {New} {Survey} from
  {Walton} {Family} {Foundation} {Finds}}.

\bibitem[{Wei et~al.(2022)Wei, Wang, Schuurmans, Bosma, Chi, Le, and
  Zhou}]{wei2022chain}
Jason Wei, Xuezhi Wang, Dale Schuurmans, Maarten Bosma, Ed~Chi, Quoc Le, and
  Denny Zhou. 2022.
\newblock Chain of thought prompting elicits reasoning in large language
  models.
\newblock \emph{arXiv preprint arXiv:2201.11903}.

\bibitem[{Wragg(2011)}]{wragg2011introduction}
Ted Wragg. 2011.
\newblock \emph{An introduction to classroom observation (Classic edition)}.
\newblock Routledge.

\end{thebibliography}

\newpage
\appendix

\begin{figure}[h]
    \centering 
    \small
    \begin{tcolorbox}[
    daprompt,
    title={\textbf{Example of Task A}},
    ]
    Consider the following classroom transcript.\\\\
    Transcript:\\
    student: Because --\\
    teacher: Nice and loud so we can hear you.\\
    student: I did 12 times 3 equals 36.\\
    teacher: Okay, 12 times 3 equals 36.\\
    student: Times 2, 36 times 2.\\
    teacher: Then 36 times 2 equals 72.  And that's your answer, right? Seventy-two inches.  Where'd you get the 12 from?\\
    \textbf{[...]}\\
    
    Based on the classroom transcript, rate the behavior management of the teacher on a scale of 1-7 (low-high). Behavior management encompasses the teacher's use of effective methods to encourage desirable behavior and prevent and re-direct misbehavior.\\ \\
    Rating (only specify a number between 1-7): 
    \end{tcolorbox}
    \begin{tcolorbox}[
    modelresponse,
    title={\textbf{Model response}},
    ]
    \small
    6
    \end{tcolorbox}
    \caption{The top box shows an example of task A for directly predicting the scores (\directAnswer) for behavior management (\classBehavioralManagement). The bottom box shows the model's output. For space reasons, the full transcript has been cut out, indicated by \textbf{[...]}.
    \label{fig:example_task_a_output}}
\end{figure}

\section{Prompts and decoding parameters}
\label{app:prompts}

This section provides all the prompts we used in our work and decoding parameters with using ChatGPT/\texttt{gpt-3.5-turbo}.
We used the OpenAI API to send queries to ChatGPT. 
We sampled from the model with temperature 0.

The subsections include the prompts for (a) scoring the teacher according to the CLASS and MQI rubric, (b) identifying highlights and missed opportunities and (c) providing actionable insights for teachers.

\begin{figure*}[h!]
    \centering
    \begin{subfigure}[b]{0.32\textwidth}
        \centering
        \includegraphics[width=\textwidth]{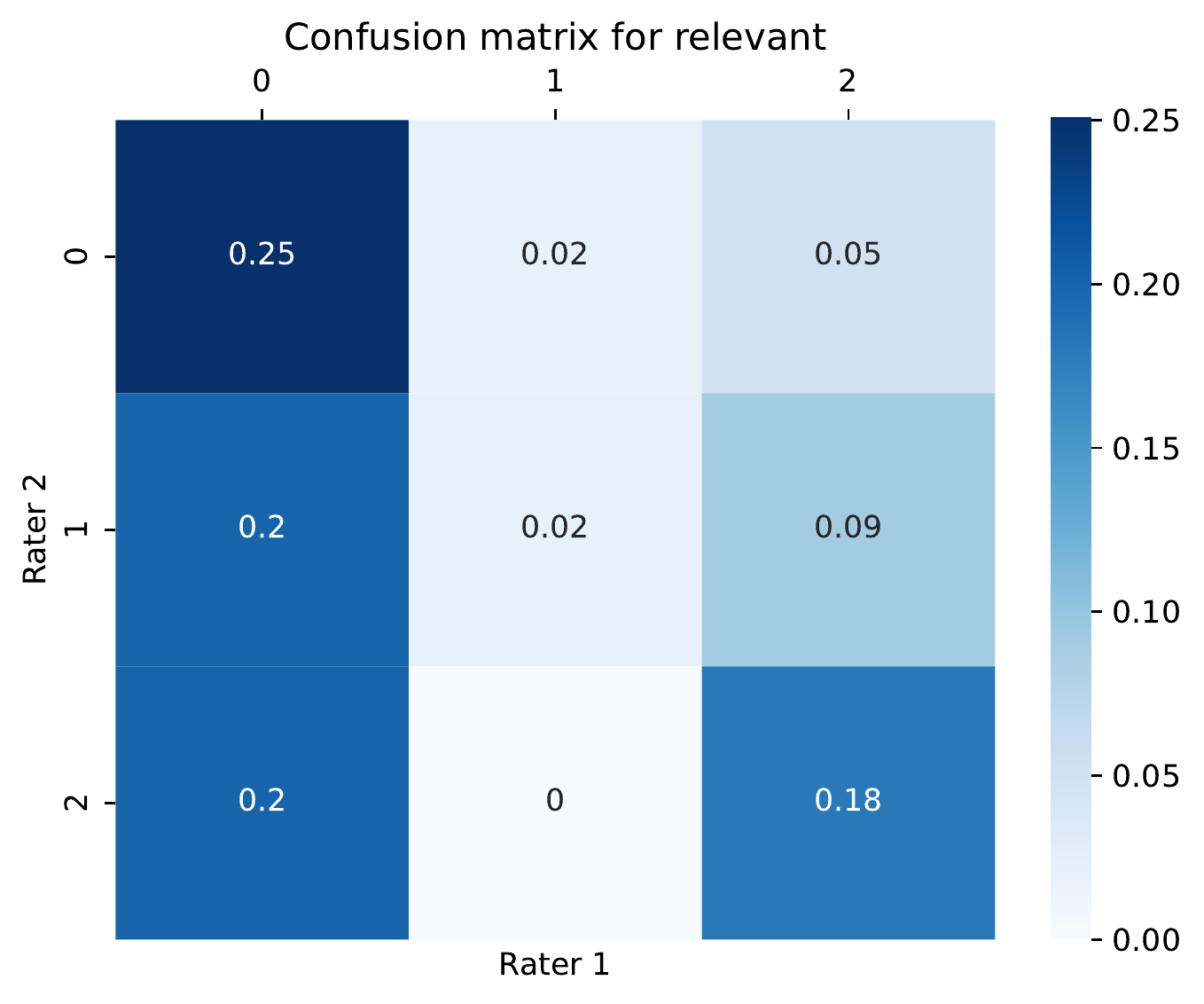}
        \caption{Relevance} 
    \end{subfigure}
    \hfill
    \begin{subfigure}[b]{0.32\textwidth}  
        \centering 
        \includegraphics[width=\textwidth]{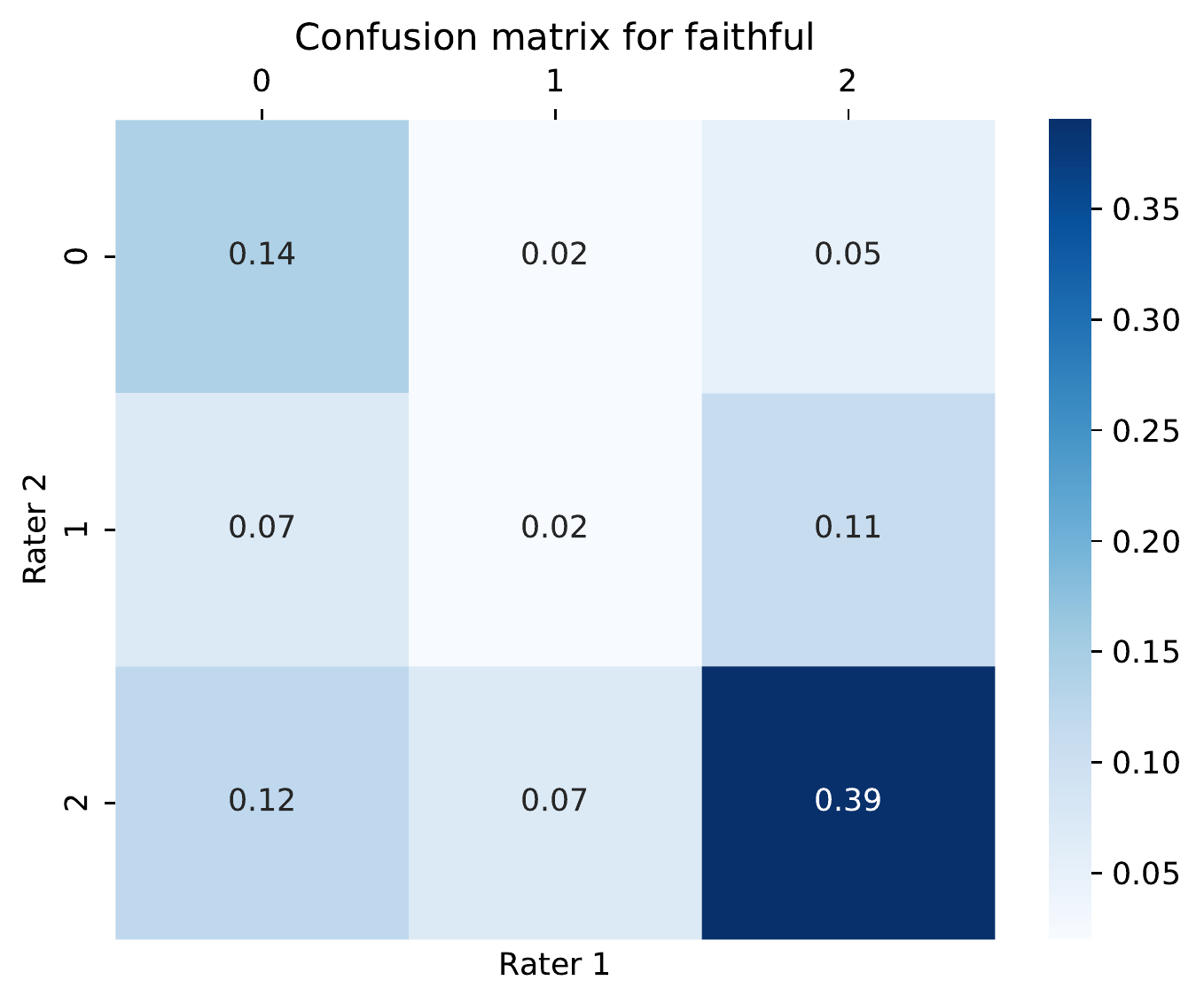}
        \caption{Faithfulness} 
    \end{subfigure}
    \hfill
    \begin{subfigure}[b]{0.32\textwidth}  
        \centering 
        \includegraphics[width=\textwidth]{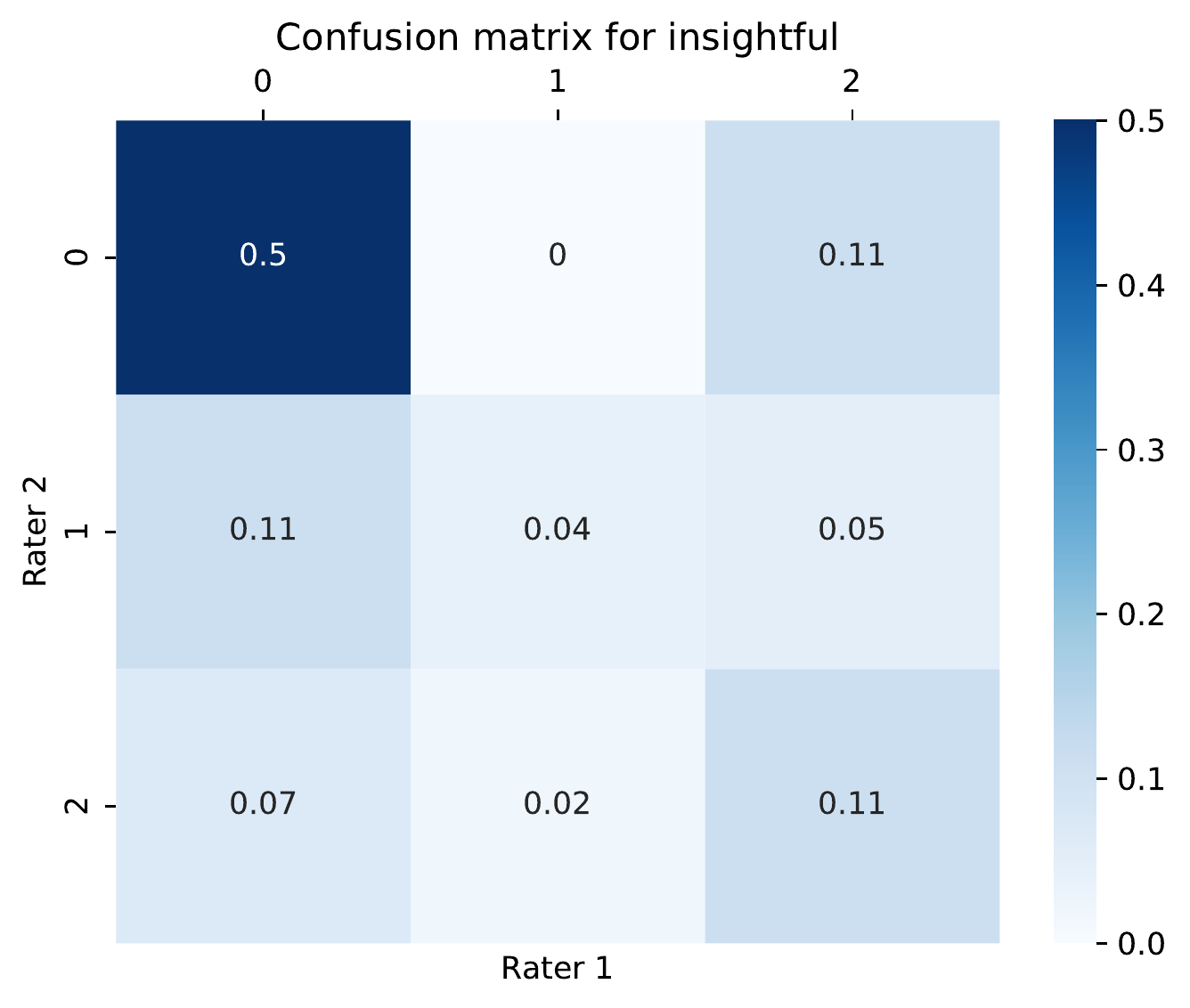}
        \caption{Insightfulness} 
    \end{subfigure}
    \caption{Confusion matrices between the two human raters on each of the criteria used in Task B: (a) \textit{relevance}, (b) \textit{faithfulness}, and (c) \textit{insightfulness}.}
    \label{fig:confusion_matrix_examples}
\end{figure*}

\subsection{Observation scores}

We prompt ChatGPT to provide scores according to the CLASS and MQI rubrics. 

Prompts for directly predicting the scores are shown in:

\begin{itemize}
    \item Figure~\ref{fig:da_clpc_prompt} for \classPositiveClimate.
    \item Figure~\ref{fig:da_clbm_prompt} for \classBehavioralManagement
    \item Figure~\ref{fig:da_clinstd_prompt} for \classInstructionalDialogue
    \item Figure~\ref{fig:da_expl_prompt} for \mqiExplanations
    \item Figure~\ref{fig:da_remed_prompt} for \mqiRemediation
    \item Figure~\ref{fig:da_langimp_prompt} for \mqiErrors
    \item Figure~\ref{fig:da_smqr_prompt} for \mqiStudentMath
\end{itemize}

\begin{figure*}[h]
    \centering 
    \begin{tcolorbox}[
    daprompt,
    title={\textbf{Prompt for direct score prediction (\directAnswer) on \classPositiveClimate}},
    ]
Consider the following classroom transcript. \\

Transcript: \\
\{transcript\} \\

Based on the classroom transcript, rate the positive climate of the classroom on a scale of 1-7 (low-high). Positive climate reflects the emotional connection and relationships among teachers and students, and the warmth, respect, and enjoyment communicated by verbal and non-verbal interactions.\\

Rating (only specify a number between 1-7):
    \end{tcolorbox}



    \caption{Prompt for directly predicting the scores (\directAnswer) on the CLASS dimension \classPositiveClimate. 
    \label{fig:da_clpc_prompt}}
\end{figure*}

\begin{figure*}[h]
    \centering 
    \begin{tcolorbox}[
    daprompt,
    title={\textbf{Prompt for direct score prediction (\directAnswer) on \classBehavioralManagement}},
    ]
        Consider the following classroom transcript.\\\\
        Transcript:\\
    \{transcript\}\\ \\
    Based on the classroom transcript, rate the behavior management of the teacher on a scale of 1-7 (low-high). Behavior management encompasses the teacher's use of effective methods to encourage desirable behavior and prevent and re-direct misbehavior.\\ \\
Rating (only specify a number between 1-7):
    \end{tcolorbox}
    \caption{Prompt for directly predicting the scores (\directAnswer) on the CLASS dimension \classBehavioralManagement. 
    \label{fig:da_clbm_prompt}}
\end{figure*}

\begin{figure*}[h]
    \centering 
    \begin{tcolorbox}[
    daprompt,
    title={\textbf{Prompt for direct score prediction (\directAnswer) on \classInstructionalDialogue}},
    ]
Consider the following classroom transcript. \\

Transcript: \\
\{transcript\} \\ 

Based on the classroom transcript, rate the instructional dialogue of the teacher on a scale of 1-7 (low-high). Instructional dialogue captures the purposeful use of content-focused discussion among teachers and students that is cumulative, with the teacher supporting students to chain ideas together in ways that lead to deeper understanding of content. Students take an active role in these dialogues and both the teacher and students use strategies that facilitate extended dialogue. \\ 

Rating (only specify a number between 1-7):
    \end{tcolorbox}
    \caption{Prompt for directly predicting the scores (\directAnswer) on the CLASS dimension \classInstructionalDialogue. 
    \label{fig:da_clinstd_prompt}}
\end{figure*}

\begin{figure*}[h]
    \centering 
    \begin{tcolorbox}[
    daprompt,
    title={\textbf{Prompt for direct score prediction (\directAnswer) on \mqiExplanations}},
    ]
Consider the following classroom transcript. \\

Transcript: \\
\{transcript\} \\

Based on the classroom transcript, rate the teacher's mathematical explanations on a scale of 1-3 (low-high). Mathematical explanations focus on the why, eg. why a procedure works, why a solution method is (in)appropriate, why an answer is true or not true, etc. Do not count `how', eg. description of the steps, or definitions unless meaning is also attached. \\

Rating (only specify a number between 1-3):
    \end{tcolorbox}
    \caption{Prompt for directly predicting the scores (\directAnswer) on the MQI dimension \mqiExplanations. 
    \label{fig:da_expl_prompt}}
\end{figure*}

\begin{figure*}[h]
    \centering 
    \begin{tcolorbox}[
    daprompt,
    title={\textbf{Prompt for direct score prediction (\directAnswer) on \mqiRemediation}},
    ]
Consider the following classroom transcript. \\

Transcript: \\
\{transcript\} \\

Based on the classroom transcript, rate the teacher's degree of remediation of student errors and difficulties on a scale of 1-3 (low-high). This means that the teacher gets at the root of student misunderstanding, rather than repairing just the procedure or fact. This is more than a simple correction of a student mistake. \\

Rating (only specify a number between 1-3):
    \end{tcolorbox}
    \caption{Prompt for directly predicting the scores (\directAnswer) on the MQI dimension \mqiRemediation. 
    \label{fig:da_remed_prompt}}
\end{figure*}

\begin{figure*}[h]
    \centering 
    \begin{tcolorbox}[
    daprompt,
    title={\textbf{Prompt for direct score prediction (\directAnswer) on \mqiErrors}},
    ]
Consider the following classroom transcript. \\

Transcript:\\
\{transcript\} \\

Based on the classroom transcript, rate the teacher's imprecision in language or notation on a scale of 1-3 (low-high). The teacher's imprecision in language or notation refers to problematic uses of mathematical language or notation. For example, errors in notation (eg. mathematical symbols), in mathematical language (eg. technical mathematical terms like "equation") or general language (eg. explaining mathematical ideas or procedures in non-technical terms). Do not count errors that are noticed and corrected within the segment.\\

Rating (only specify a number between 1-3):
    \end{tcolorbox}
    \caption{Prompt for directly predicting the scores (\directAnswer) on the MQI dimension \mqiErrors. 
    \label{fig:da_langimp_prompt}}
\end{figure*}

\begin{figure*}[h]
    \centering 
    \begin{tcolorbox}[
    daprompt,
    title={\textbf{Prompt for direct score prediction (\directAnswer) on \mqiStudentMath}},
    ]
Consider the following classroom transcript. \\

Transcript:\\
\{transcript\} \\

Based on the classroom transcript, rate the degree of student mathematical questioning and reasoning on a scale of 1-3 (low-high). Student mathematical questioning and reasoning means that students engage in mathematical thinking. Examples include but are not limited to: Students provide counter-claims in response to a proposed mathematical statement or idea, ask mathematically motivated questions requesting explanations, make conjectures about the mathematics discussed in the lesson, etc.\\

Rating (only specify a number between 1-3):
    \end{tcolorbox}
    \caption{Prompt for directly predicting the scores (\directAnswer) on the MQI dimension \mqiStudentMath. 
    \label{fig:da_smqr_prompt}}
\end{figure*}

Prompts for directly predicting the scores with additional rubric descriptions are shown in:

\begin{itemize}
    \item Figure~\ref{fig:dad_clpc_prompt} for \classPositiveClimate.
    \item Figure~\ref{fig:dad_clbm_prompt} for \classBehavioralManagement
    \item Figure~\ref{fig:dad_clinstd_prompt} for \classInstructionalDialogue
    \item Figure~\ref{fig:dad_expl_prompt} for \mqiExplanations
    \item Figure~\ref{fig:dad_remed_prompt} for \mqiRemediation
    \item Figure~\ref{fig:dad_langimp_prompt} for \mqiErrors
    \item Figure~\ref{fig:dad_smqr_prompt} for \mqiStudentMath
\end{itemize}

\begin{figure*}[h]
    \centering 
    \begin{tcolorbox}[
    daprompt,
    title={\textbf{Prompt with rubric description for direct score prediction  (\directAnswerDescription) on \classPositiveClimate}},
    ]
Consider the following classroom transcript. \\

Transcript:\\
\{transcript\} \\

Based on the classroom transcript, rate the positive climate of the classroom on a scale of 1-7 (low-high). Positive climate reflects the emotional connection and relationships among teachers and students, and the warmth, respect, and enjoyment communicated by verbal and non-verbal interactions.\\

Explanation of ratings: \\
1, 2: The teacher and students seem distant from one another, display flat affect, do not provide positive comments, or rarely demonstrate respect for one another.\\
3, 4, 5: There is some display of a supportive relationship, of positive affect, of positive communication, or of respect between the teacher and the students. \\
6, 7: There are many displays of a supportive relationship, of positive affect, of positive communication, or of respect between the teacher and the students.\\

Rating (only specify a number between 1-7):
    \end{tcolorbox}
    \caption{Prompt for directly predicting the scores (\directAnswerDescription) on the CLASS dimension \classPositiveClimate. 
    \label{fig:dad_clpc_prompt}}
\end{figure*}

\begin{figure*}[h]
    \centering 
    \begin{tcolorbox}[
    daprompt,
    title={\textbf{Prompt with rubric description for direct score prediction (\directAnswerDescription) on \classBehavioralManagement}},
    ]
    Consider the following classroom transcript.\\

Transcript: \\
\{transcript\} \\

Based on the classroom transcript, rate the behavior management of the teacher on a scale of 1-7 (low-high). Behavior management encompasses the teacher's use of effective methods to encourage desirable behavior and prevent and re-direct misbehavior. \\

Explanation of ratings: \\
1, 2: Teacher does not set expectations of the rules or inconsistently enforces them, teacher is reactive to behavioral issues or does not monitor students, teacher uses ineffective methods to redirect misbehavior, students are defiant. \\
3, 4, 5: Teacher sets some expectations of the rules but inconsistently enforces them, teacher uses a mix of proactive and reactive approaches to behavioral issues and sometimes monitors students, teacher uses a mix of effective and ineffective strategies to misdirect behavior, students periodically misbehave.\\
6, 7: Teacher sets clear expectations of the rules, teacher is proactive and monitors students, teacher consistently uses effective strategies to redirect mishavior, students are compliant.\\ 

Rating (only specify a number between 1-7):
    \end{tcolorbox}
    \caption{Prompt for directly predicting the scores (\directAnswerDescription) on the CLASS dimension \classBehavioralManagement. 
    \label{fig:dad_clbm_prompt}}
\end{figure*}

\begin{figure*}[h]
    \centering 
    \begin{tcolorbox}[
    daprompt,
    title={\textbf{Prompt with rubric description for direct score prediction  (\directAnswerDescription) on \classInstructionalDialogue}},
    ]
Consider the following classroom transcript. \\

Transcript: \\
\{transcript\} \\

Based on the classroom transcript, rate the instructional dialogue of the teacher on a scale of 1-7 (low-high). Instructional dialogue captures the purposeful use of content-focused discussion among teachers and students that is cumulative, with the teacher supporting students to chain ideas together in ways that lead to deeper understanding of content. Students take an active role in these dialogues and both the teacher and students use strategies that facilitate extended dialogue. \\

Explanation of ratings:\\
1, 2: There are no or few discussions in class or discussions unrelated to content, class is dominated by teacher talk, the teacher and students ask closed questions or rarely acknowledge/repeat/extend others' comments.\\
3, 4, 5: There are occasional brief content-based discussions in class among teachers and students, the class is mostly dominated by teacher talk, the teacher and students sometimes use facilitation strategies to encourage more elaborated dialogue.\\
6, 7: There are frequent, content-driven discussions in the class between teachers and students, class dialogues are distributed amongst the teacher and the majority of students, the teacher and students frequently use facilitation strategies that encourage more elaborated dialogue. \\

Rating (only specify a number between 1-7):
    \end{tcolorbox}
    \caption{Prompt for directly predicting the scores (\directAnswerDescription) on the CLASS dimension \classInstructionalDialogue. 
    \label{fig:dad_clinstd_prompt}}
\end{figure*}

\begin{figure*}[h]
    \centering 
    \begin{tcolorbox}[
    daprompt,
    title={\textbf{Prompt with rubric description for direct score prediction  (\directAnswerDescription) on \mqiExplanations}},
    ]
    Consider the following classroom transcript. \\

Transcript: \\
\{transcript\} \\

Based on the classroom transcript, rate the teacher's mathematical explanations on a scale of 1-3 (low-high).Mathematical explanations focus on the why, eg. why a procedure works, why a solution method is (in)appropriate, why an answer is true or not true, etc. Do not count `how', eg. description of the steps, or definitions unless meaning is also attached. \\

Explanation of ratings:\\
1: A mathematical explanation occurs as an isolated instance in the segment.\\
2: Two or more brief explanations occur in the segment OR an explanation is more than briefly present but not the focus of instruction.\\
3: One of more mathematical explanation(s) is a focus of instruction in the segment. The explanation(s) need not be most or even a majority of the segment; what distinguishes a High is the fact that the explanation(s) are a major feature of the teacher-student work (e.g., working for 2-3 minutes to elucidate the simplifying example above).\\

Rating (only specify a number between 1-3):
    \end{tcolorbox}
    \caption{Prompt for directly predicting the scores (\directAnswerDescription) on the CLASS dimension \mqiExplanations. 
    \label{fig:dad_expl_prompt}}
\end{figure*}

\begin{figure*}[h]
    \centering 
    \begin{tcolorbox}[
    daprompt,
    title={\textbf{Prompt with rubric description for direct score prediction (\directAnswerDescription) on \mqiRemediation}},
    ]
    Consider the following classroom transcript. \\

Transcript:\\
\{transcript\} \\

Based on the classroom transcript, rate the teacher's degree of remediation of student errors and difficulties on a scale of 1-3 (low-high). This means that the teacher gets at the root of student misunderstanding, rather than repairing just the procedure or fact. This is more than a simple correction of a student mistake.\\

Explanation of ratings:\\
1: Brief conceptual or procedural remediation occurs. \\
2: Moderate conceptual or procedural remediation occurs or brief pre-remediation (calling students' attention to a common error) occurs.\\
3: Teach engages in conceptual remediation systematically and at length. Examples include identifying the source of student errors or misconceptions, discussing how student errors illustrate broader misunderstanding and then addressing those erorrs, or extended pre-remediation.\\

Rating (only specify a number between 1-3):
    \end{tcolorbox}
    \caption{Prompt for directly predicting the scores (\directAnswerDescription) on the CLASS dimension \mqiRemediation. 
    \label{fig:dad_remed_prompt}}
\end{figure*}

\begin{figure*}[h]
    \centering 
    \begin{tcolorbox}[
    daprompt,
    title={\textbf{Prompt with rubric description for direct score prediction (\directAnswerDescription) on \mqiErrors}},
    ]
Consider the following classroom transcript.\\

Transcript:\\
\{transcript\} \\

Based on the classroom transcript, rate the teacher's imprecision in language or notation on a scale of 1-3 (low-high). The teacher's imprecision in language or notation refers to problematic uses of mathematical language or notation. For example, errors in notation (eg. mathematical symbols), in mathematical language (eg. technical mathematical terms like "equation") or general language (eg. explaining mathematical ideas or procedures in non-technical terms). Do not count errors that are noticed and corrected within the segment.\\

Explanation of ratings:\\
1: Brief instance of imprecision. Does not obscure the mathematics of the segment. \\
2: Imprecision occurs in part(s) of the segment or imprecision obscures the mathematics but for only part of the segment.\\
3: Imprecision occurs in most or all of the segment or imprecision obscures the mathematics of the segment.\\

Rating (only specify a number between 1-3):
    \end{tcolorbox}
    \caption{Prompt for directly predicting the scores (\directAnswerDescription) on the CLASS dimension \mqiErrors. 
    \label{fig:dad_langimp_prompt}}
\end{figure*}

\begin{figure*}[h]
    \centering 
    \begin{tcolorbox}[
    daprompt,
    title={\textbf{Prompt with rubric description for direct score prediction (\directAnswerDescription) on \mqiStudentMath}},
    ]    
Consider the following classroom transcript. \\

Transcript:  \\
\{transcript\}  \\

Based on the classroom transcript, rate the degree of student mathematical questioning and reasoning on a scale of 1-3 (low-high). Student mathematical questioning and reasoning means that students engage in mathematical thinking. Examples include but are not limited to: Students provide counter-claims in response to a proposed mathematical statement or idea, ask mathematically motivated questions requesting explanations, make conjectures about the mathematics discussed in the lesson, etc.  \\

Explanation of ratings: \\
1: One of two instances of brief student mathematical questioning or reasoning are present. \\
2: Student mathematical questioning or reasoning is more sustained or more frequent, but it is not characteristic of the segment. \\
3: Student mathematical questioning or reasoning characterizes much of the segment. \\

Rating (only specify a number between 1-3):
    \end{tcolorbox}
    \caption{Prompt for directly predicting the scores (\directAnswerDescription) on the CLASS dimension \mqiStudentMath. 
    \label{fig:dad_smqr_prompt}}
\end{figure*}

Prompts for reasoning then predicting the scores are shown in:
\begin{itemize}
    \item Figure~\ref{fig:ra_clpc_prompt} for \classPositiveClimate.
    \item Figure~\ref{fig:ra_clbm_prompt} for \classBehavioralManagement
    \item Figure~\ref{fig:ra_clinstd_prompt} for \classInstructionalDialogue
    \item Figure~\ref{fig:ra_expl_prompt} for \mqiExplanations
    \item Figure~\ref{fig:ra_remed_prompt} for \mqiRemediation
    \item Figure~\ref{fig:ra_langimp_prompt} for \mqiErrors
    \item Figure~\ref{fig:ra_smqr_prompt} for \mqiStudentMath
\end{itemize}

\begin{figure*}[h]
    \centering 
    \begin{tcolorbox}[
    daprompt,
    title={\textbf{Prompting with reasoning, then predicting the score (\reasoningAnswer) on \classPositiveClimate}},
    ]
    Consider the following classroom transcript. \\

Transcript:\\
\{transcript\}\\

Please do the following.\\
1. Think step-by-step how you would rate the positive climate of the classroom on a scale of 1-7 (low-high). Positive climate reflects the emotional connection and relationships among teachers and students, and the warmth, respect, and enjoyment communicated by verbal and non-verbal interactions.\\
2. Provide your rating as a number between 1 and 7.\\

Format your answer as:\\
Reasoning:\\
Rating (only specify a number between 1-7):\\

Reasoning:\\
    \end{tcolorbox}
    \caption{Prompt for reasoning, then predicting the score  (\reasoningAnswer) on the CLASS dimension \classPositiveClimate. 
    \label{fig:ra_clpc_prompt}}
\end{figure*}

\begin{figure*}[h]
    \centering 
    \begin{tcolorbox}[
    daprompt,
    title={\textbf{Prompting with reasoning, then predicting the score (\reasoningAnswer) on \classBehavioralManagement}},
    ]
    Consider the following classroom transcript. \\

Transcript:\\
\{transcript\}\\

Please do the following.\\
1. Think step-by-step how you would rate the behavior management of the teacher on a scale of 1-7 (low-high). Behavior management encompasses the teacher's use of effective methods to encourage desirable behavior and prevent and re-direct misbehavior.\\
2. Provide your rating as a number between 1 and 7.\\

Format your answer as:\\
Reasoning:\\
Rating (only specify a number between 1-7):\\

Reasoning:
    \end{tcolorbox}
    \caption{Prompt for reasoning, then predicting the score  (\reasoningAnswer) on the CLASS dimension \classBehavioralManagement. 
    \label{fig:ra_clbm_prompt}}
\end{figure*}

\begin{figure*}[h]
    \centering 
    \begin{tcolorbox}[
    daprompt,
    title={\textbf{Prompting with reasoning, then predicting the score (\reasoningAnswer) on \classInstructionalDialogue}},
    ]
    Consider the following classroom transcript. \\

Transcript:\\
\{transcript\}\\

Please do the following.\\
1. Think step-by-step how you would rate the instructional dialogue of the teacher on a scale of 1-7 (low-high). Instructional dialogue captures the purposeful use of content-focused discussion among teachers and students that is cumulative, with the teacher supporting students to chain ideas together in ways that lead to deeper understanding of content. Students take an active role in these dialogues and both the teacher and students use strategies that facilitate extended dialogue.\\
2. Provide your rating as a number between 1 and 7.\\

Format your answer as:\\
Reasoning:\\
Rating (only specify a number between 1-7):\\

Reasoning:
    \end{tcolorbox}
    \caption{Prompt for reasoning, then predicting the score  (\reasoningAnswer) on the CLASS dimension \classInstructionalDialogue. 
    \label{fig:ra_clinstd_prompt}}
\end{figure*}

\begin{figure*}[h]
    \centering 
    \begin{tcolorbox}[
    daprompt,
    title={\textbf{Prompting with reasoning, then predicting the score (\reasoningAnswer) on \mqiExplanations}},
    ]
    Consider the following classroom transcript. \\

Transcript:\\
\{transcript\}\\

Please do the following.\\
1. Think step-by-step how you would rate the teacher's mathematical explanations on a scale of 1-3 (low-high). Mathematical explanations focus on the why, eg. why a procedure works, why a solution method is (in)appropriate, why an answer is true or not true, etc. Do not count `how', eg. description of the steps, or definitions unless meaning is also attached.\\
2. Provide your rating as a number between 1 and 3.\\

Format your answer as:\\
Reasoning:\\
Rating (only specify a number between 1-3):\\

Reasoning:
    \end{tcolorbox}
    \caption{Prompt for reasoning, then predicting the score  (\reasoningAnswer) on the CLASS dimension \mqiExplanations. 
    \label{fig:ra_expl_prompt}}
\end{figure*}

\begin{figure*}[h]
    \centering 
    \begin{tcolorbox}[
    daprompt,
    title={\textbf{Prompting with reasoning, then predicting the score (\reasoningAnswer) on \mqiRemediation}},
    ]
    Consider the following classroom transcript.\\

Transcript:\\
\{transcript\}\\

Please do the following.\\
1. Think step-by-step how you would rate the teacher's degree of remediation of student errors and difficulties on a scale of 1-3 (low-high). This means that the teacher gets at the root of student misunderstanding, rather than repairing just the procedure or fact. This is more than a simple correction of a student mistake.\\
2. Provide your rating as a number between 1 and 3.\\

Format your answer as:\\
Reasoning:\\
Rating (only specify a number between 1-3):\\

Reasoning:
    \end{tcolorbox}
    \caption{Prompt for reasoning, then predicting the score  (\reasoningAnswer) on the CLASS dimension \mqiRemediation. 
    \label{fig:ra_remed_prompt}}
\end{figure*}

\begin{figure*}[h]
    \centering 
    \begin{tcolorbox}[
    daprompt,
    title={\textbf{Prompting with reasoning, then predicting the score (\reasoningAnswer) on \mqiErrors}},
    ]
Consider the following classroom transcript.\\
    
Transcript:\\
\{transcript\}\\

Please do the following.\\
1. Think step-by-step how you would rate the teacher's imprecision in language or notation on a scale of 1-3 (low-high). The teacher's imprecision in language or notation refers to problematic uses of mathematical language or notation. For example, errors in notation (eg. mathematical symbols), in mathematical language (eg. technical mathematical terms like "equation") or general language (eg. explaining mathematical ideas or procedures in non-technical terms). Do not count errors that are noticed and corrected within the segment.\\
2. Provide your rating as a number between 1 and 3.\\

Format your answer as:\\
Reasoning:\\
Rating (only specify a number between 1-3):\\

Reasoning:
    \end{tcolorbox}
    \caption{Prompt for reasoning, then predicting the score  (\reasoningAnswer) on the CLASS dimension \mqiErrors. 
    \label{fig:ra_langimp_prompt}}
\end{figure*}

\begin{figure*}[h]
    \centering 
    \begin{tcolorbox}[
    daprompt,
    title={\textbf{Prompting with reasoning, then predicting the score (\reasoningAnswer) on \mqiStudentMath}},
    ]
    Consider the following classroom transcript. \\

Transcript:\\
\{transcript\}\\

Please do the following.\\
1. Think step-by-step how you would rate the degree of student mathematical questioning and reasoning on a scale of 1-3 (low-high). Student mathematical questioning and reasoning means that students engage in mathematical thinking. Examples include but are not limited to: Students provide counter-claims in response to a proposed mathematical statement or idea, ask mathematically motivated questions requesting explanations, make conjectures about the mathematics discussed in the lesson, etc.\\
2. Provide your rating as a number between 1 and 3.\\

Format your answer as:\\
Reasoning:\\
Rating (only specify a number between 1-3):\\

Reasoning:
    \end{tcolorbox}
    \caption{Prompt for reasoning, then predicting the score  (\reasoningAnswer) on the CLASS dimension \mqiStudentMath. 
    \label{fig:ra_smqr_prompt}}
\end{figure*}

\subsection{Highlights and missed opportunities}

We prompt ChatGPT to identify highlights and missed opportunities according to the CLASS and MQI dimensions. 
The prompts for each dimension are shown in: 

\begin{itemize}
    \item Figure~\ref{fig:example_clpc_prompt} for \classPositiveClimate
    \item Figure~\ref{fig:example_clbm_prompt} for \classBehavioralManagement
    \item Figure~\ref{fig:example_clinstd_prompt} for \classInstructionalDialogue
    \item Figure~\ref{fig:example_expl_prompt} for \mqiExplanations
    \item Figure~\ref{fig:example_remed_prompt} for \mqiRemediation
    \item Figure~\ref{fig:example_langimp_prompt} for \mqiErrors
    \item Figure~\ref{fig:example_smqr_prompt} for \mqiStudentMath
\end{itemize}

\begin{figure*}[h]
    \centering 
    \begin{tcolorbox}[
    exampleprompt,
    title={\textbf{Prompt for identifying highlights and missed opportunity on \classPositiveClimate}},
    ]
    Consider the following classroom transcript.\\

Transcript:\\
\{transcript\}\\

Please do the following.\\
1. Provide up to 5 good examples of the classroom's positive climate. Positive climate reflects the emotional connection and relationships among teachers and students, and the warmth, respect, and enjoyment communicated by verbal and non-verbal interactions.\\
2. Provide up to 5 bad examples (eg. missed opportunities or poor execution) of the classroom's positive climate.\\

Format your answer as:\\
Good examples\\
1. Line number: <specify line number>, Segment: "<copied from transcript>", Reason: <specify why this is a good example>\\
2. ...\\

Bad examples\\
1. Line number: <specify line number>, Segment: "<copied from transcript>", Reason: <specify why this is a bad example>\\
2. ...\\

Good examples:
    \end{tcolorbox}
    \caption{Prompt for identifying highlights and missed opportunity on \classPositiveClimate. 
    \label{fig:example_clpc_prompt}}
\end{figure*}

\begin{figure*}[h]
    \centering 
    \begin{tcolorbox}[
    exampleprompt,
    title={\textbf{Prompt for identifying highlights and missed opportunity on \classBehavioralManagement}},
    ]
    Consider the following classroom transcript. \\

Transcript:\\
\{transcript\}\\

Please do the following.\\
1. Provide up to 5 good examples of the teacher's behavior management. Behavior management encompasses the teacher's use of effective methods to encourage desirable behavior and prevent and re-direct misbehavior.\\
2. Provide up to 5 bad examples (eg. missed opportunities or poor execution) of the teacher's behavior management.\\

Format your answer as:\\
Good examples\\
1. Line number: <specify line number>, Segment: "<copied from transcript>", Reason: <specify why this is a good example>\\
2. ...\\

Bad examples\\
1. Line number: <specify line number>, Segment: "<copied from transcript>", Reason: <specify why this is a bad example>\\
2. ...\\

Good examples:
    \end{tcolorbox}
    \caption{Prompt for identifying highlights and missed opportunity on \classBehavioralManagement. 
    \label{fig:example_clbm_prompt}}
\end{figure*}

\begin{figure*}[h]
    \centering 
    \begin{tcolorbox}[
    exampleprompt,
    title={\textbf{Prompt for identifying highlights and missed opportunity on \classInstructionalDialogue}},
    ]
    Consider the following classroom transcript.\\

Transcript:\\
\{transcript\}\\

Please do the following.\\
1. Provide up to 5 good examples of the teacher's instructional dialogue. Instructional dialogue captures the purposeful use of content-focused discussion among teachers and students that is cumulative, with the teacher supporting students to chain ideas together in ways that lead to deeper understanding of content. Students take an active role in these dialogues and both the teacher and students use strategies that facilitate extended dialogue.\\
2. Provide up to 5 bad examples of (eg. missed opportunities or poor execution) the teacher's instructional dialogue.\\

Format your answer as:\\
Good examples\\
1. Line number: <specify line number>, Segment: "<copied from transcript>", Reason: <specify why this is a good example>\\
2. ...\\

Bad examples\\
1. Line number: <specify line number>, Segment: "<copied from transcript>", Reason: <specify why this is a bad example>\\
2. ...\\

Good examples:
    \end{tcolorbox}
    \caption{Prompt for identifying highlights and missed opportunity on \classInstructionalDialogue. 
    \label{fig:example_clinstd_prompt}}
\end{figure*}

\begin{figure*}[h]
    \centering 
    \begin{tcolorbox}[
    exampleprompt,
    title={\textbf{Prompt for identifying highlights and missed opportunity on \mqiExplanations}},
    ]
    Consider the following classroom transcript.\\

Transcript:\\
\{transcript\}\\

Please do the following.\\
1. Provide up to 5 good examples of the teacher's mathematical explanations. Mathematical explanations focus on the why, eg. why a procedure works, why a solution method is (in)appropriate, why an answer is true or not true, etc. Do not count 'how', eg. description of the steps, or definitions unless meaning is also attached.\\
2. Provide up to 5 bad examples (eg. missed opportunities or poor execution) of the teacher's mathematical explanations.\\

Format your answer as:\\
Good examples\\
1. Line number: <specify line number>, Segment: "<copied from transcript>", Reason: <specify why this is a good example>\\
2. ...\\

Bad examples\\
1. Line number: <specify line number>, Segment: "<copied from transcript>", Reason: <specify why this is a bad example>\\
2. ...\\

Good examples:
    \end{tcolorbox}
    \caption{Prompt for identifying highlights and missed opportunity on \mqiExplanations. 
    \label{fig:example_expl_prompt}}
\end{figure*}

\begin{figure*}[h]
    \centering 
    \begin{tcolorbox}[
    exampleprompt,
    title={\textbf{Prompt for identifying highlights and missed opportunity on \mqiRemediation}},
    ]
    Consider the following classroom transcript.\\

Transcript:\\
\{transcript\}\\

Please do the following.\\
1. Provide up to 5 good examples of the teacher's remediation of student errors and difficulties. This means that the teacher gets at the root of student misunderstanding, rather than repairing just the procedure or fact. This is more than a simple correction of a student mistake.\\
2. Provide up to 5 bad examples (eg. missed opportunities or poor execution) of the teacher's remediation of student errors and difficulties.\\

Format your answer as:\\
Good examples\\
1. Line number: <specify line number>, Segment: "<copied from transcript>", Reason: <specify why this is a good example>\\
2. ...\\

Bad examples\\
1. Line number: <specify line number>, Segment: "<copied from transcript>", Reason: <specify why this is a bad example>\\
2. ...\\

Good examples:
    \end{tcolorbox}
    \caption{Prompt for identifying highlights and missed opportunity on \mqiRemediation. 
    \label{fig:example_remed_prompt}}
\end{figure*}

\begin{figure*}[h]
    \centering 
    \begin{tcolorbox}[
    exampleprompt,
    title={\textbf{Prompt for identifying highlights and missed opportunity on \mqiErrors}},
    ]
    Consider the following classroom transcript.\\

Transcript:\\
\{transcript\}\\

Please do the following.\\
1. Provide up to 5 good examples of the teacher's imprecision in language or notation. The teacher's imprecision in language or notation refers to problematic uses of mathematical language or notation. For example, errors in notation (eg. mathematical symbols), in mathematical language (eg. technical mathematical terms like "equation") or general language (eg. explaining mathematical ideas or procedures in non-technical terms). Do not count errors that are noticed and corrected within the segment.\\
2. Provide up to 5 bad examples (eg. missed opportunities or poor execution) of the teacher's imprecision in language or notation.\\

Format your answer as:\\
Good examples\\
1. Line number: <specify line number>, Segment: "<copied from transcript>", Reason: <specify why this is a good example>\\
2. ...\\

Bad examples\\
1. Line number: <specify line number>, Segment: "<copied from transcript>", Reason: <specify why this is a bad example>\\
2. ...\\

Good examples:
    \end{tcolorbox}
    \caption{Prompt for identifying highlights and missed opportunity on \mqiErrors. 
    \label{fig:example_langimp_prompt}}
\end{figure*}

\begin{figure*}[h]
    \centering 
    \begin{tcolorbox}[
    exampleprompt,
    title={\textbf{Prompt for identifying highlights and missed opportunity on \mqiStudentMath}},
    ]
    Consider the following classroom transcript.\\

Transcript:\\
\{transcript\}\\

Please do the following.\\
1. Provide up to 5 good examples of the students' mathematical questioning and reasoning. Student mathematical questioning and reasoning means that students engage in mathematical thinking. Examples include but are not limited to: Students provide counter-claims in response to a proposed mathematical statement or idea, ask mathematically motivated questions requesting explanations, make conjectures about the mathematics discussed in the lesson, etc.\\
2. Provide up to 5 bad examples (eg. missed opportunities or poor execution) of the students' mathematical questioning and reasoning.\\

Format your answer as:\\
Good examples\\
1. Line number: <specify line number>, Segment: "<copied from transcript>", Reason: <specify why this is a good example>\\
2. ...\\

Bad examples\\
1. Line number: <specify line number>, Segment: "<copied from transcript>", Reason: <specify why this is a bad example>\\
2. ...\\

Good examples:
    \end{tcolorbox}
    \caption{Prompt for identifying highlights and missed opportunity on \mqiStudentMath. 
    \label{fig:example_smqr_prompt}}
\end{figure*}

\subsection{Actionable suggestions}

We prompt ChatGPT to make actionable suggestions to the teacher for eliciting more student mathematical reasoning in the classroom.
The prompt used for this task is shown in Figure~\ref{fig:suggestion_prompt}.

\begin{figure*}[t]
    \centering 
    \begin{tcolorbox}[
    suggestionprompt,
    title={\textbf{Prompt for suggestions on eliciting more student reasoning in the classroom}},
    ]
    Consider the following classroom transcript. \\ \\
    Transcript: \\
    \{transcript\}\\\\
    The transcript contains many short student responses. Please provide 5 suggestions for the teacher on how the teacher could elicit more student reasoning in the classroom. Student reasoning is counted broadly as students asking questions, engaging in mathematical discourse with their teacher or peers, and providing explanations such as justifying their answers. \\ \\
    Format your answer as: \\
    Advice to the teacher: \\
    1. Line number: <specify line number>, Segment: "<copied from transcript>", Suggestion: <specify advice to the teacher> \\
    2. ...\\ \\
    Advice to the teacher:
    \end{tcolorbox}
    \caption{Prompt for suggestions on eliciting more student mathematical reasoning in the classroom.
    \label{fig:suggestion_prompt}}
\end{figure*}

\section{Human experiments \label{app:human_experiments}}

We recruited 2 experienced human teachers to evaluate the generated model responses. 
As illustrated in our main figure (Figure~\ref{fig:main_figure}), there are three main responses that are being evaluated by the human teachers: the highlights, missed opportunities and suggestions. 
Every observation code has their own generated highlights and missed opportunities. 

\subsection{Collecting model responses to evaluate}
\paragraph{Highlights and missed opportunities}

From the transcripts which have less than 10\% student contributions including ``[inaudible]'' markers, we sample \numCLASSSegments{} random 15-minutes transcript segments for the CLASS codes, and \numMQISegments{} random 7.5 minutes transcript segments for the MQI codes. 
Every code has 2 model-generated highlights and missed opportunities. 
In total, we have \textbf{\totalCLASSitems{} CLASS-annotated items}. The calculation is: \numCLASSSegments{} segments $\times 3$ CLASS  codes $\times$ ($2$ highlights $+2$ missed opportunities) = \totalCLASSitems{} items.
In total, we have \textbf{\totalMQIitems{} MQI-annotated items}. 
The calculation is: \numMQISegments{} segments $\times 4$ MQI  codes $\times$ ($2$ highlights $+2$ missed opportunities) = \totalMQIitems{} items.

\paragraph{Suggestions}
We use the same \numSuggestionSegments{} random MQI 7.5-minutes transcript segments for prompting the model for suggestions. 
In total, we have \textbf{\totalSuggestions{} item suggestions}. The calculation is \numSuggestionSegments{} segments $\times 2 $ suggestions = \totalSuggestions{} items. 

\subsection{Evaluation axes and human interface}
This section details what we ask the teachers to evaluate qualitatively. 
Some of the details are repeated from Section~\ref{sec:validation_methods} for completeness. 
We additionally include screenshots of the human experiment interface.

\paragraph{Highlights and missed opportunities}
The teachers evaluate the model examples along three axes. 
One is \textbf{relevance}: Is the model's response relevant to the CLASS or MQI dimension of interest? 
Two is \textbf{faithfulness}: Does the model's response have the right interpretation of the events that occur in the classroom transcript? We evaluate along this dimension because the model sometimes can hallucinate or misinterpret the events in the transcript when providing examples.
Three is \textbf{insightfulness}: Does the model's response reveal something beyond the line segment's obvious meaning in the transcript?
We ask the teachers to evaluate on a 3-point scale (yes, somewhat, no).
Optionally, the teacher may additionally provide a free text comment, if they want to elaborate their answer. 

\begin{figure*}[h]
    \newcommand{\ratio}{0.30}
    \centering
    \begin{subfigure}{\linewidth}
        \includegraphics[width=\linewidth]{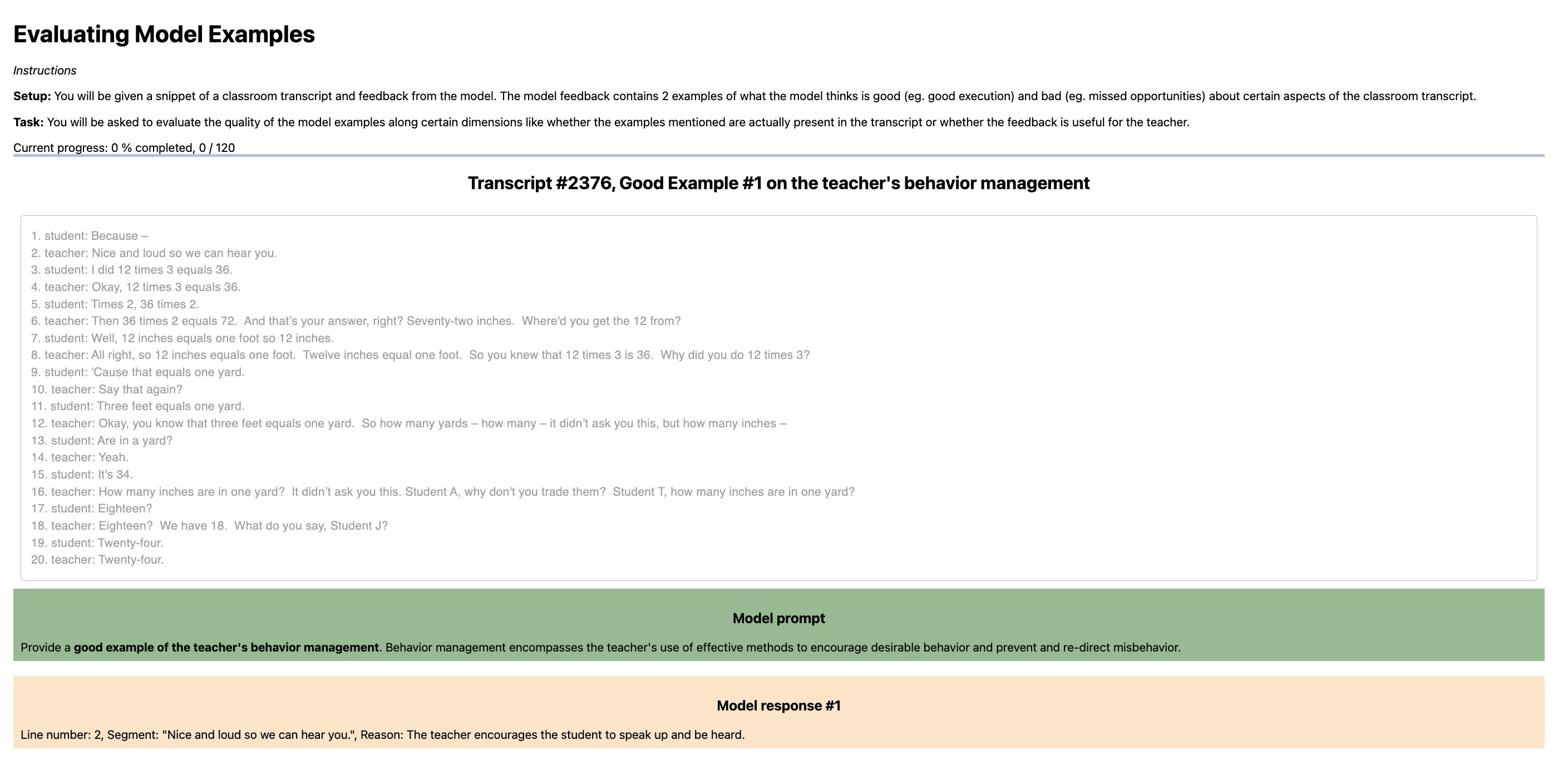}
    \end{subfigure}
    \begin{subfigure}{\linewidth}
        \includegraphics[width=\linewidth]{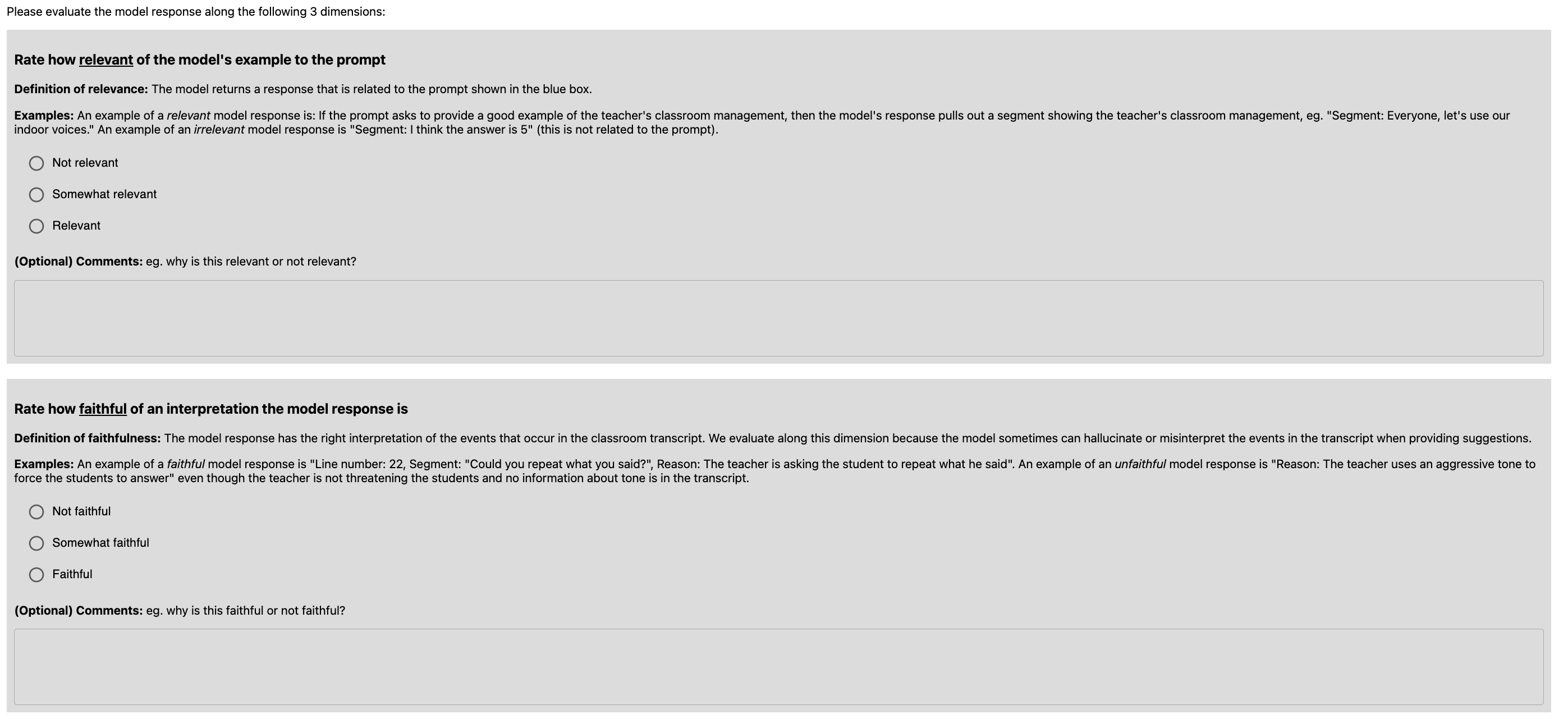}
    \end{subfigure}
    \begin{subfigure}{\linewidth}
        \includegraphics[width=\linewidth]{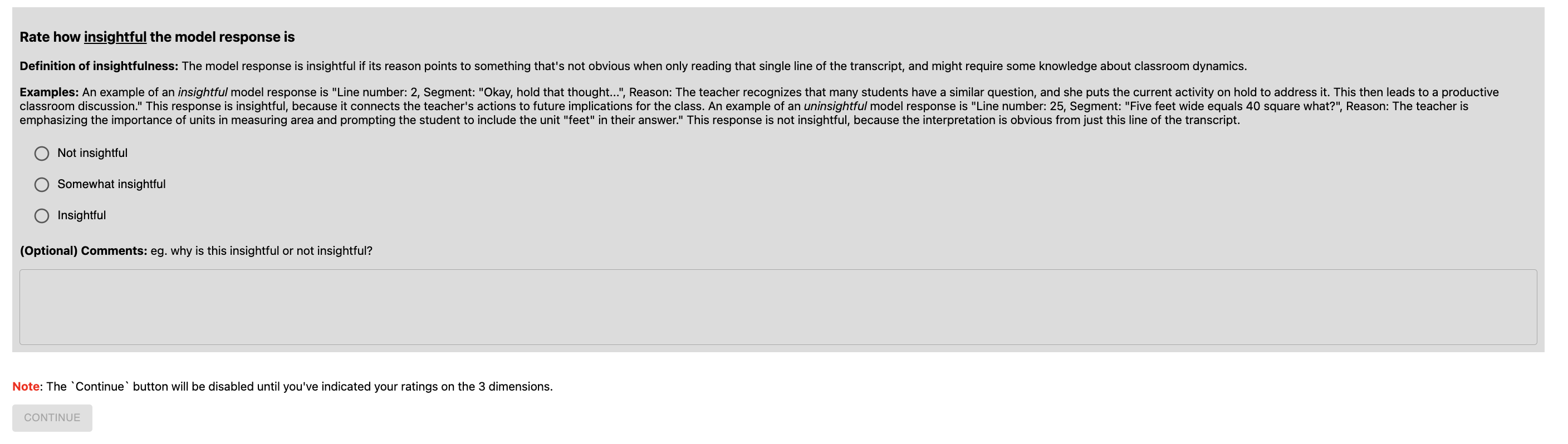}
    \end{subfigure}
    \caption{
    Human interface for evaluating the highlights (good examples) and missed opportunities (bad examples) on CLASS observation items generated by the model.
    \label{fig:class_example_interface}}
\end{figure*}

\begin{figure*}[h]
    \newcommand{\ratio}{0.30}
    \centering
    \begin{subfigure}{\linewidth}
        \includegraphics[width=\linewidth]{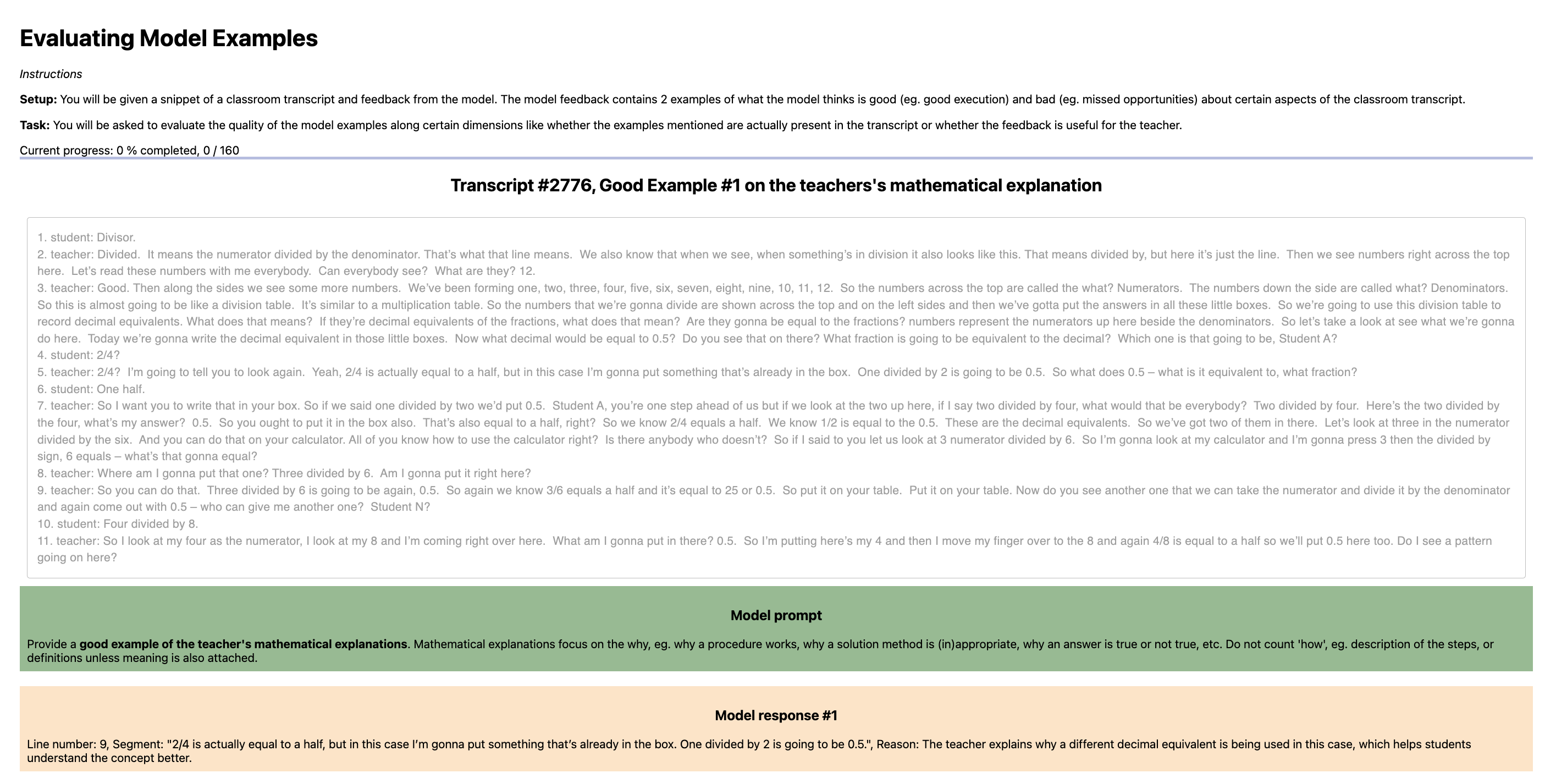}
    \end{subfigure}
    \begin{subfigure}{\linewidth}
        \includegraphics[width=\linewidth]{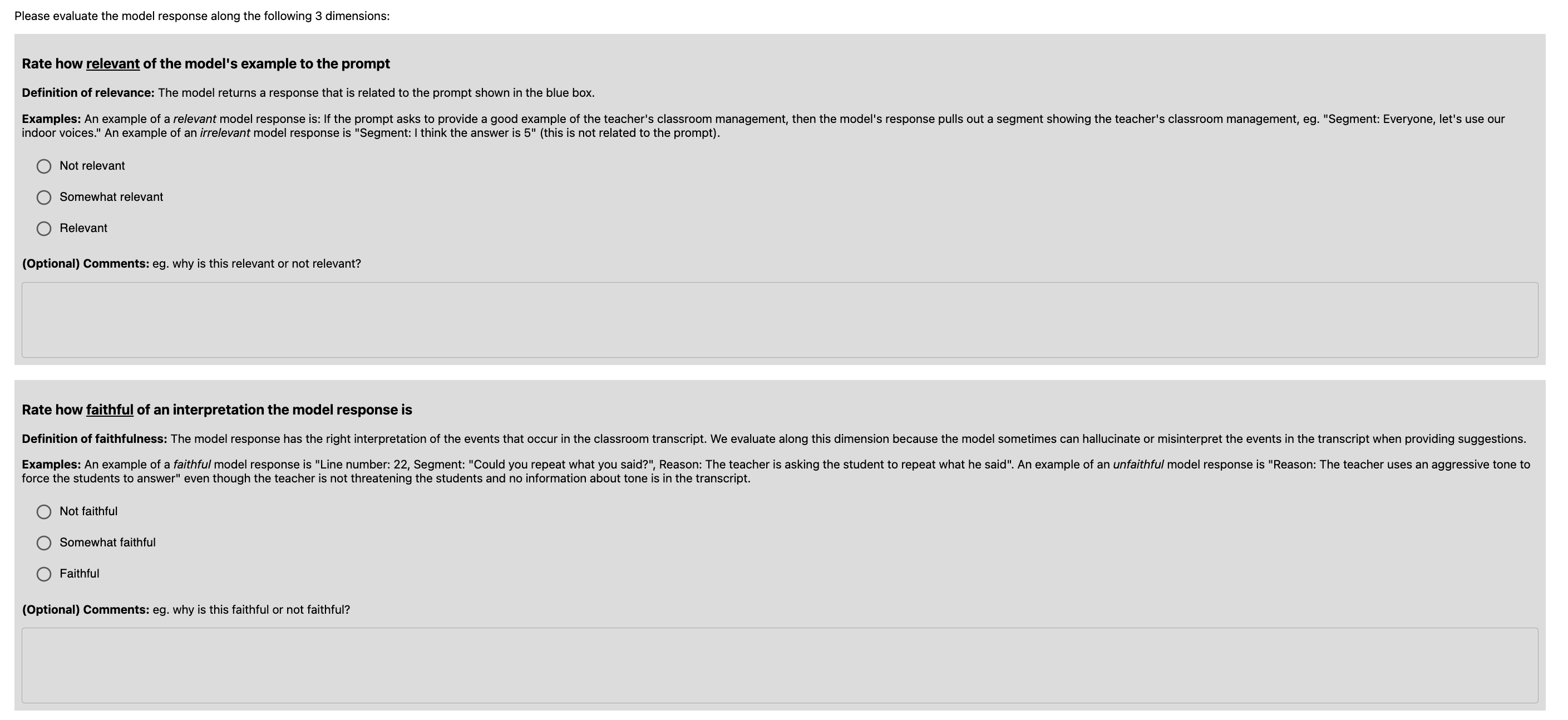}
    \end{subfigure}
    \begin{subfigure}{\linewidth}
        \includegraphics[width=\linewidth]{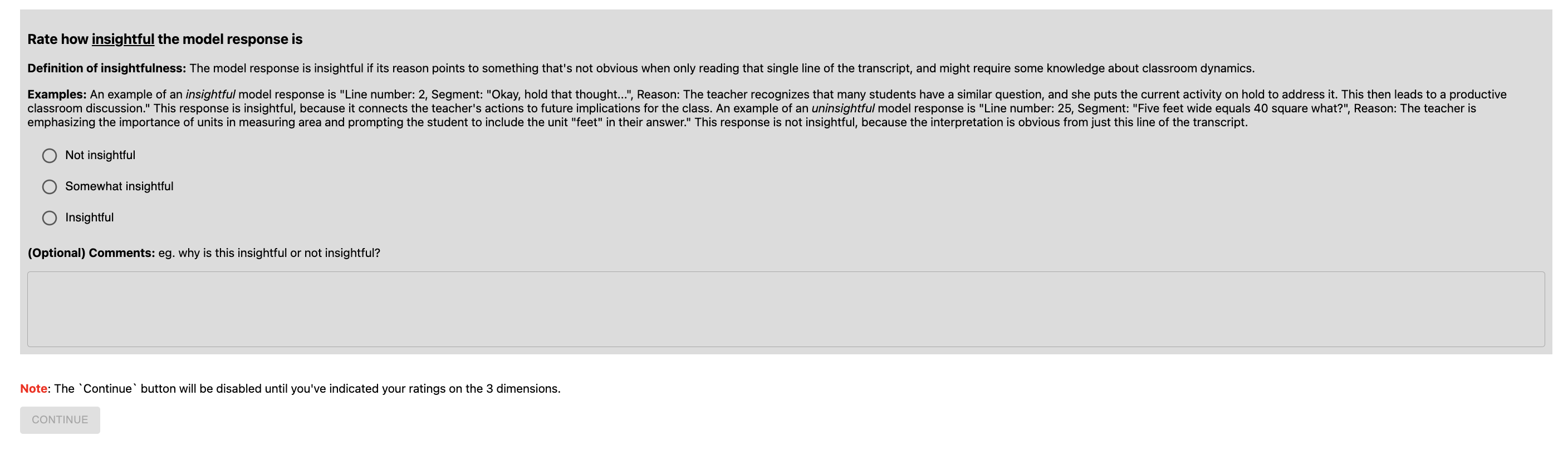}
    \end{subfigure}
    \caption{
    Human interface for evaluating the highlights (good examples) and missed opportunities (bad examples) on MQI observation items generated by the model.
    \label{fig:mqi_example_interface}}
\end{figure*}

Figure~\ref{fig:class_example_interface} shows the human interface for evaluating the CLASS observation items, and Figure~\ref{fig:mqi_example_interface} for evaluating the MQI observation items.

\paragraph{Suggestions}
The teachers evaluate the model suggestions along four axes. 
One is \textbf{relevance}: Is the model's response relevant to eliciting more student mathematical reasoning in the classroom?
Two is \textbf{faithfulness}: Does the model's response have the right interpretation of the events that occur in the classroom transcript?
Similar to the previous research question, we evaluate along this dimension because the model sometimes can hallucinate or misinterpret the events in the transcript when providing suggestions.
Three is \textbf{actionability}: Is the model's suggestion something that the teacher can easily translate into practice for improving their teaching or encouraging student mathematical reasoning?
Finally, four is \textbf{novelty}: Is the model suggestion something that the teacher already does in the transcript?
Similar to the previous section, we ask the teachers to evaluate on a 3-point scale (yes, somewhat, no). 

\begin{figure*}[h]
    \newcommand{\ratio}{0.30}
    \centering
    \begin{subfigure}{\linewidth}
        \includegraphics[width=\linewidth]{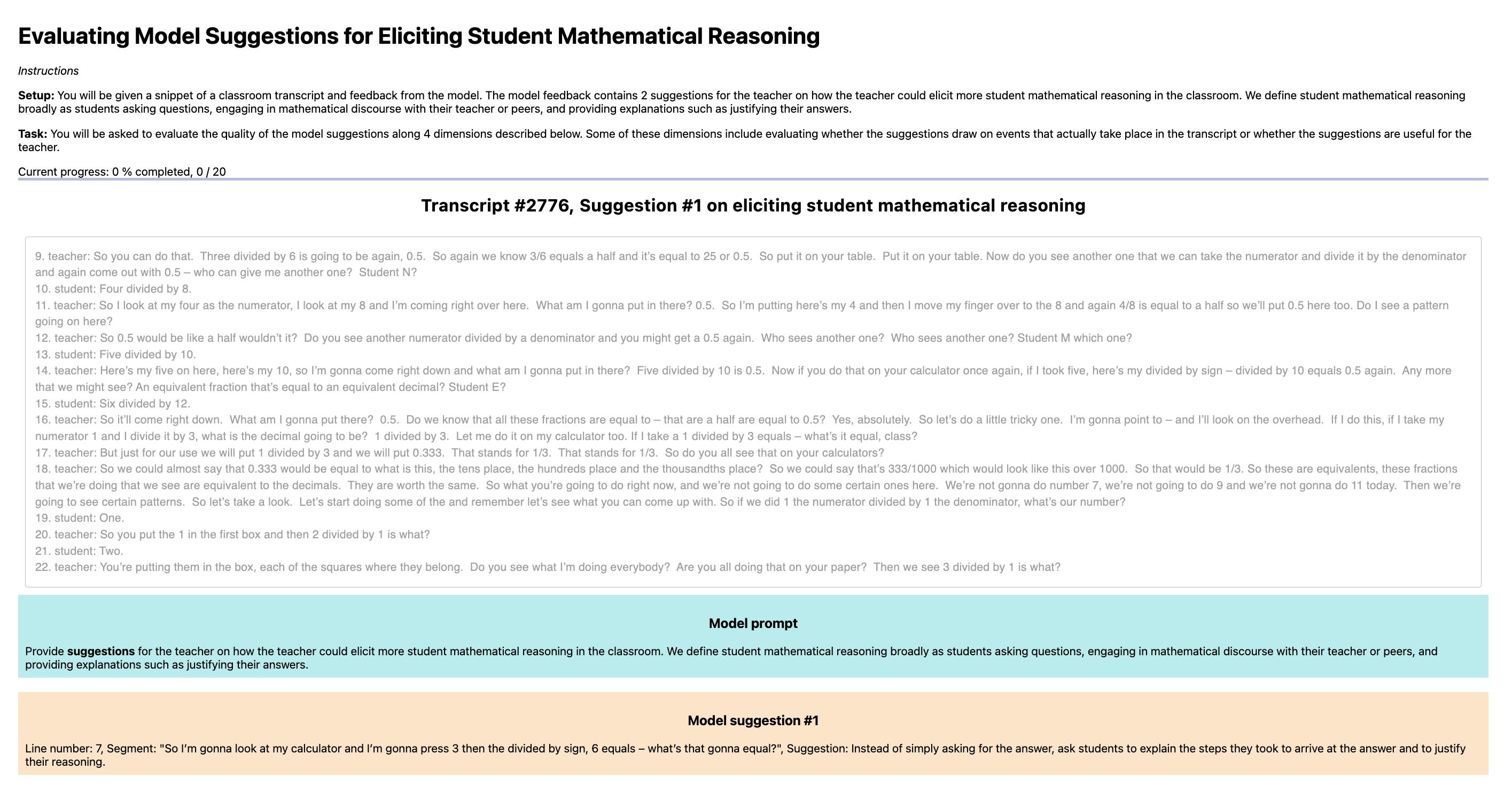}
    \end{subfigure}
    \begin{subfigure}{\linewidth}
        \includegraphics[width=\linewidth]{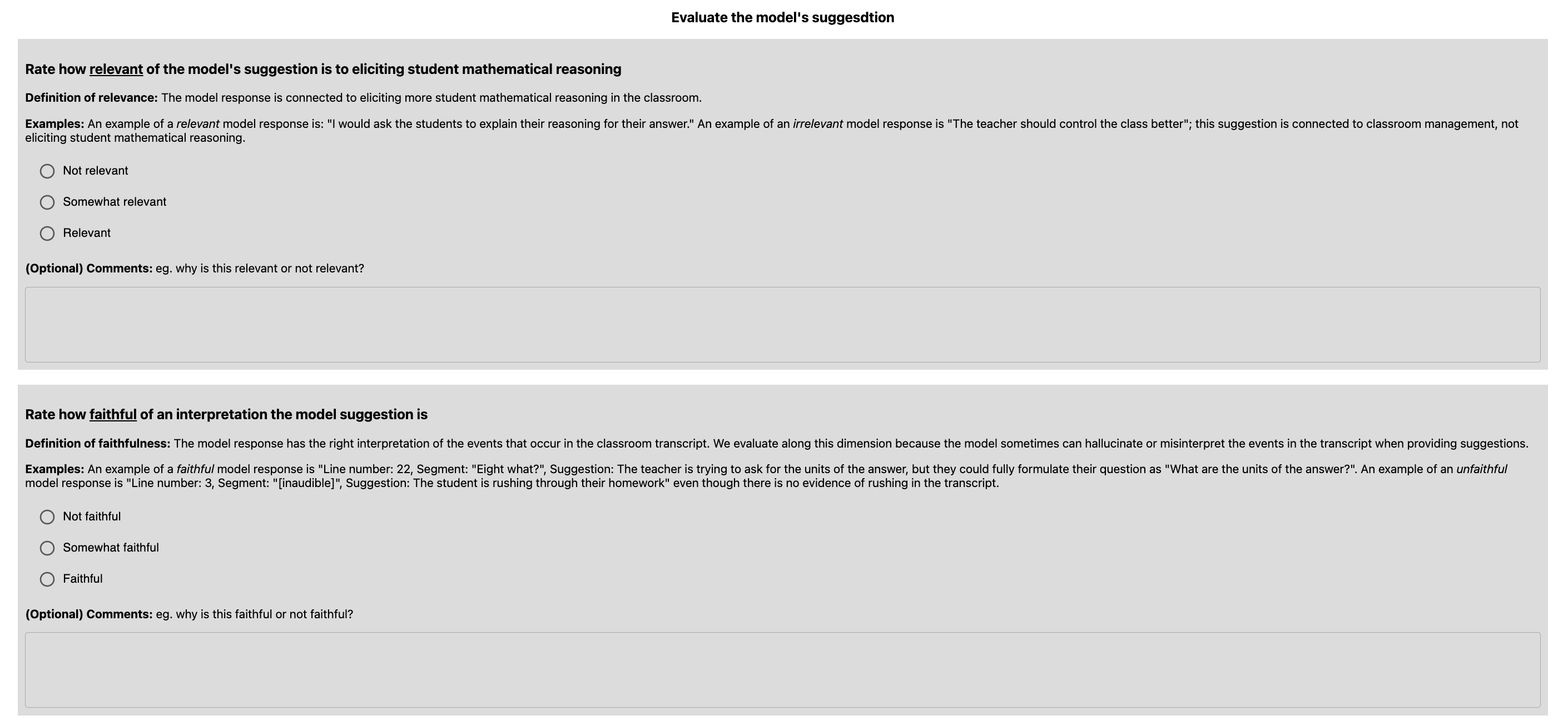}
    \end{subfigure}
    \begin{subfigure}{\linewidth}
        \includegraphics[width=\linewidth]{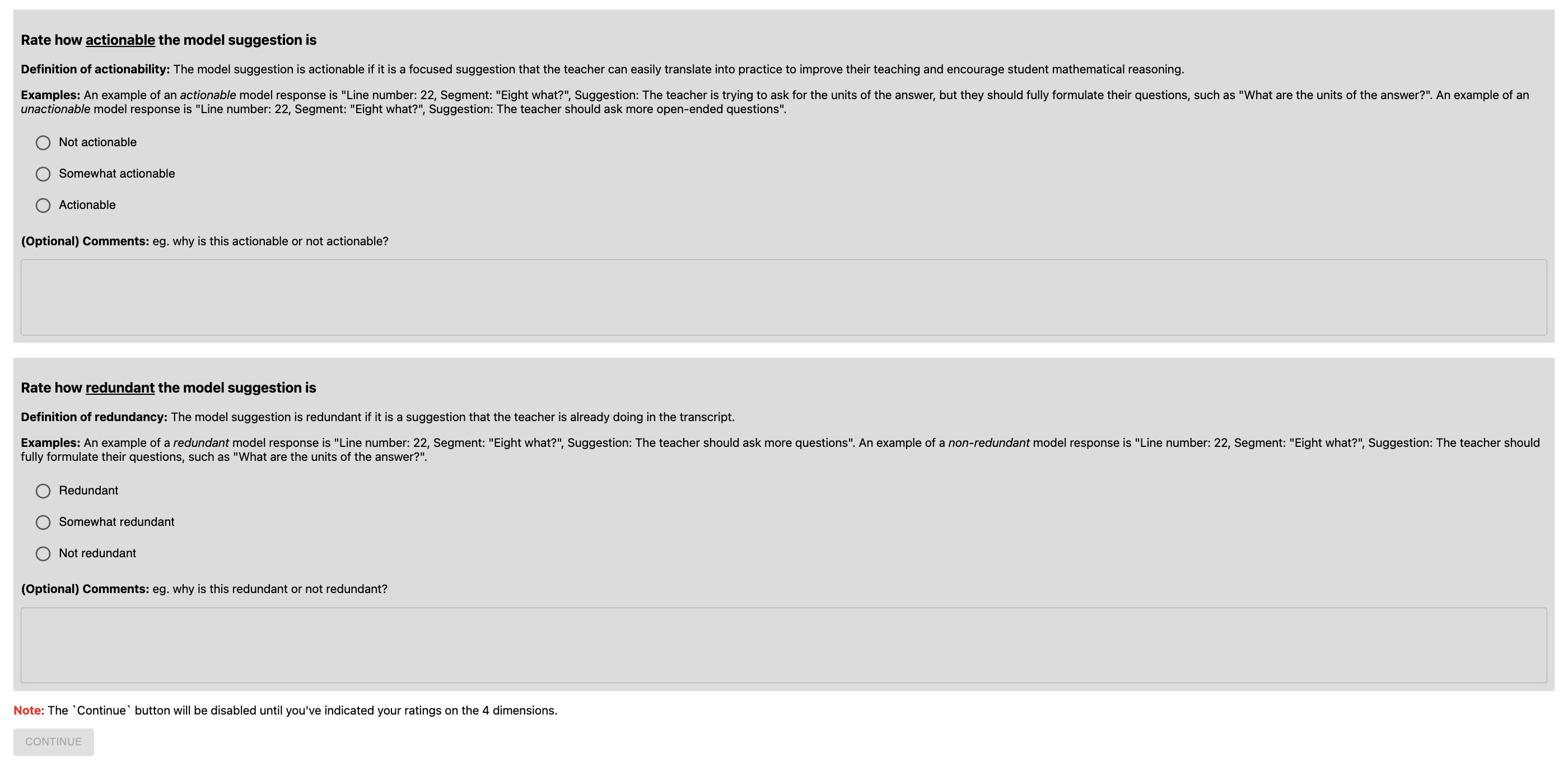}
    \end{subfigure}
    \caption{
    Human interface for evaluating the model suggestions.
    \label{fig:suggestions_interface}}
\end{figure*}

Figure~\ref{fig:suggestions_interface} shows the human interface for evaluating the model suggestions.

\section{Additional results on quantitative scoring \label{app:additional_results}}

We include the additional results on the the quantitative scoring task. 

\begin{figure*}[h]
    \newcommand{\ratio}{0.30}
    \centering
    \begin{tabular}{cccc}
        \multicolumn{1}{c}{} & \multicolumn{3}{c}{} \\ 
        & \classPositiveClimate & \classBehavioralManagement & \classInstructionalDialogue \\ 
            \multirow{-10}{*}{\rotatebox[origin=c]{90}{\directAnswer}} & 
            \begin{subfigure}{\ratio\linewidth}
                \includegraphics[width=\linewidth]{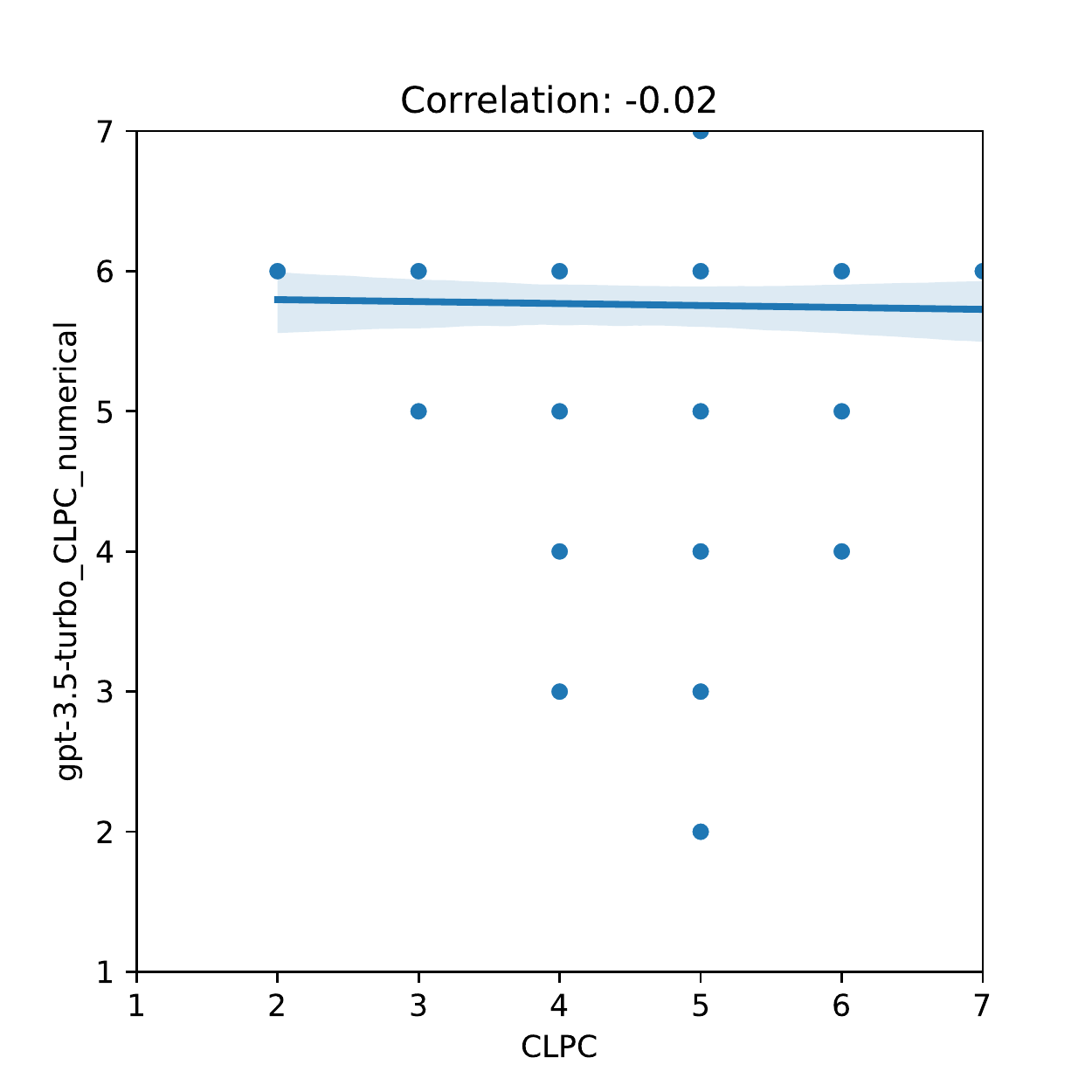}
              \end{subfigure}
            & \begin{subfigure}{\ratio\linewidth}
                \includegraphics[width=\linewidth]{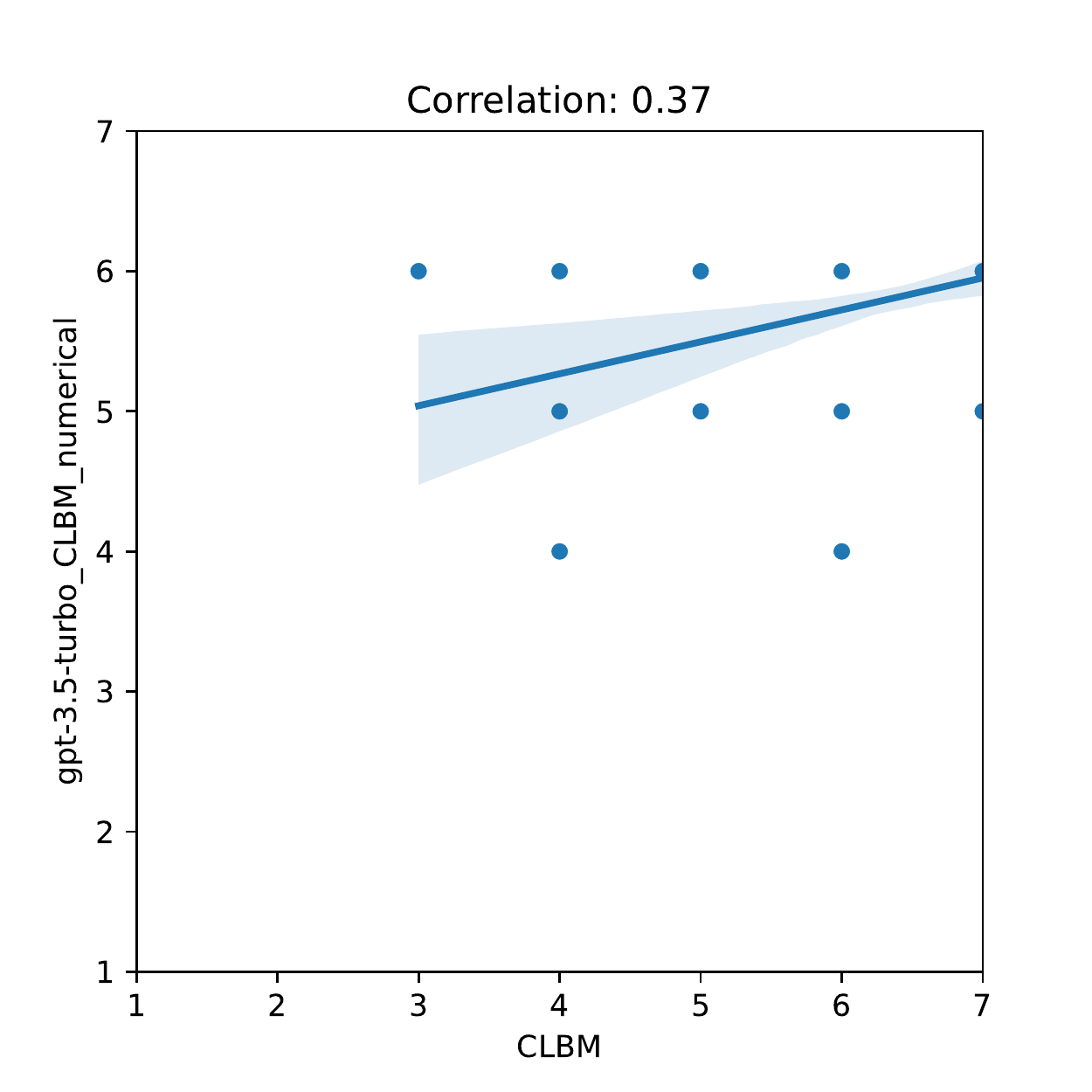}
              \end{subfigure}
            & \begin{subfigure}{\ratio\linewidth}
                \includegraphics[width=\linewidth]{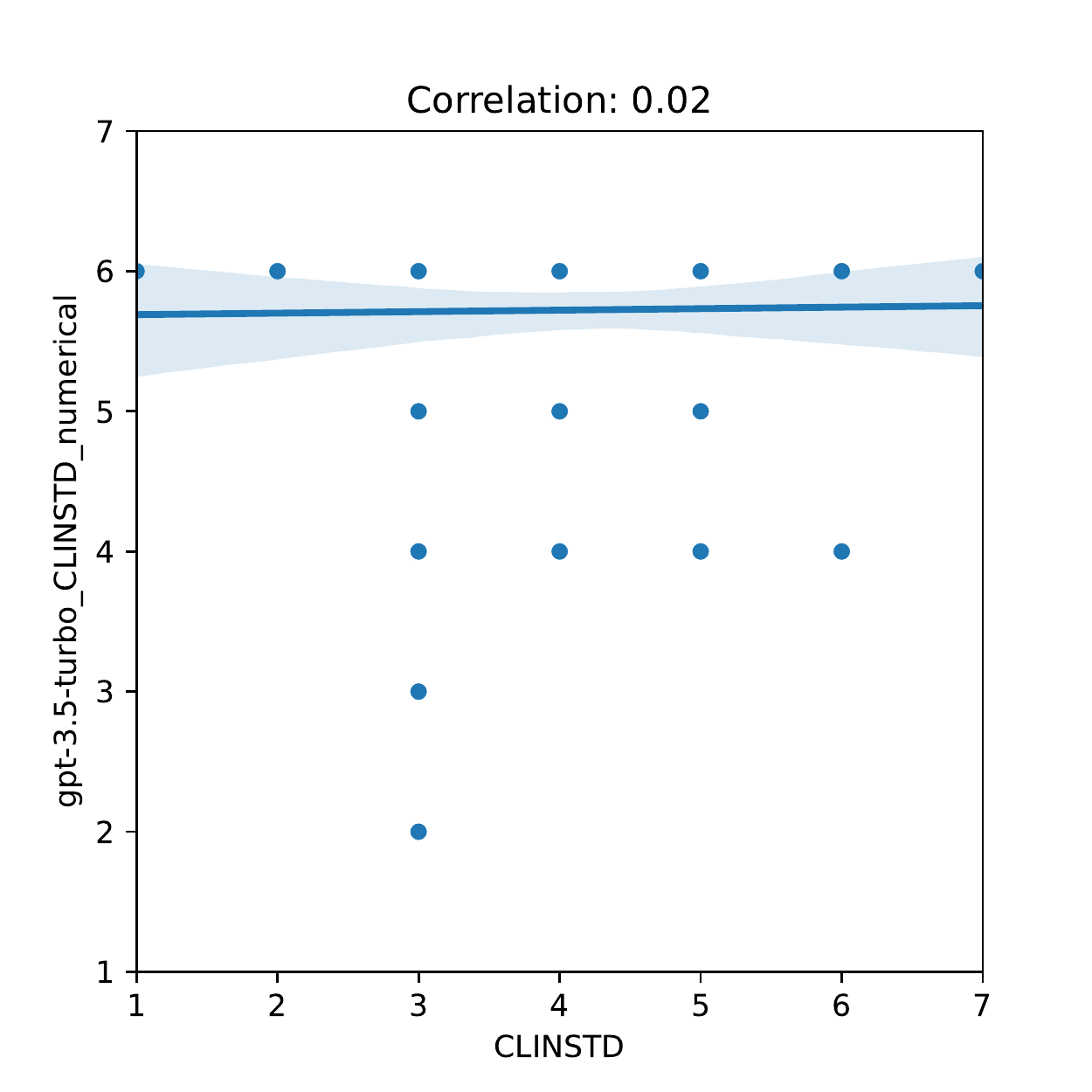}
              \end{subfigure} \\
            \multirow{-10}{*}{\rotatebox[origin=c]{90}{\directAnswerDescription}} & \begin{subfigure}{\ratio\linewidth}
                \includegraphics[width=\linewidth]{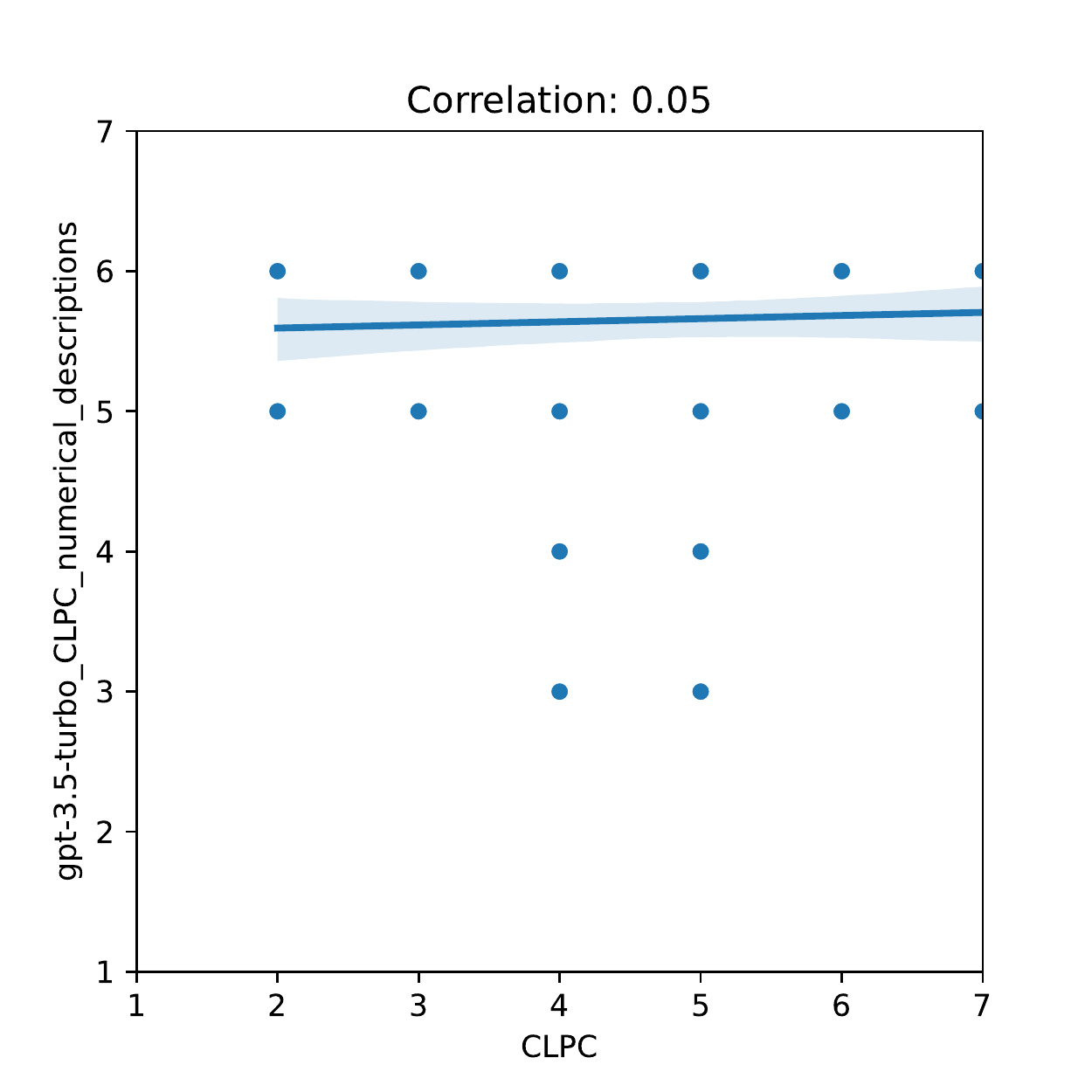}
              \end{subfigure}
            & \begin{subfigure}{\ratio\linewidth}
                \includegraphics[width=\linewidth]{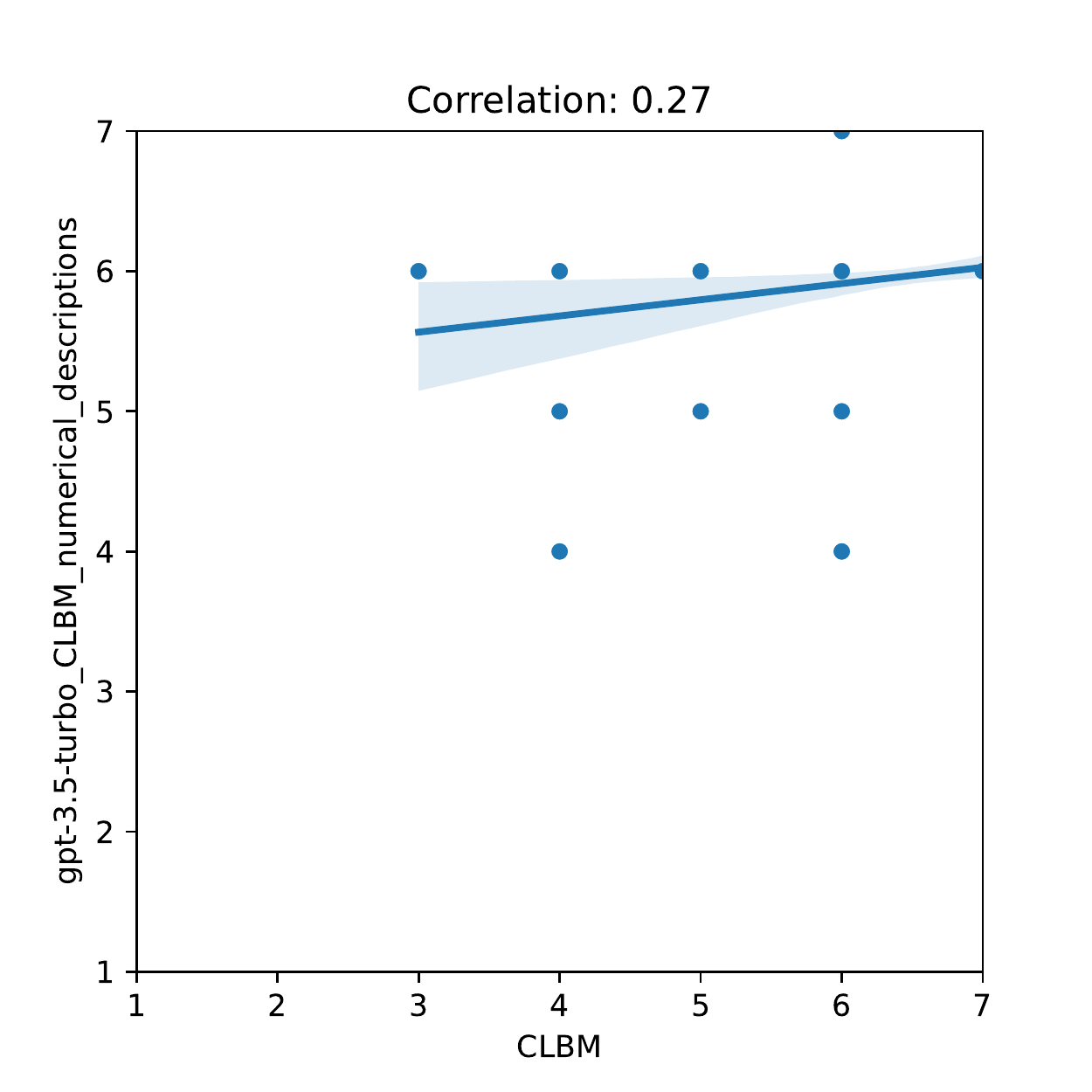}
              \end{subfigure}
            & \begin{subfigure}{\ratio\linewidth}
                \includegraphics[width=\linewidth]{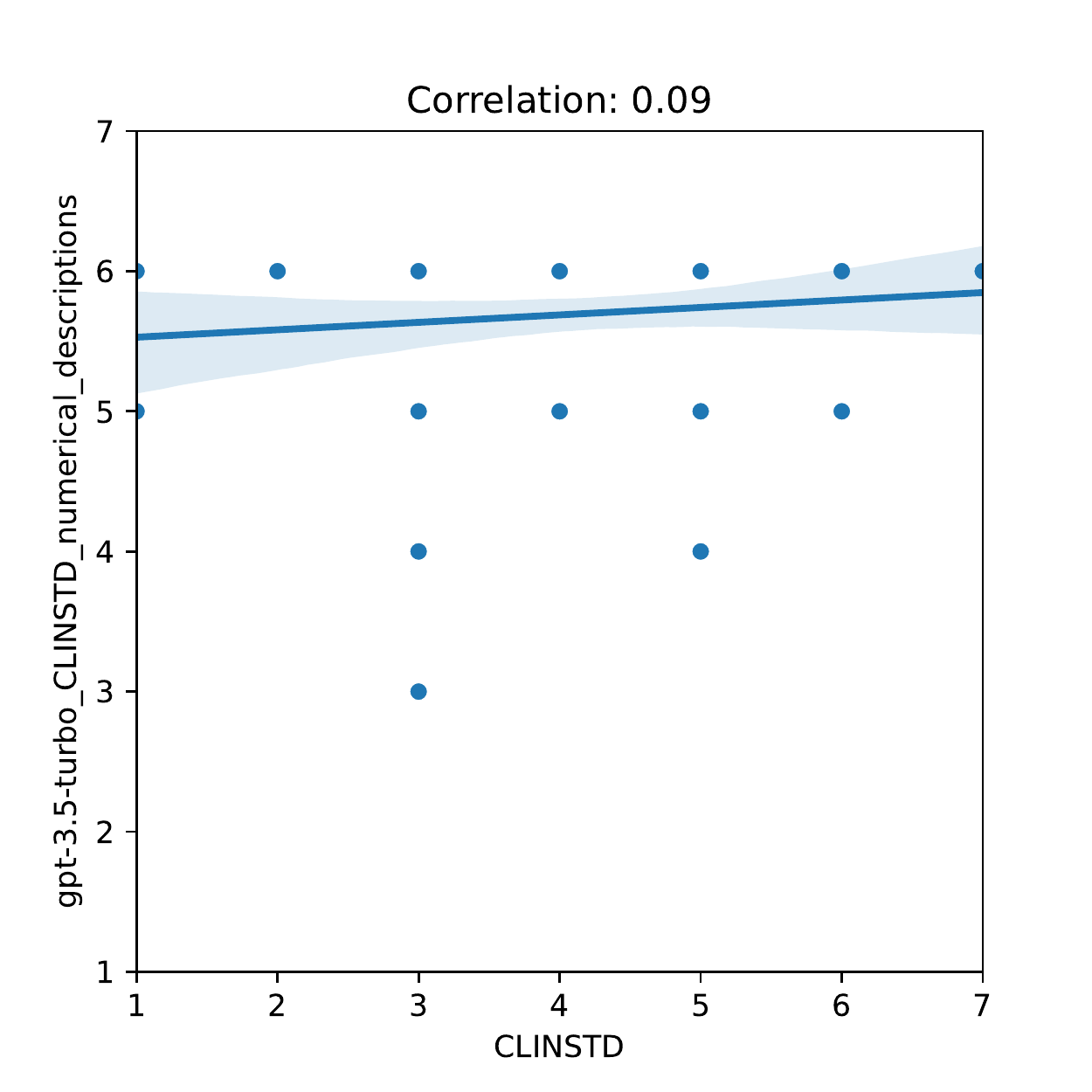}
              \end{subfigure} \\
            \multirow{-10}{*}{\rotatebox[origin=c]{90}{\reasoningAnswer}}& \begin{subfigure}{\ratio\linewidth}
                \includegraphics[width=\linewidth]{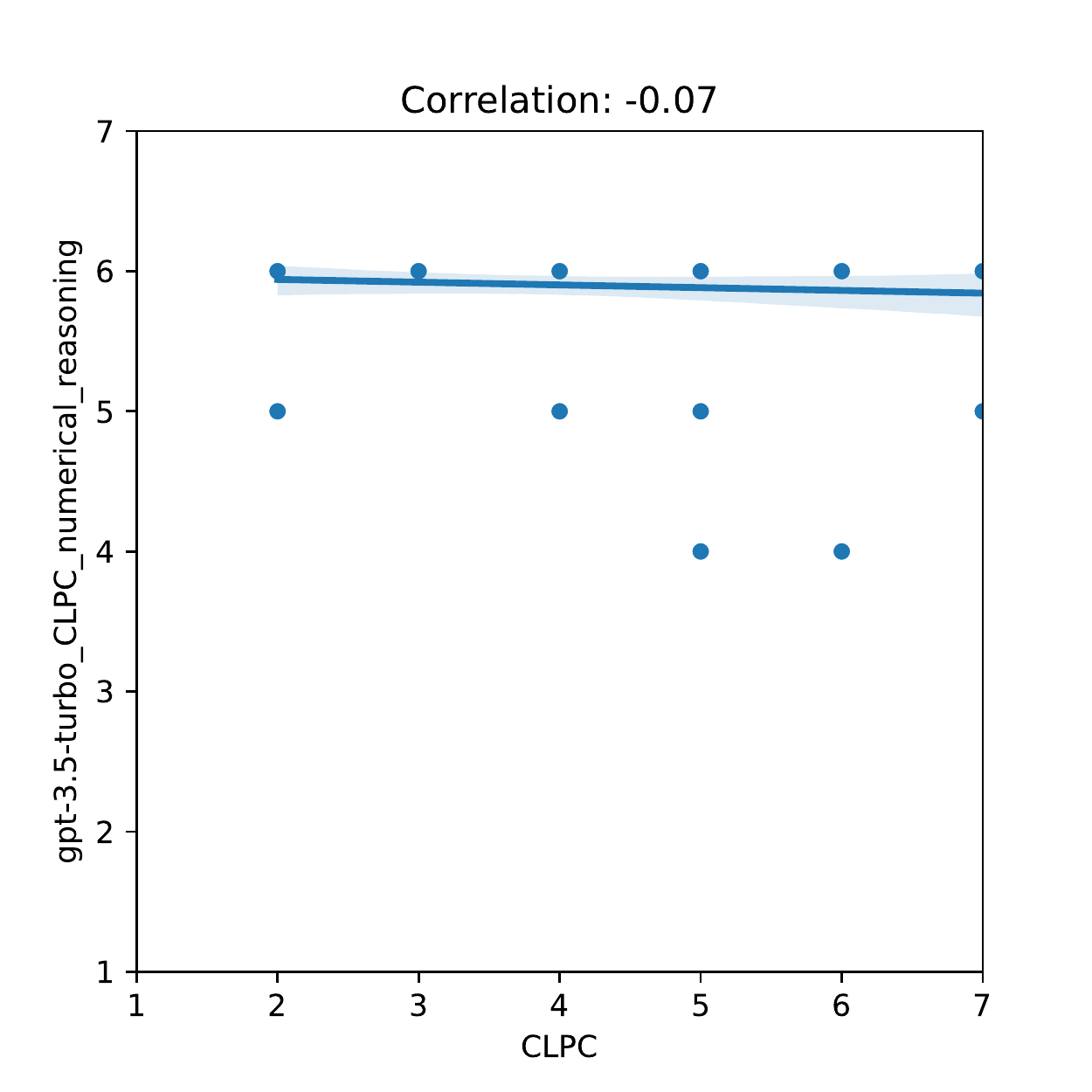}
              \end{subfigure}
            & \begin{subfigure}{\ratio\linewidth}
                \includegraphics[width=\linewidth]{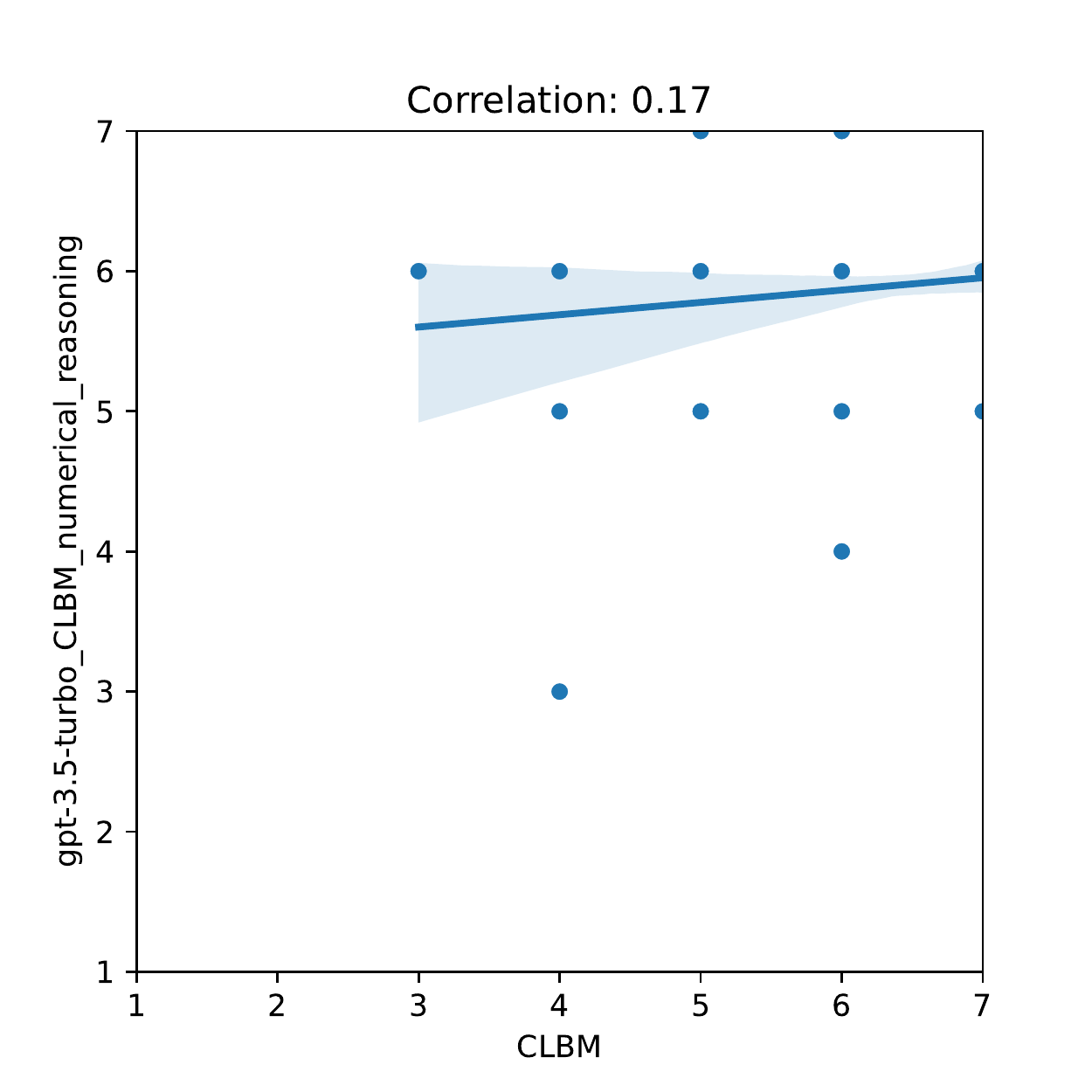}
              \end{subfigure}
            & \begin{subfigure}{\ratio\linewidth}
                \includegraphics[width=\linewidth]{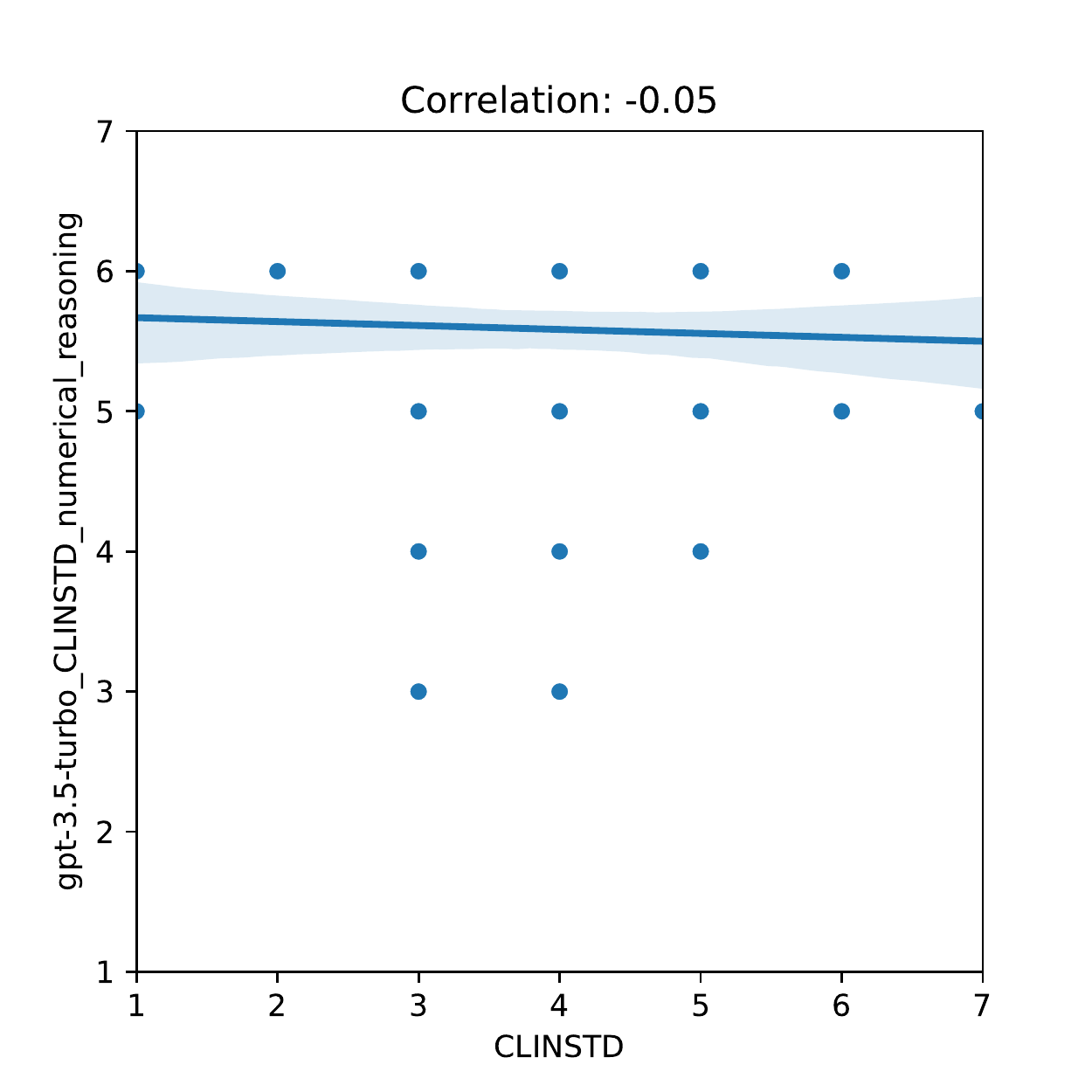}
              \end{subfigure} \\
        \end{tabular}
        \caption{Correlation between CLASS annotations and model predictions.}
        \label{fig:class_prediction_correlation}
    \end{figure*}

\begin{figure*}[h]
    \newcommand{\ratio}{0.30}
    \newcommand{\offset}{-7}
    \centering
    \begin{tabular}{cccc}
        \multicolumn{1}{c}{} & \multicolumn{3}{c}{} \\ 
        & \classPositiveClimate & \classBehavioralManagement & \classInstructionalDialogue \\
        \multirow{0}{*}{\rotatebox{90}{}} 
            \multirow{\offset}{*}{\rotatebox[origin=c]{90}{\directAnswer}} & \begin{subfigure}{\ratio\linewidth}
                \includegraphics[width=\linewidth]{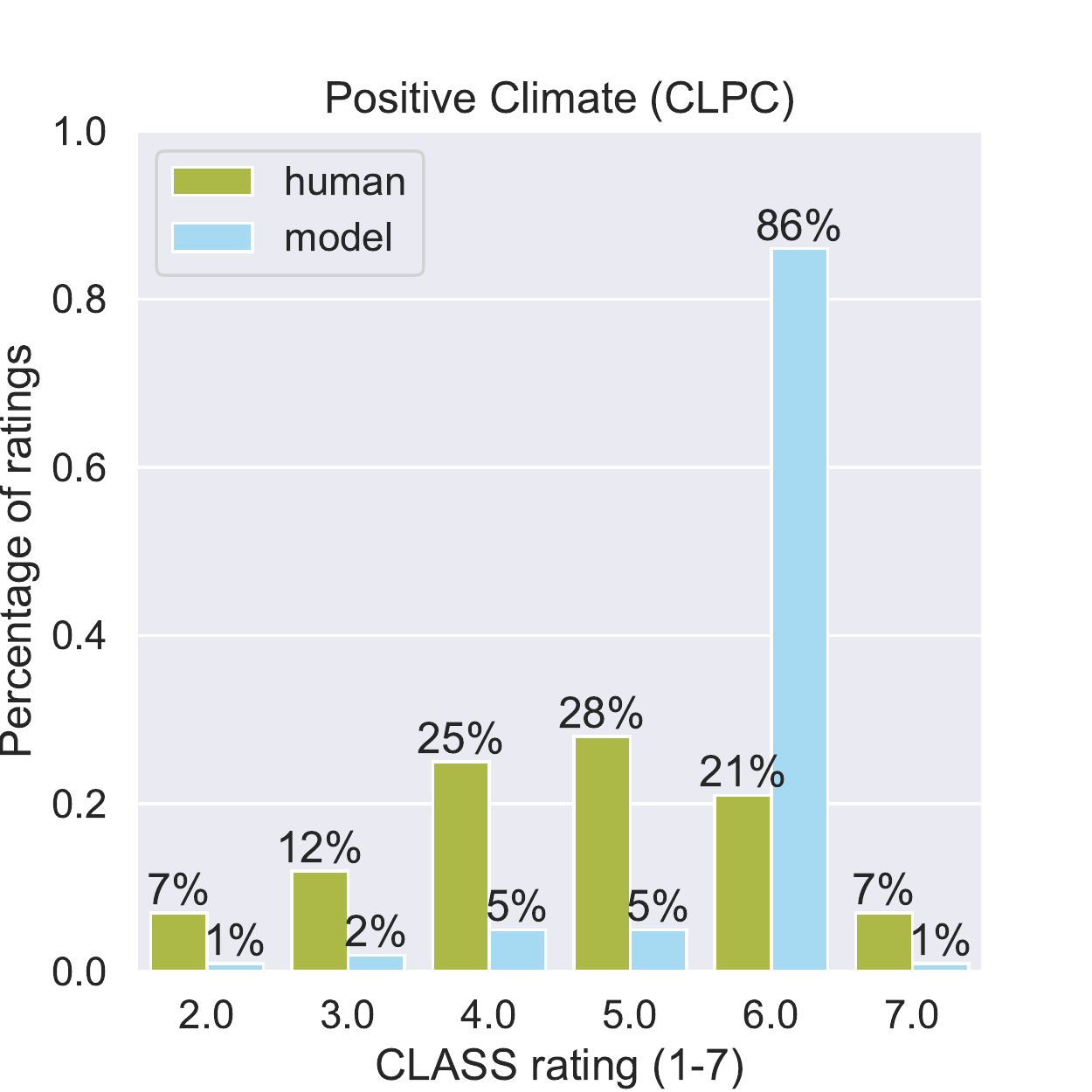}
              \end{subfigure}
            & \begin{subfigure}{\ratio\linewidth}
                \includegraphics[width=\linewidth]{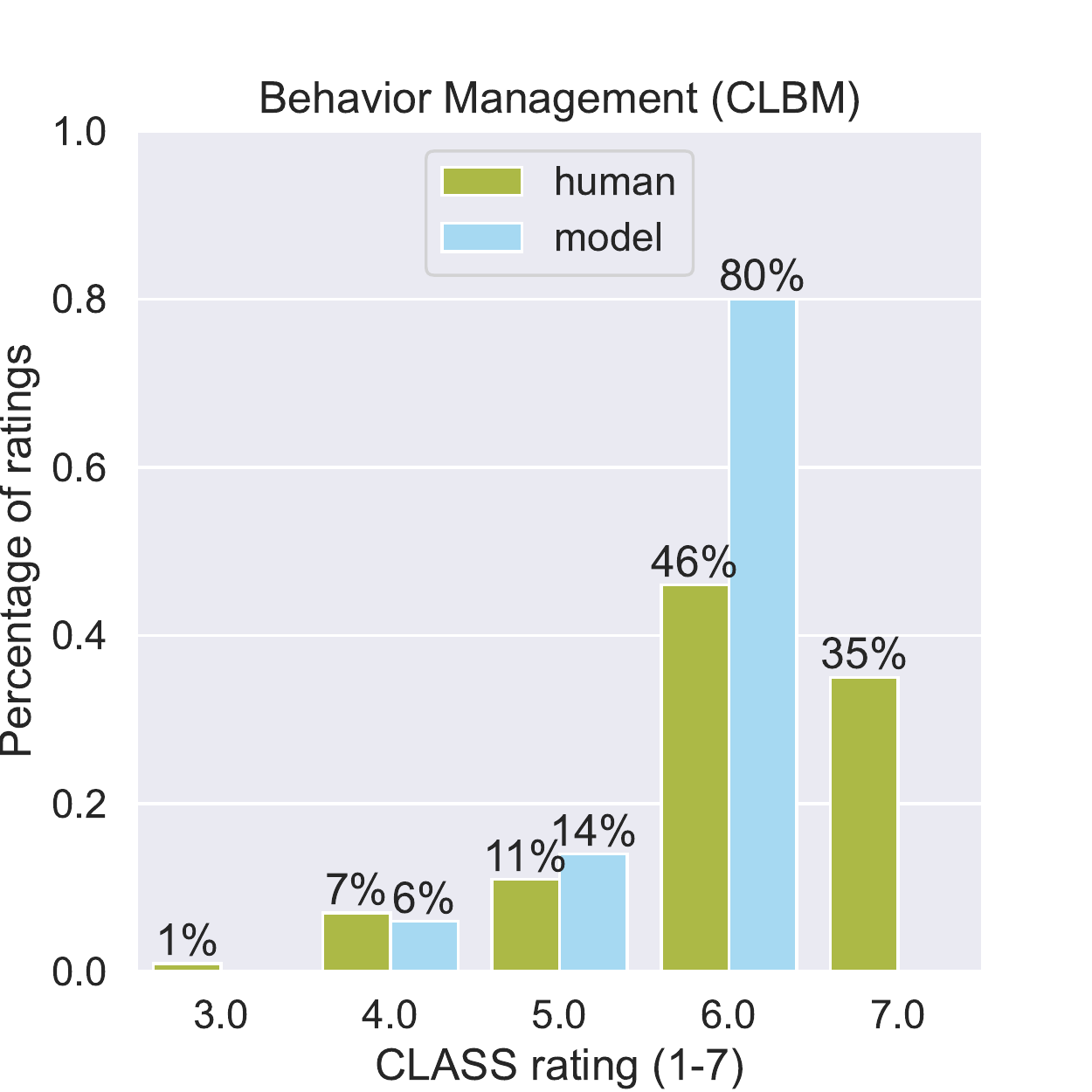}
              \end{subfigure}
            & \begin{subfigure}{\ratio\linewidth}
                \includegraphics[width=\linewidth]{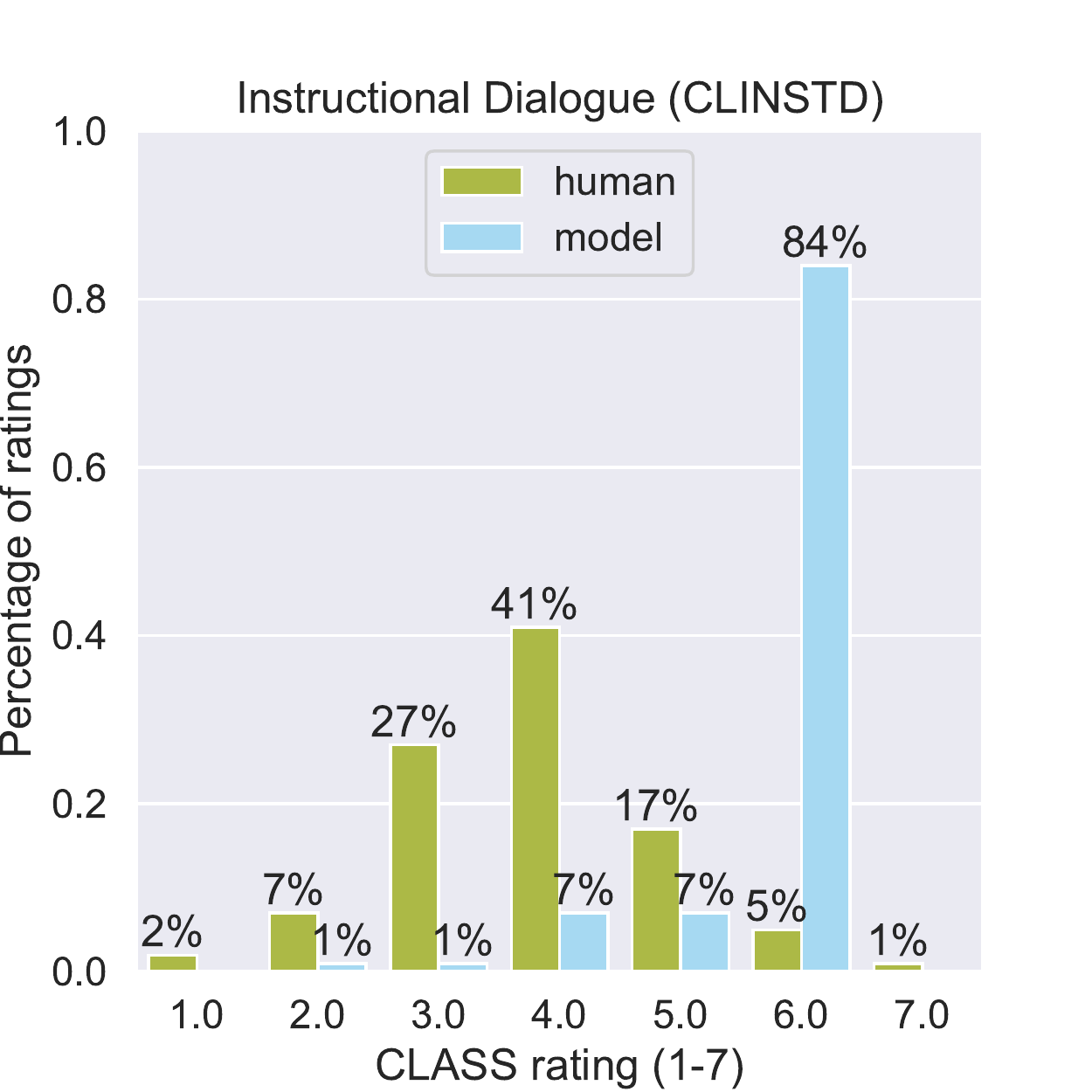}
              \end{subfigure} \\
            \multirow{\offset}{*}{\rotatebox[origin=c]{90}{\directAnswerDescription}} & \begin{subfigure}{\ratio\linewidth}
                \includegraphics[width=\linewidth]{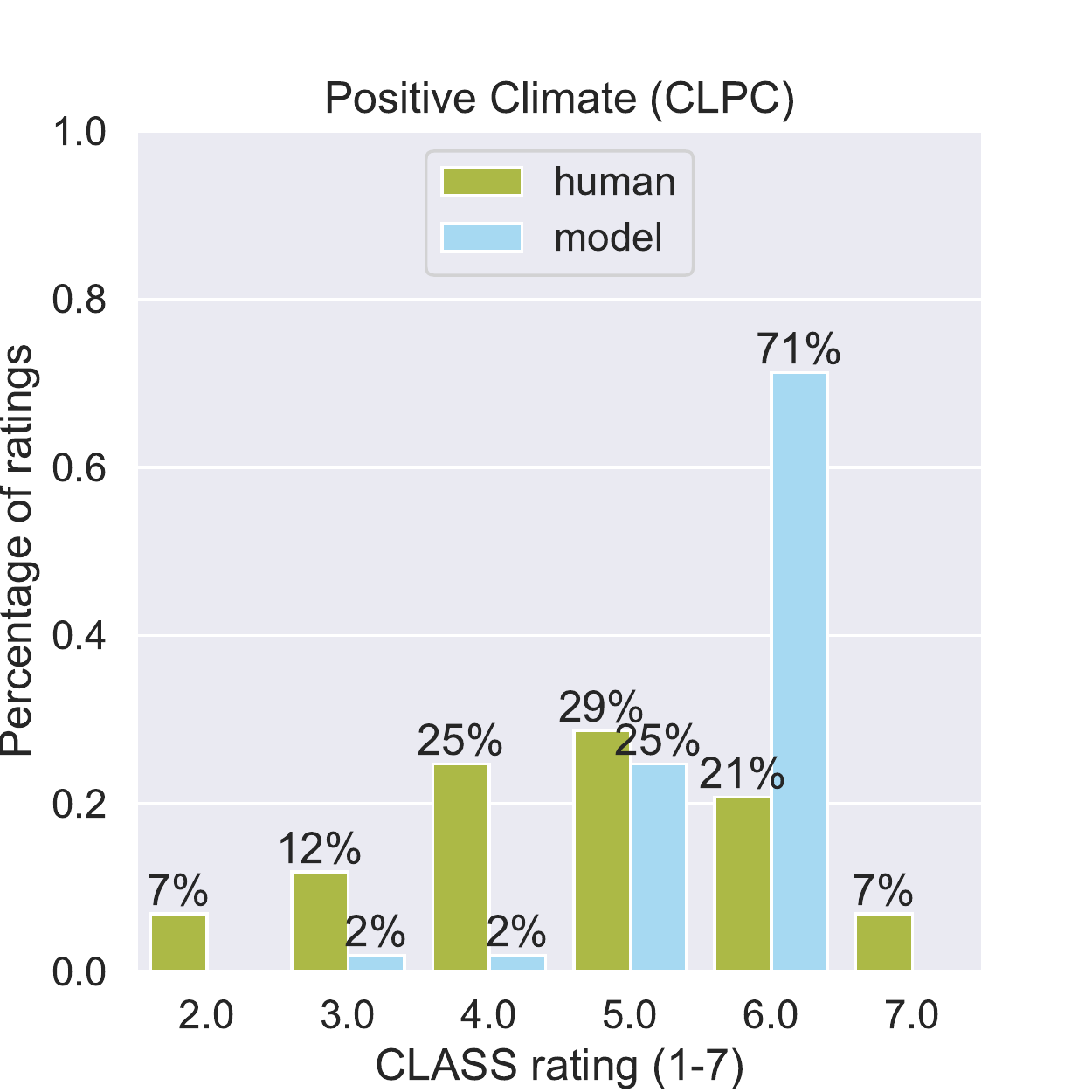}
              \end{subfigure}
            & \begin{subfigure}{\ratio\linewidth}
                \includegraphics[width=\linewidth]{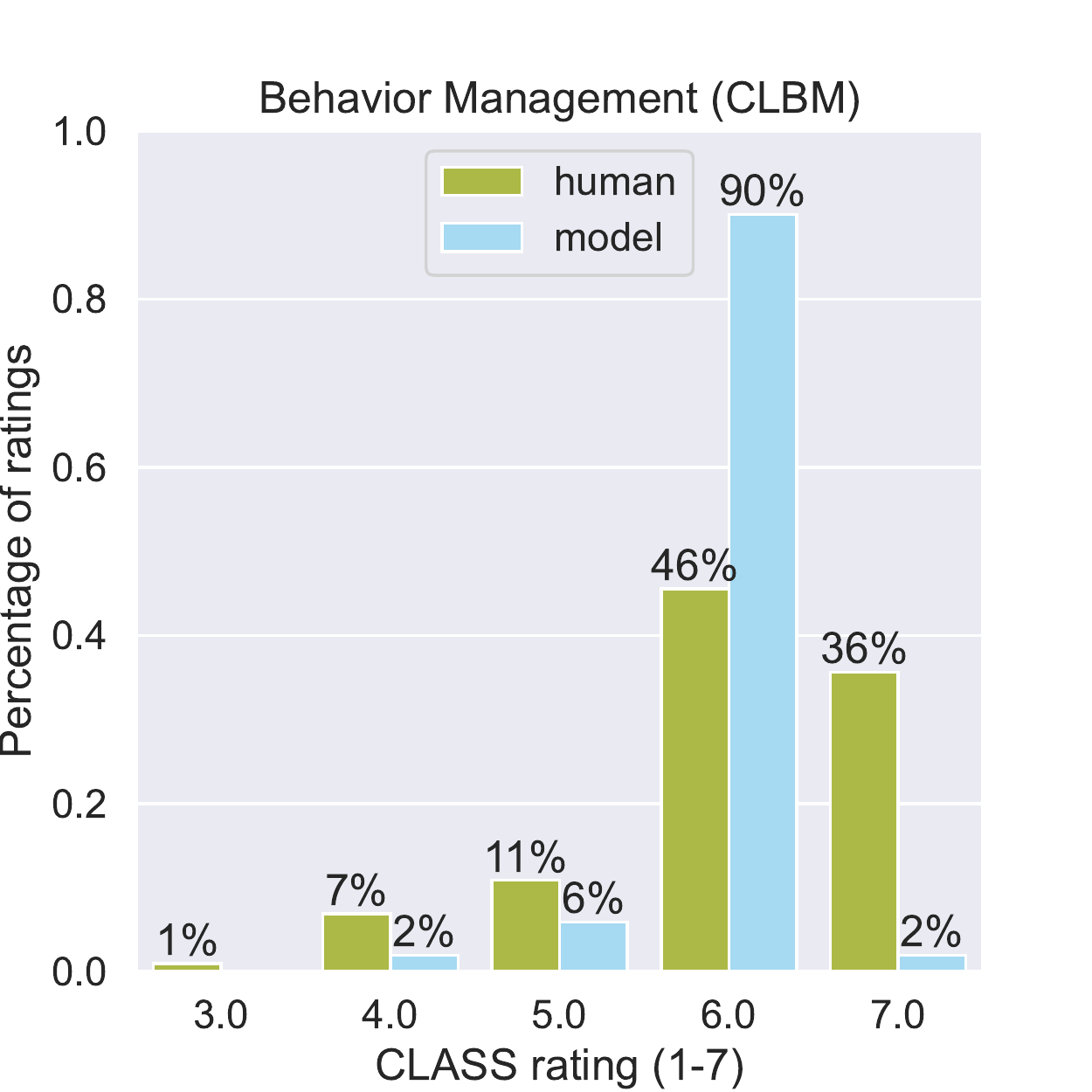}
              \end{subfigure}
            & \begin{subfigure}{\ratio\linewidth}
                \includegraphics[width=\linewidth]{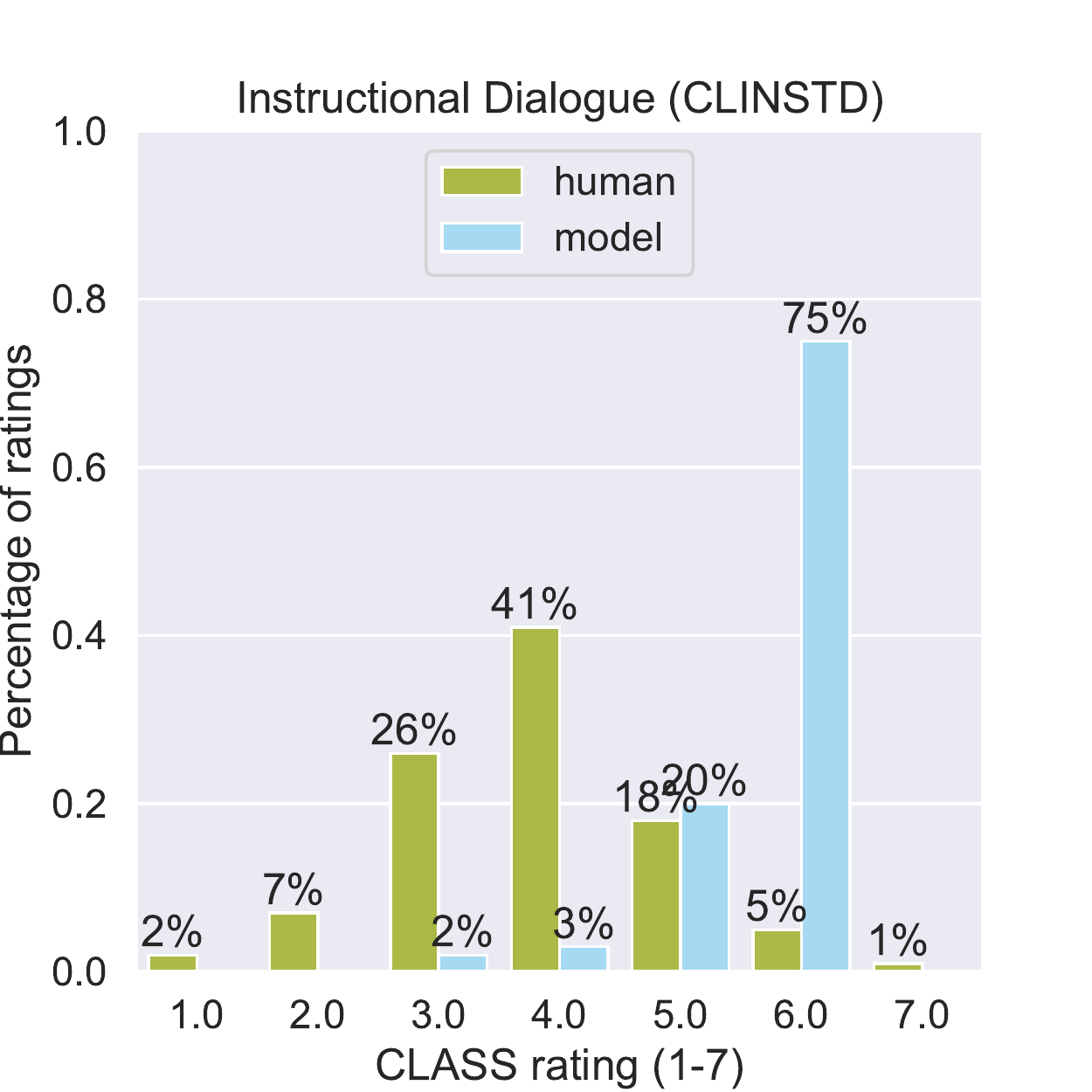}
              \end{subfigure} \\
            \multirow{\offset}{*}{\rotatebox[origin=c]{90}{\reasoningAnswer}}& \begin{subfigure}{\ratio\linewidth}
                \includegraphics[width=\linewidth]{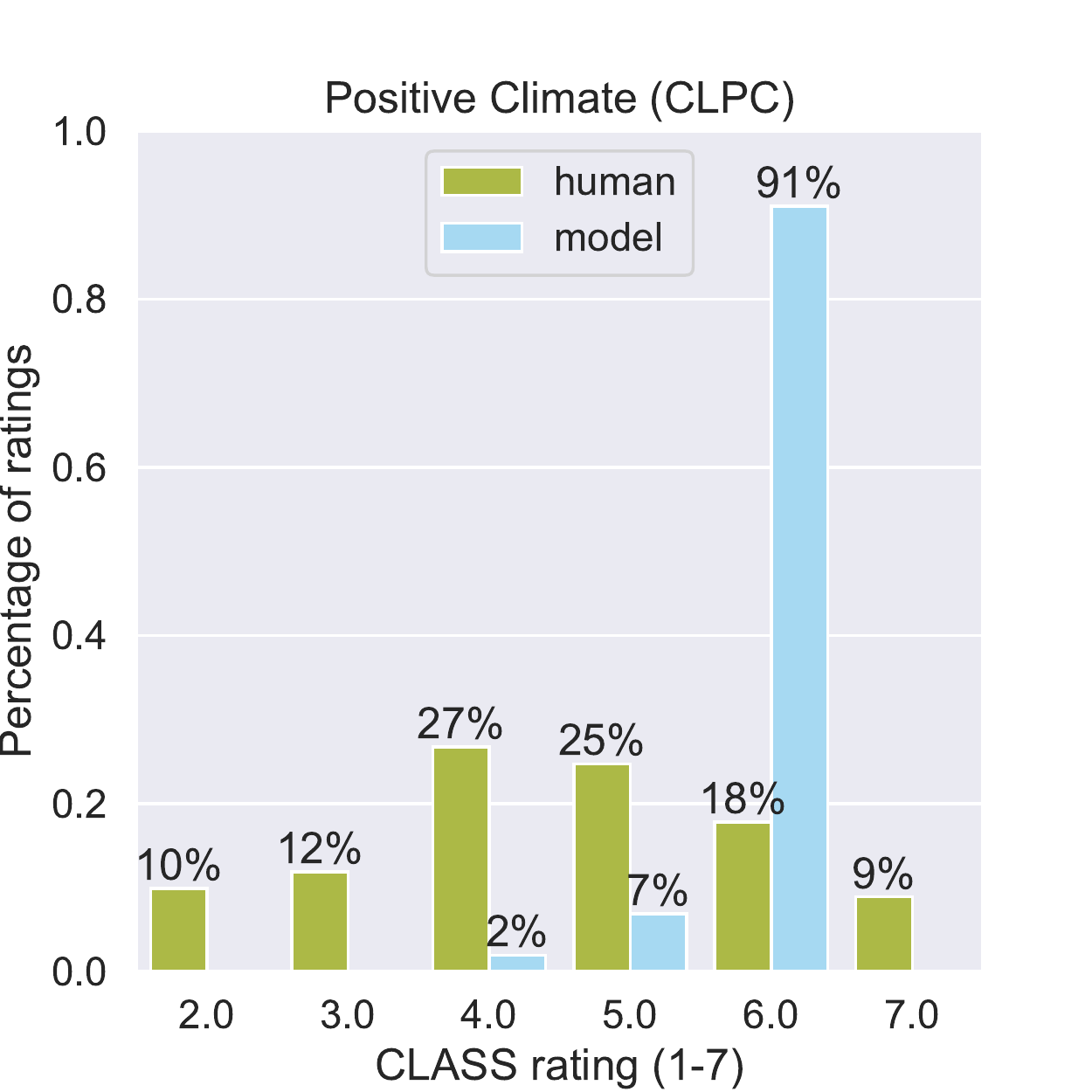}
                \caption{}
              \end{subfigure}
            & \begin{subfigure}{\ratio\linewidth}
                \includegraphics[width=\linewidth]{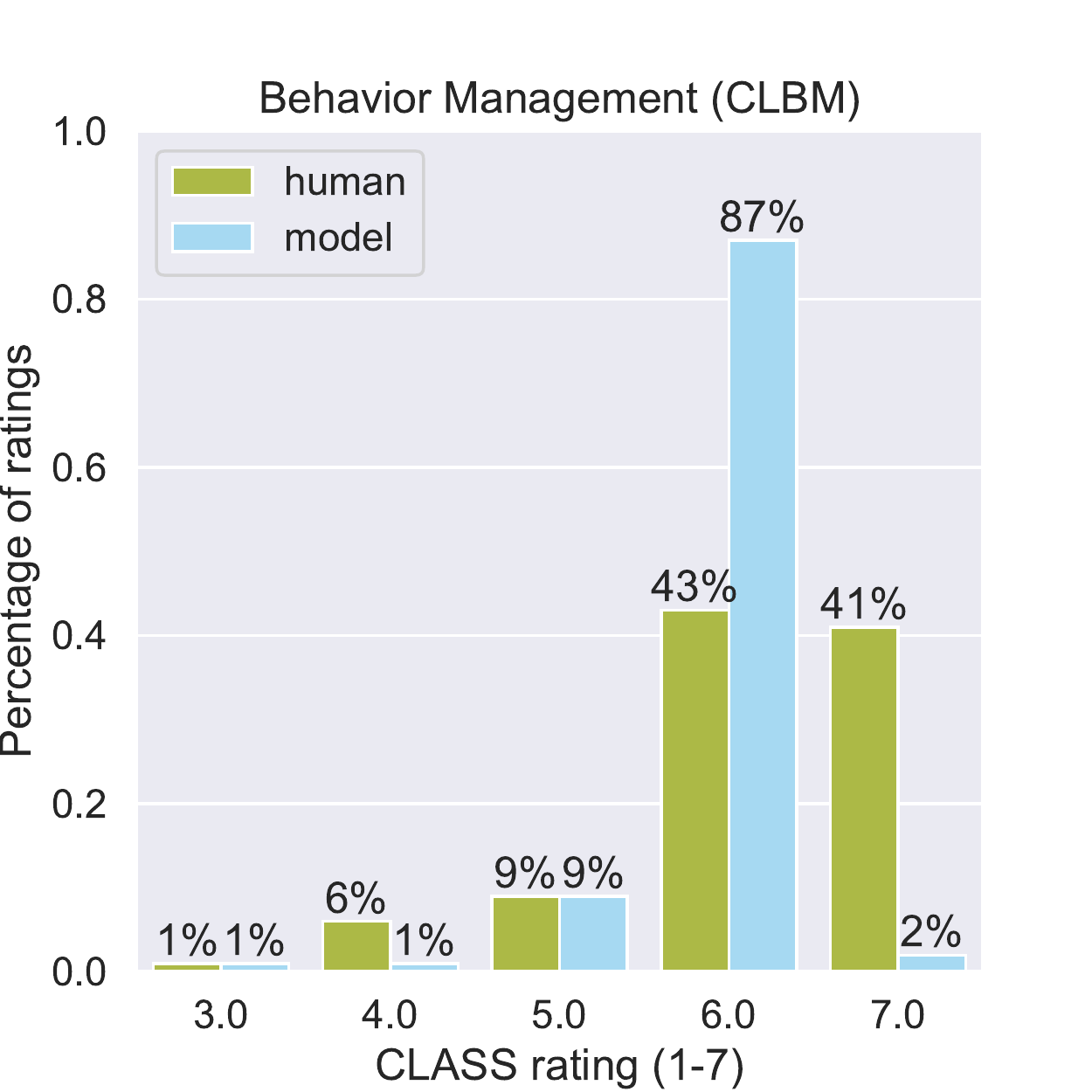}
                \caption{}
              \end{subfigure}
            & \begin{subfigure}{\ratio\linewidth}
                \includegraphics[width=\linewidth]{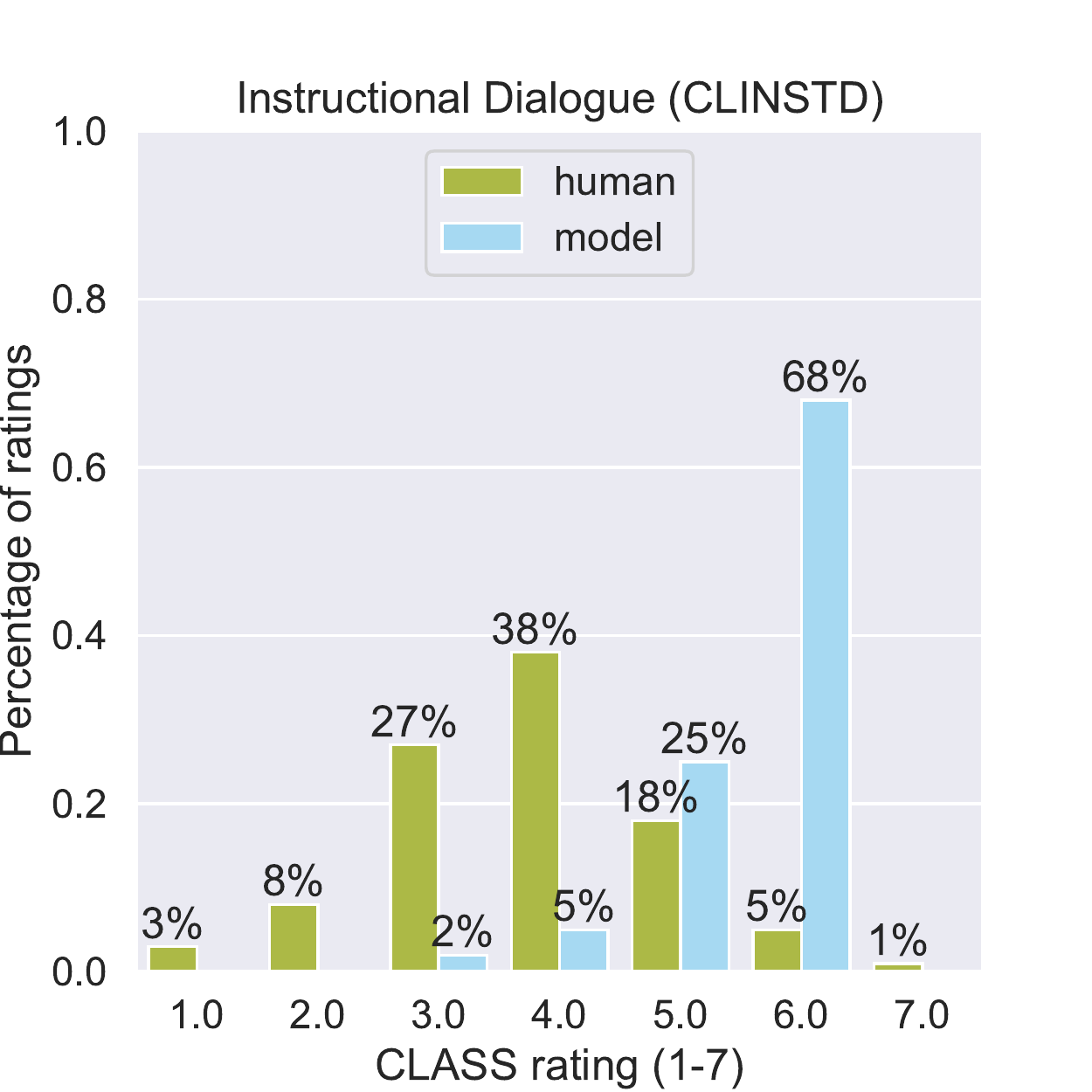}
                \caption{}
              \end{subfigure} \\
        \end{tabular}
        \caption{Bar plots comparing CLASS scores from humans vs. ChatGPT model.}
    \label{fig:class_prediction_barplots}
    \end{figure*}

\paragraph{CLASS}
Figure~\ref{fig:class_prediction_correlation} shows scatter plots of the model predicted scores vs. the human scores. 
It shows this across CLASS observation items and prompting methods (\directAnswer, \directAnswerDescription, and \reasoningAnswer). 
Figure~\ref{fig:class_prediction_barplots} shows the same data, but compares the human and model predicted score distribution.

\begin{figure*}[h]
    \newcommand{\ratio}{0.20}
    \newcommand{\offset}{-7}
    \centering
    \begin{tabular}{ccccc}
        \multicolumn{1}{c}{} & \multicolumn{4}{c}{} \\ 
        & \mqiExplanations & \mqiRemediation & \mqiErrors & \mqiStudentMath \\ 
            \multirow{-10}{*}{\rotatebox[origin=c]{90}{\directAnswer}} & 
            \begin{subfigure}{\ratio\linewidth}
                \includegraphics[width=\linewidth]{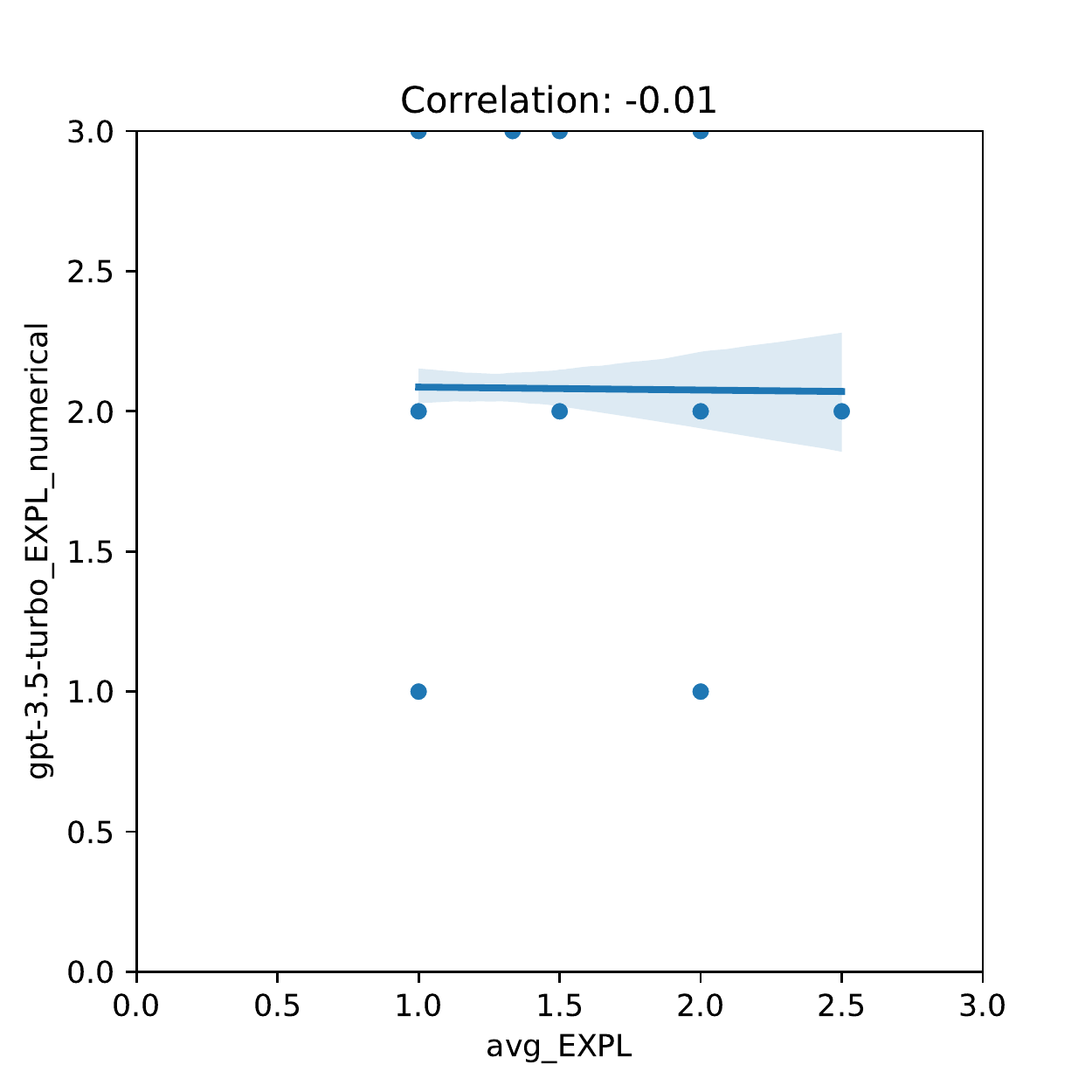}
              \end{subfigure}
            & \begin{subfigure}{\ratio\linewidth}
                \includegraphics[width=\linewidth]{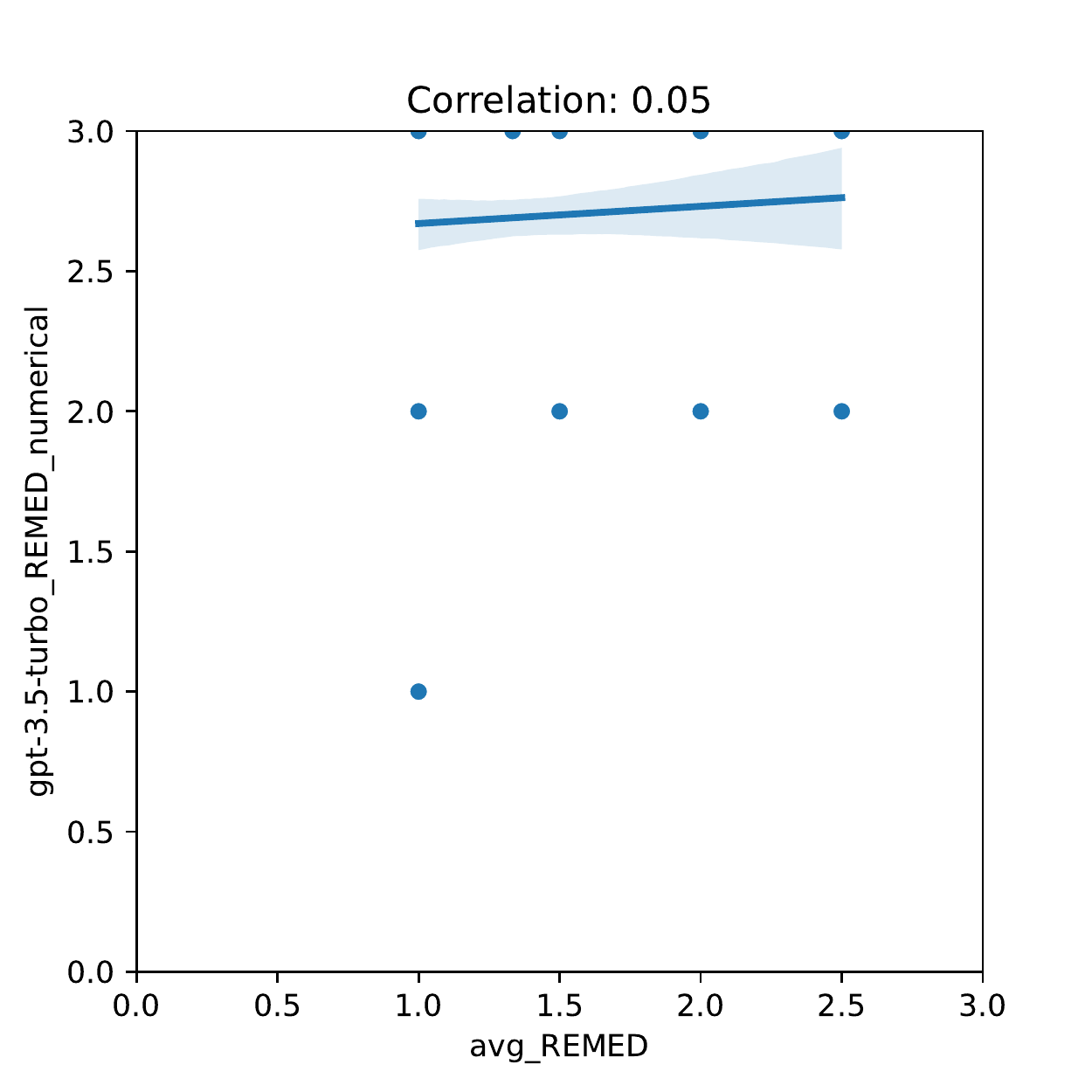}
              \end{subfigure}
            & \begin{subfigure}{\ratio\linewidth}
                \includegraphics[width=\linewidth]{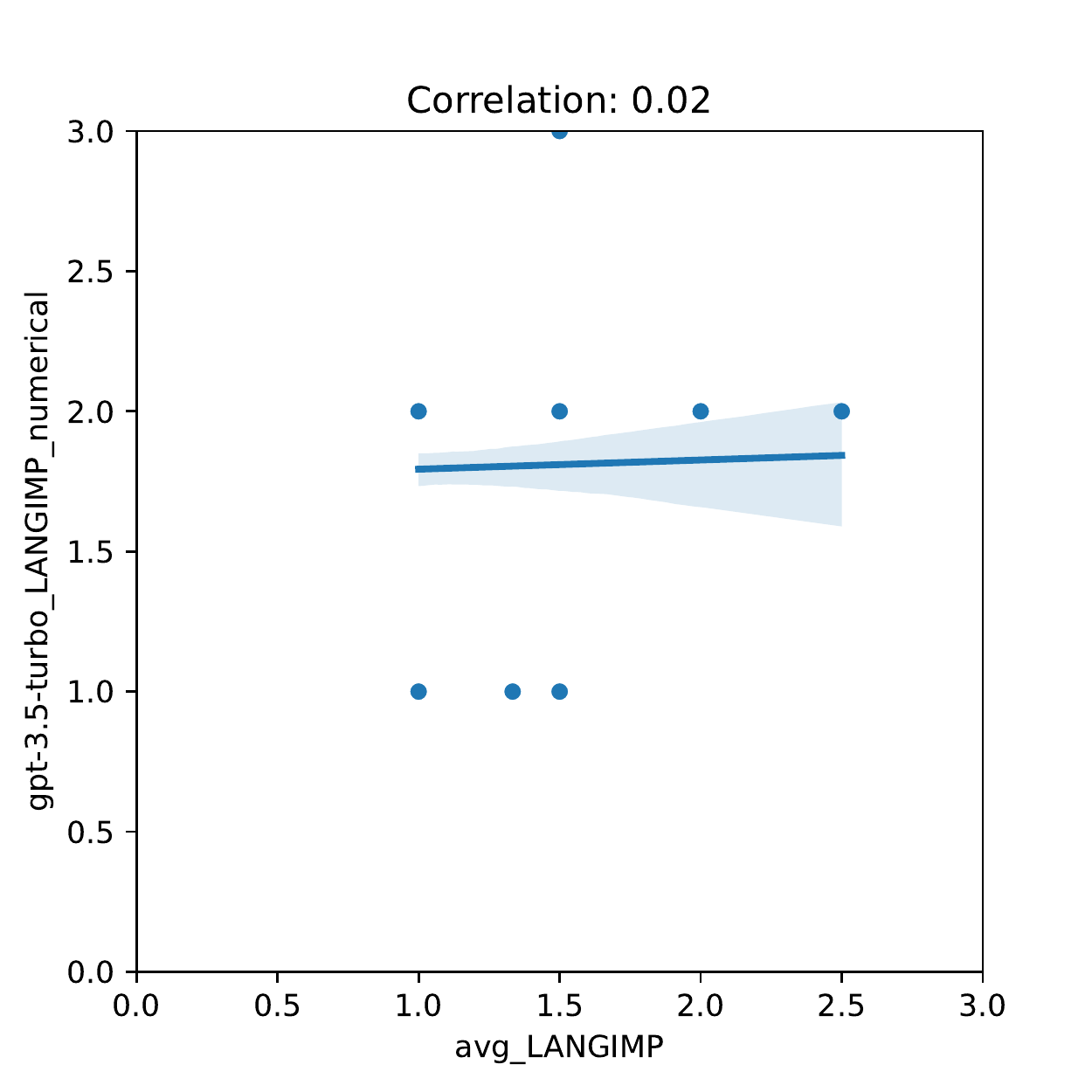}
              \end{subfigure}
            & \begin{subfigure}{\ratio\linewidth}
                \includegraphics[width=\linewidth]{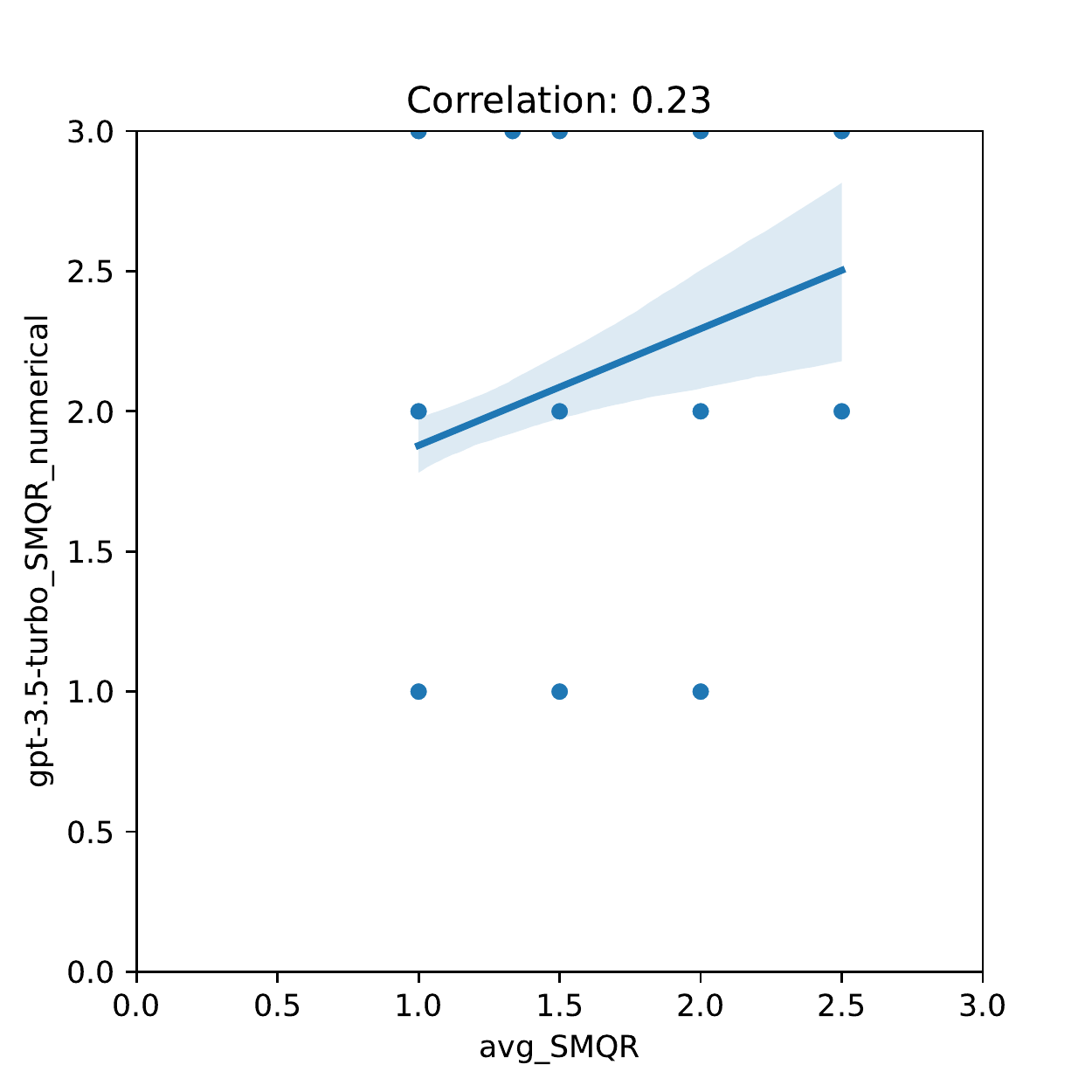}
              \end{subfigure} \\
            \multirow{-10}{*}{\rotatebox[origin=c]{90}{\directAnswerDescription}} & \begin{subfigure}{\ratio\linewidth}
                \includegraphics[width=\linewidth]{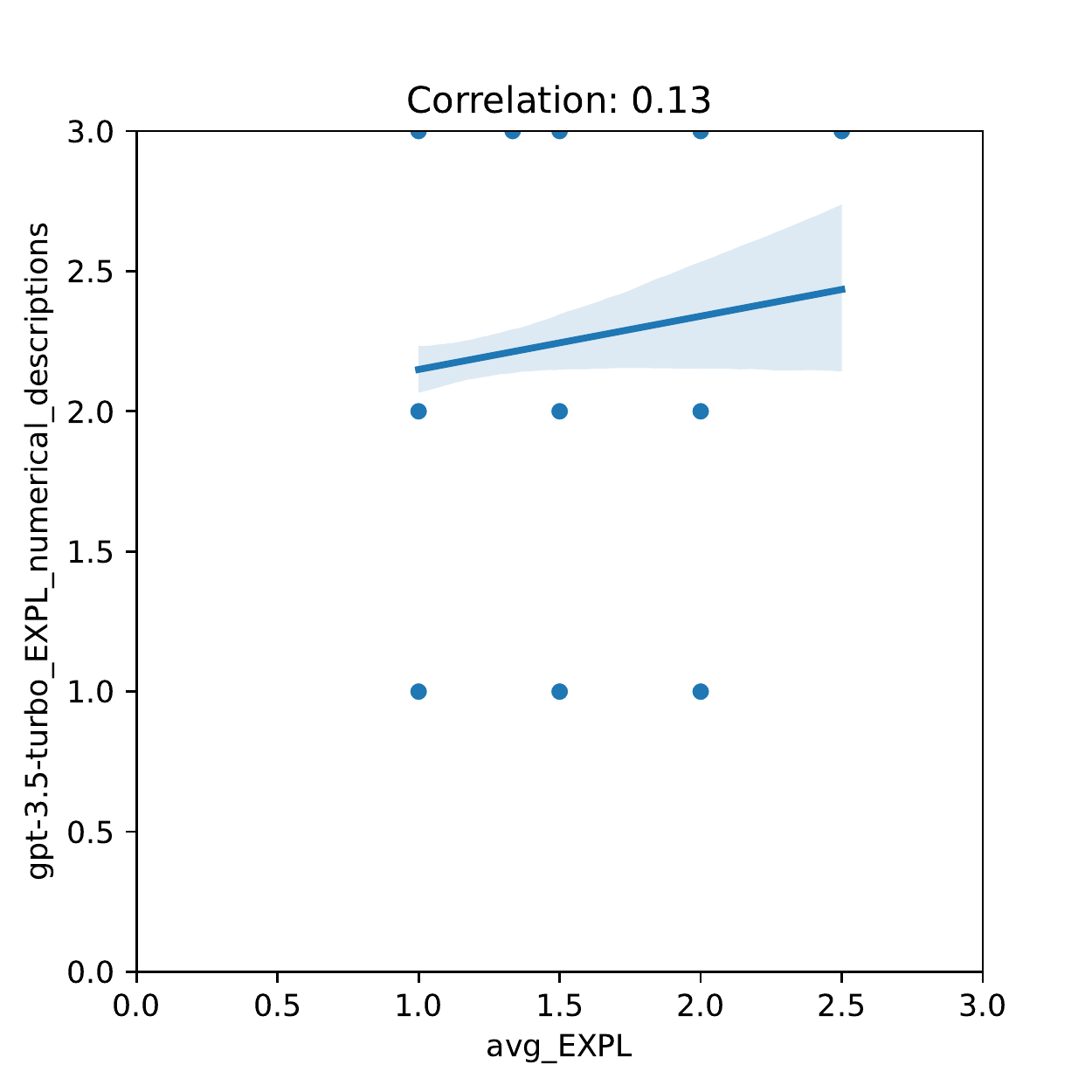}
              \end{subfigure}
            & \begin{subfigure}{\ratio\linewidth}
                \includegraphics[width=\linewidth]{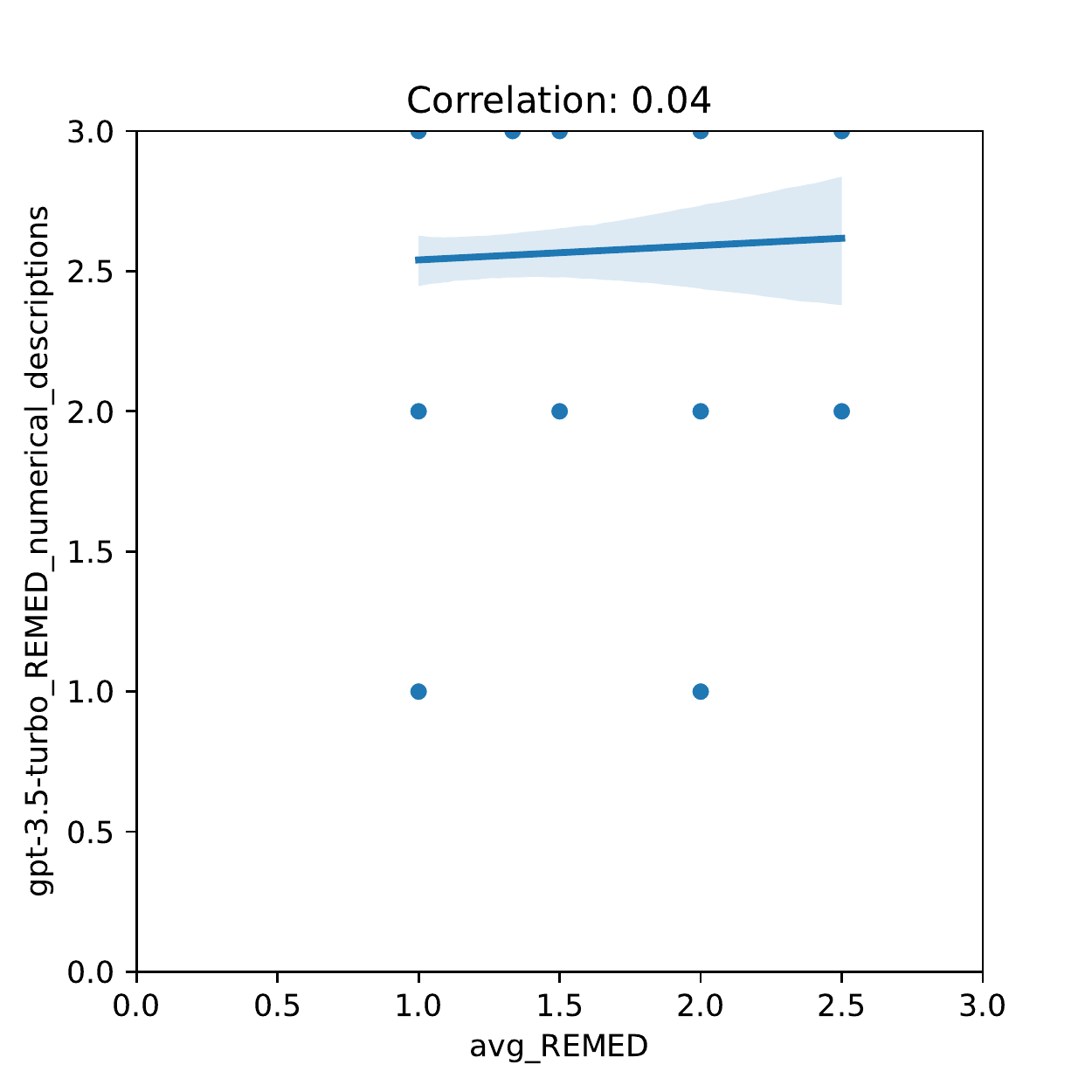}
              \end{subfigure}
            & \begin{subfigure}{\ratio\linewidth}
                \includegraphics[width=\linewidth]{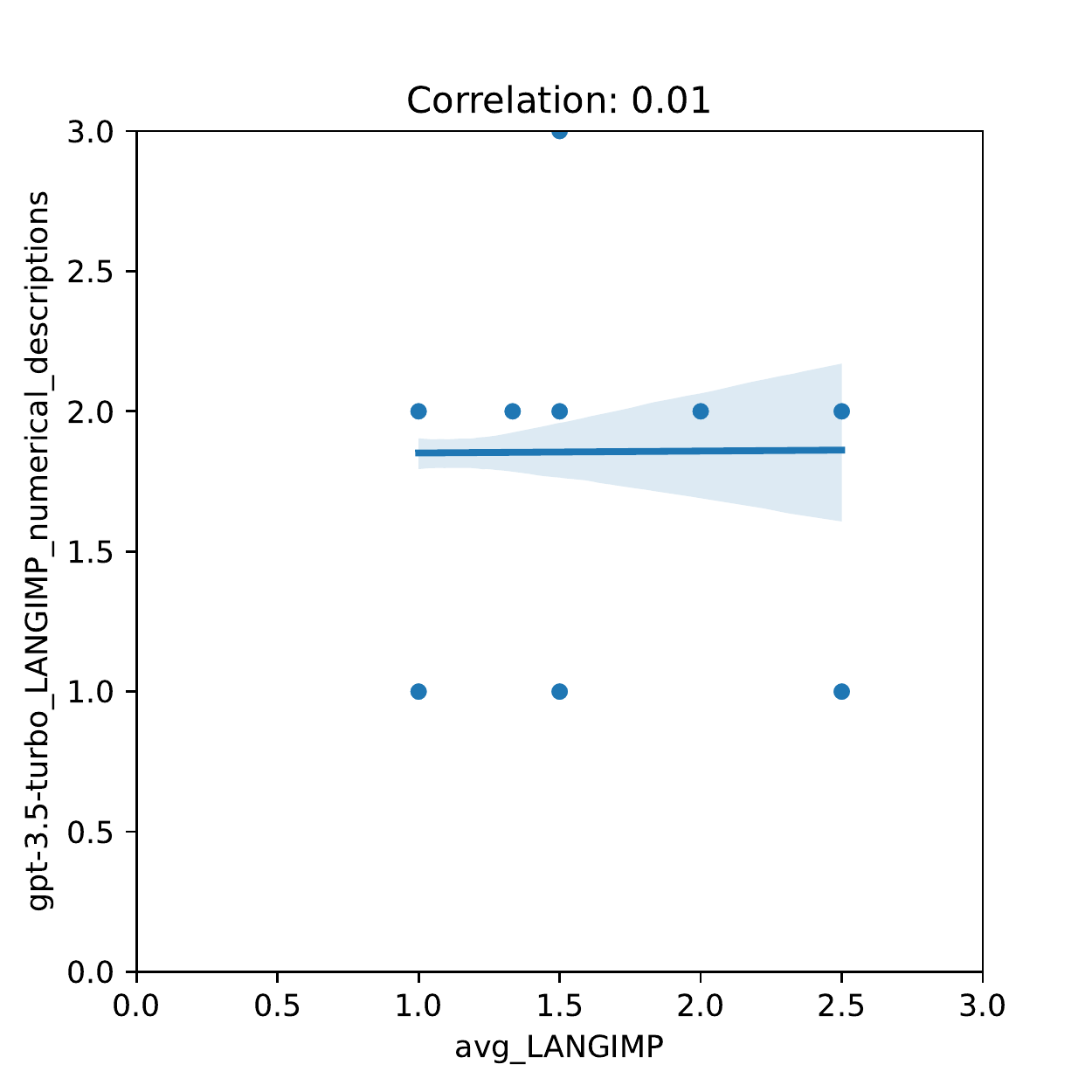}
              \end{subfigure}
            & \begin{subfigure}{\ratio\linewidth}
                \includegraphics[width=\linewidth]{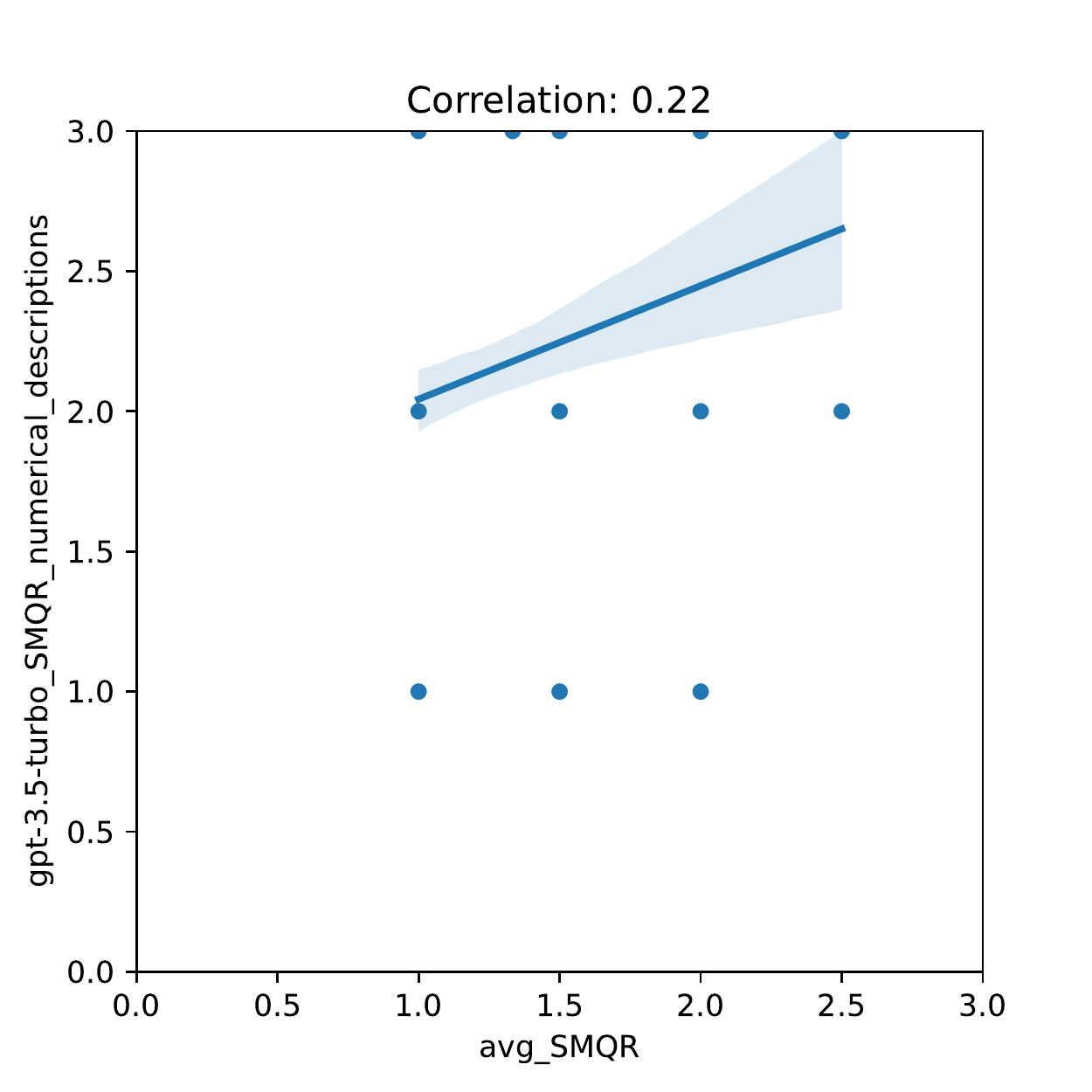}
              \end{subfigure} \\
            \multirow{-10}{*}{\rotatebox[origin=c]{90}{\reasoningAnswer}}& \begin{subfigure}{\ratio\linewidth}
                \includegraphics[width=\linewidth]{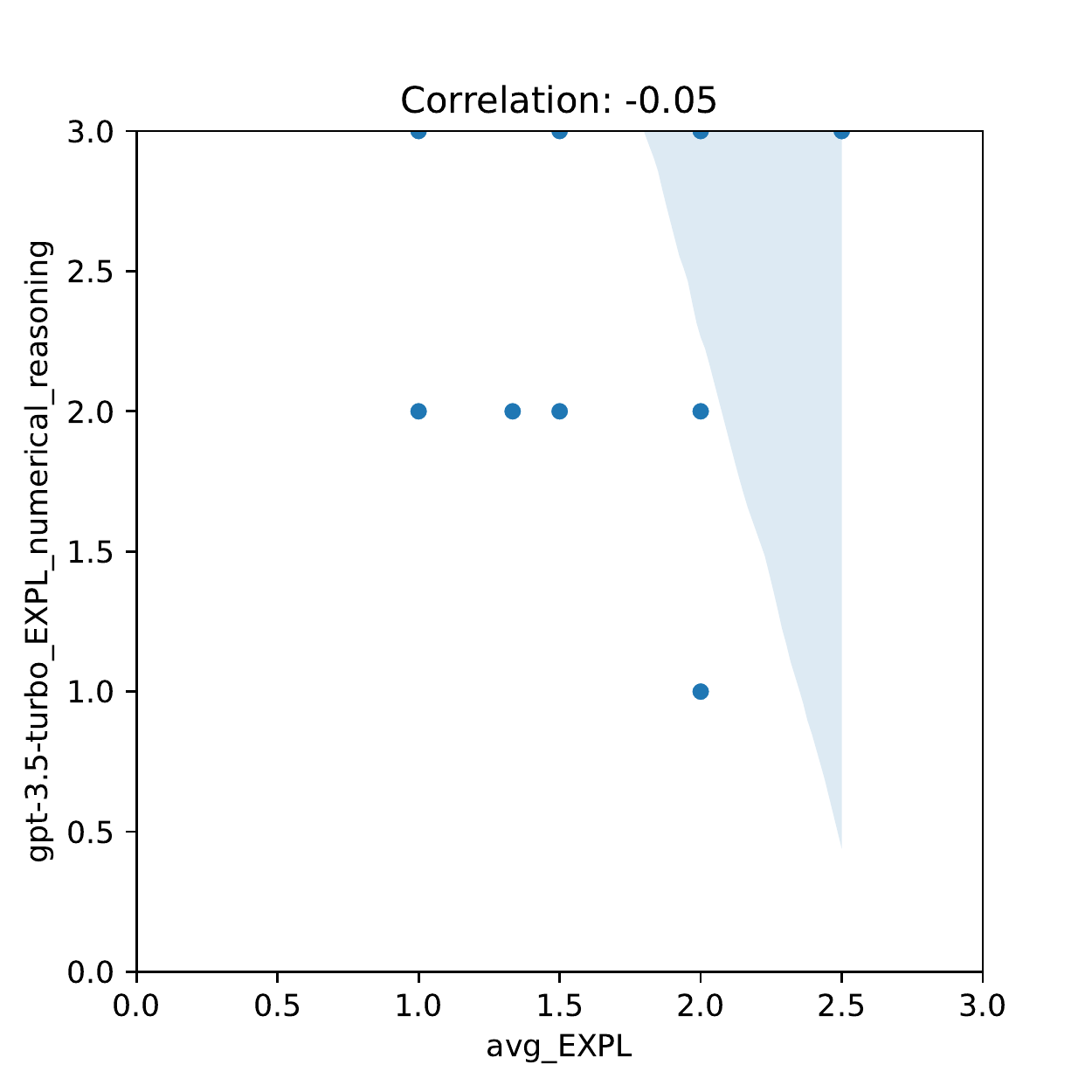}
              \end{subfigure}
            & \begin{subfigure}{\ratio\linewidth}
                \includegraphics[width=\linewidth]{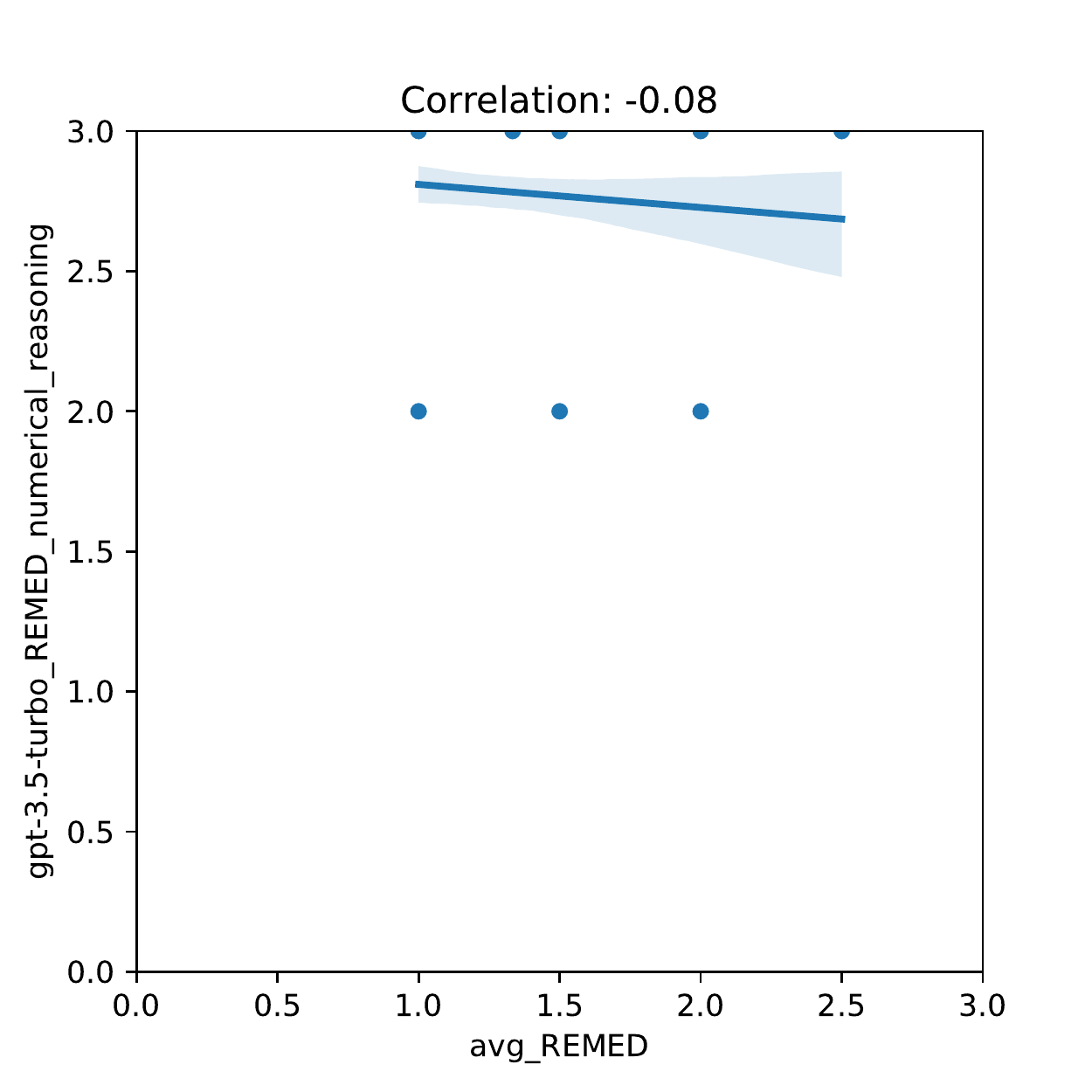}
              \end{subfigure}
            & \begin{subfigure}{\ratio\linewidth}
                \includegraphics[width=\linewidth]{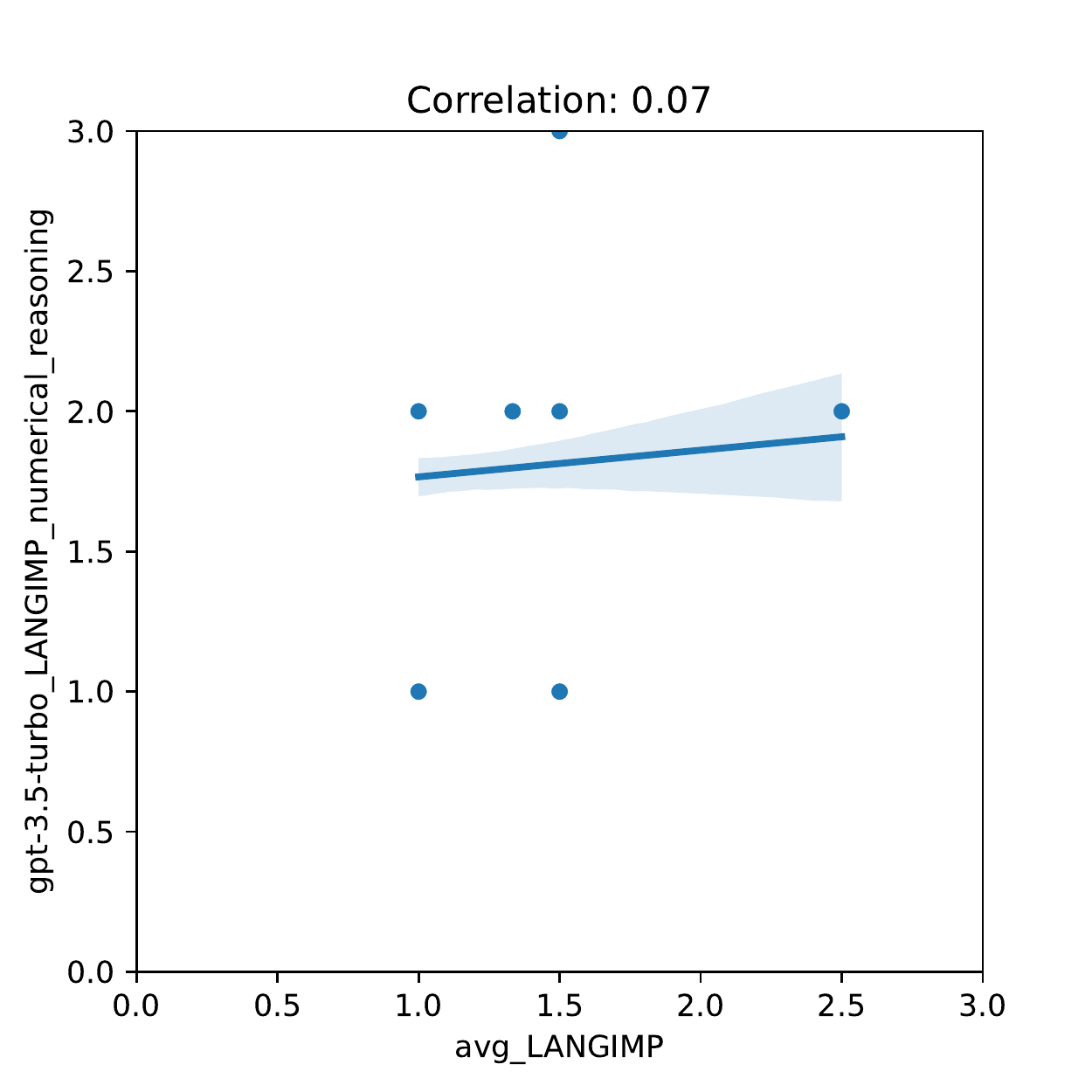}
              \end{subfigure}
            & \begin{subfigure}{\ratio\linewidth}
                \includegraphics[width=\linewidth]{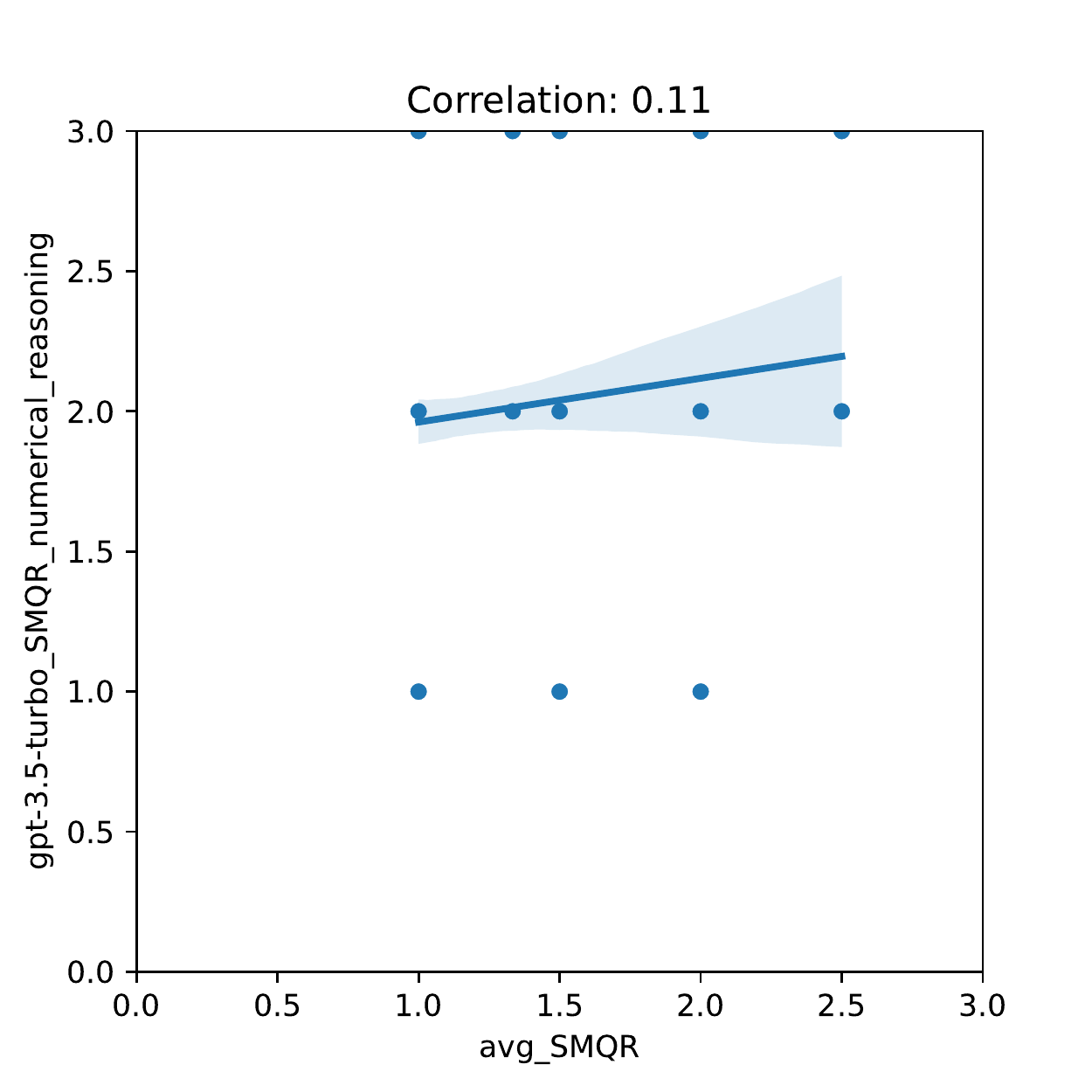}
              \end{subfigure} \\
        \end{tabular}
    \caption{Correlation between MQI annotations and model predictions.}    
    \label{fig:mqi_prediction_correlation}
\end{figure*}

\begin{figure*}[h]
    \newcommand{\ratio}{0.20}
    \newcommand{\offset}{-20}
    \centering
    \begin{tabular}{ccccc}
        \multicolumn{1}{c}{} & \multicolumn{4}{c}{} \\ 
        & \mqiExplanations & \mqiRemediation & \mqiErrors & \mqiStudentMath \\ 
            \multirow{-10}{*}{\rotatebox[origin=c]{90}{\directAnswer}} & 
            \begin{subfigure}{\ratio\linewidth}
                \includegraphics[width=\linewidth]{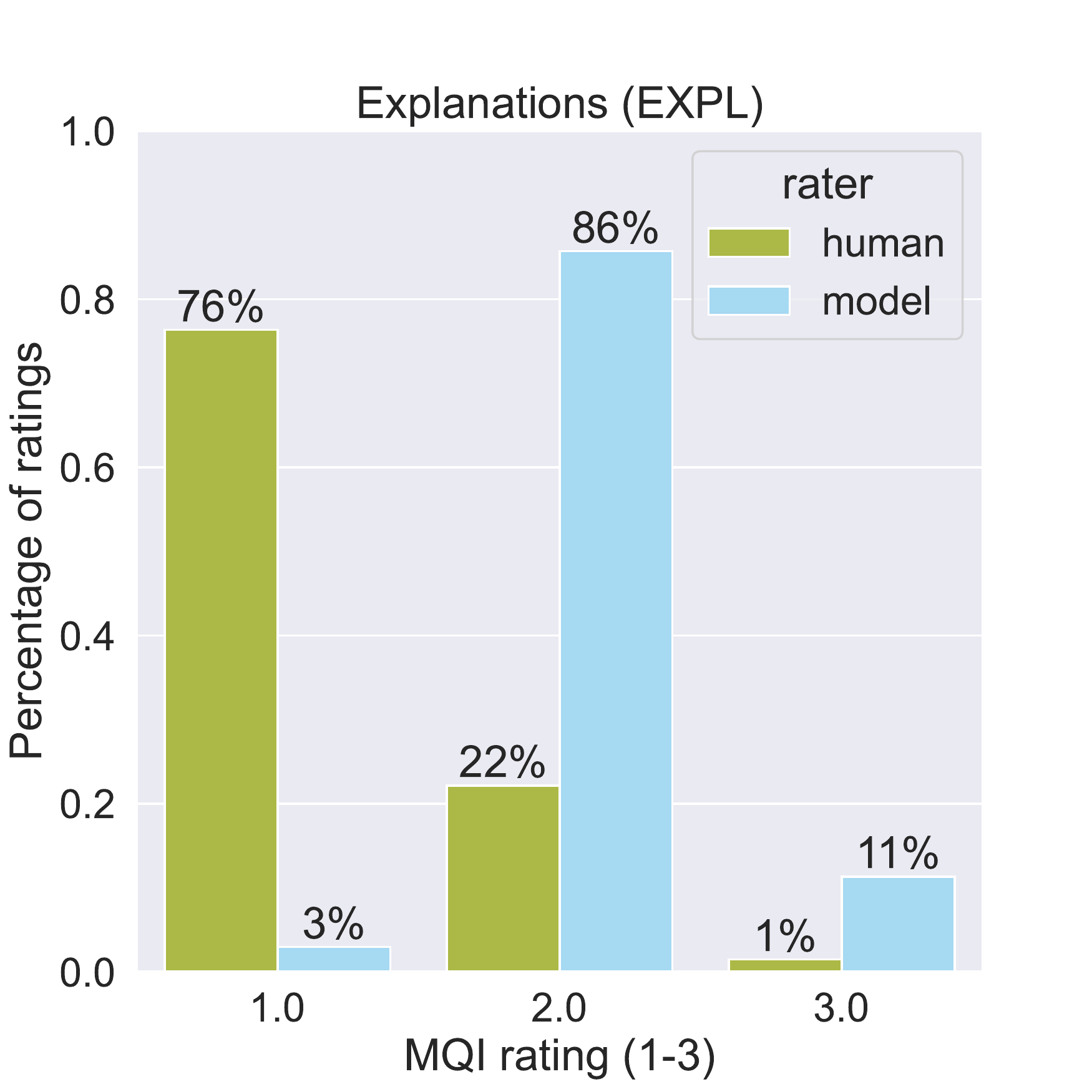}
              \end{subfigure}
            & \begin{subfigure}{\ratio\linewidth}
                \includegraphics[width=\linewidth]{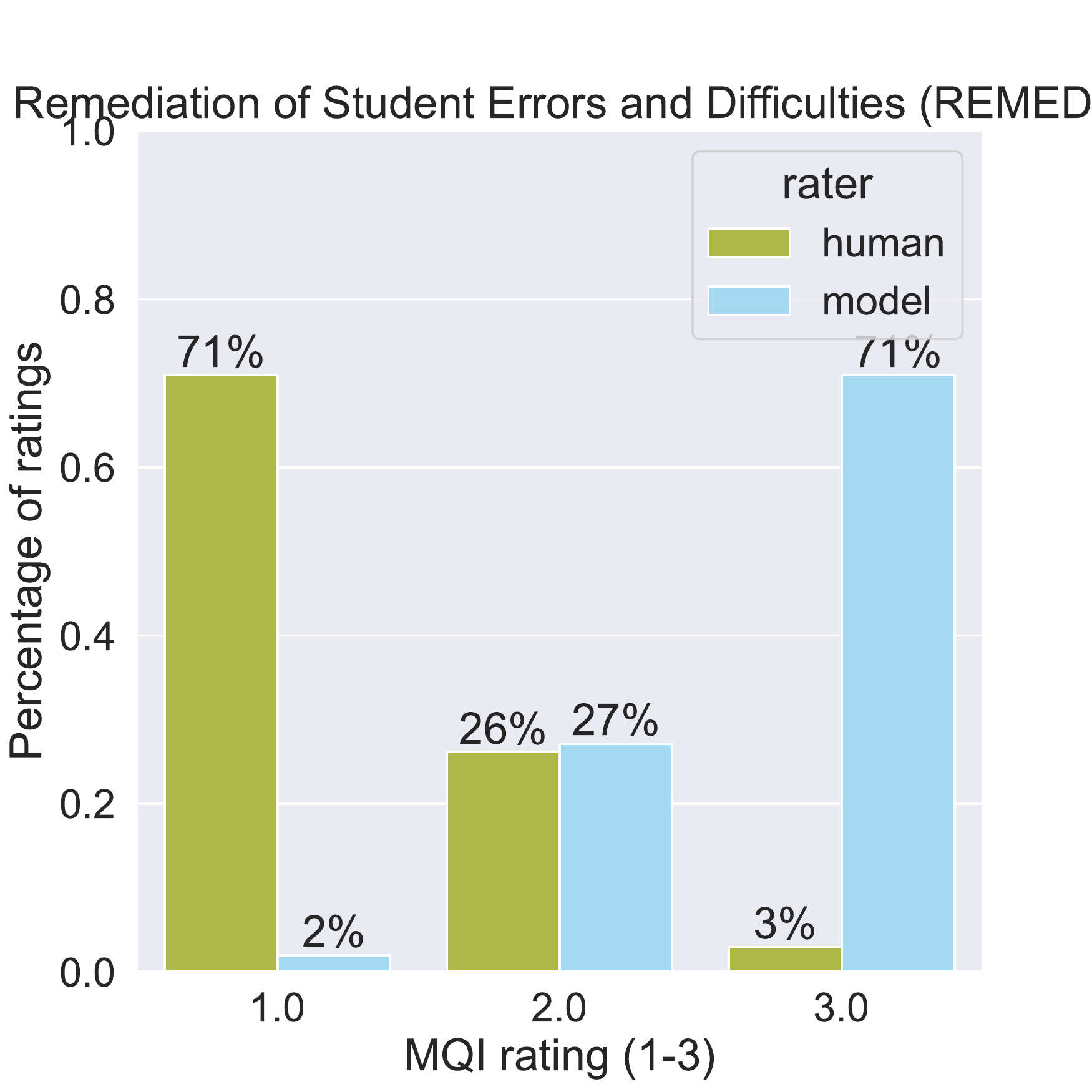}
              \end{subfigure}
            & \begin{subfigure}{\ratio\linewidth}
                \includegraphics[width=\linewidth]{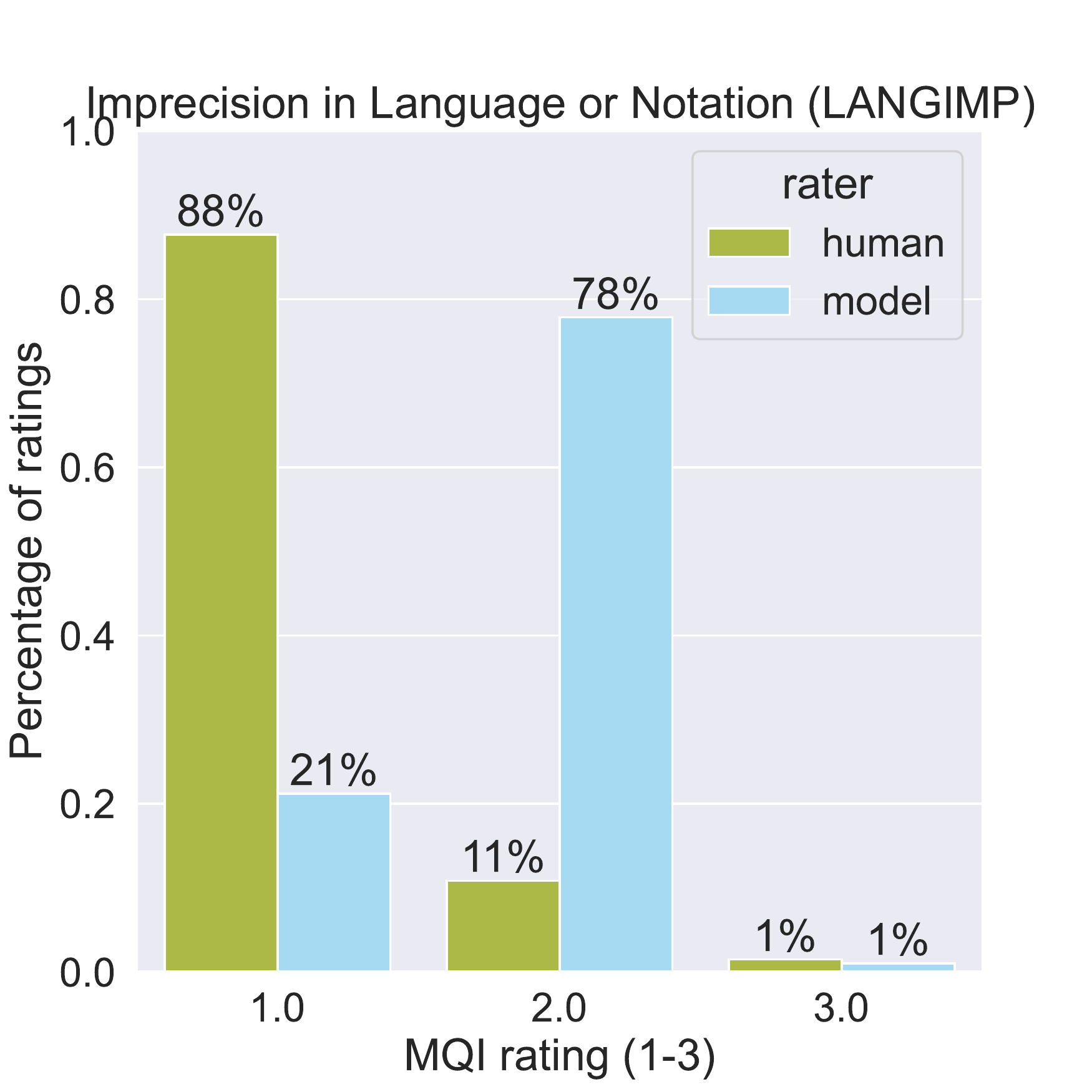}
              \end{subfigure}
            & \begin{subfigure}{\ratio\linewidth}
                \includegraphics[width=\linewidth]{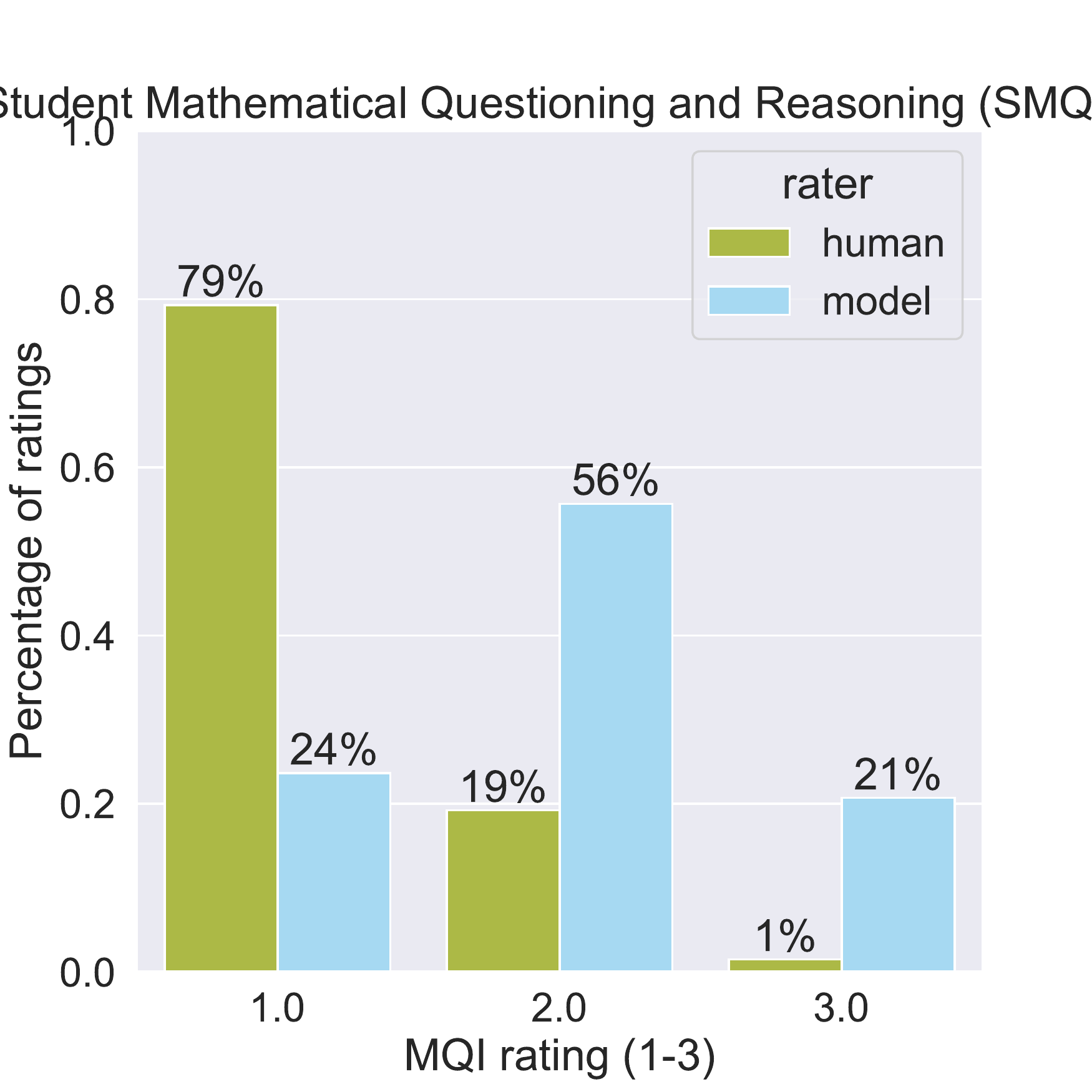}
              \end{subfigure} \\
            \multirow{-10}{*}{\rotatebox[origin=c]{90}{\directAnswerDescription}} & \begin{subfigure}{\ratio\linewidth}
                \includegraphics[width=\linewidth]{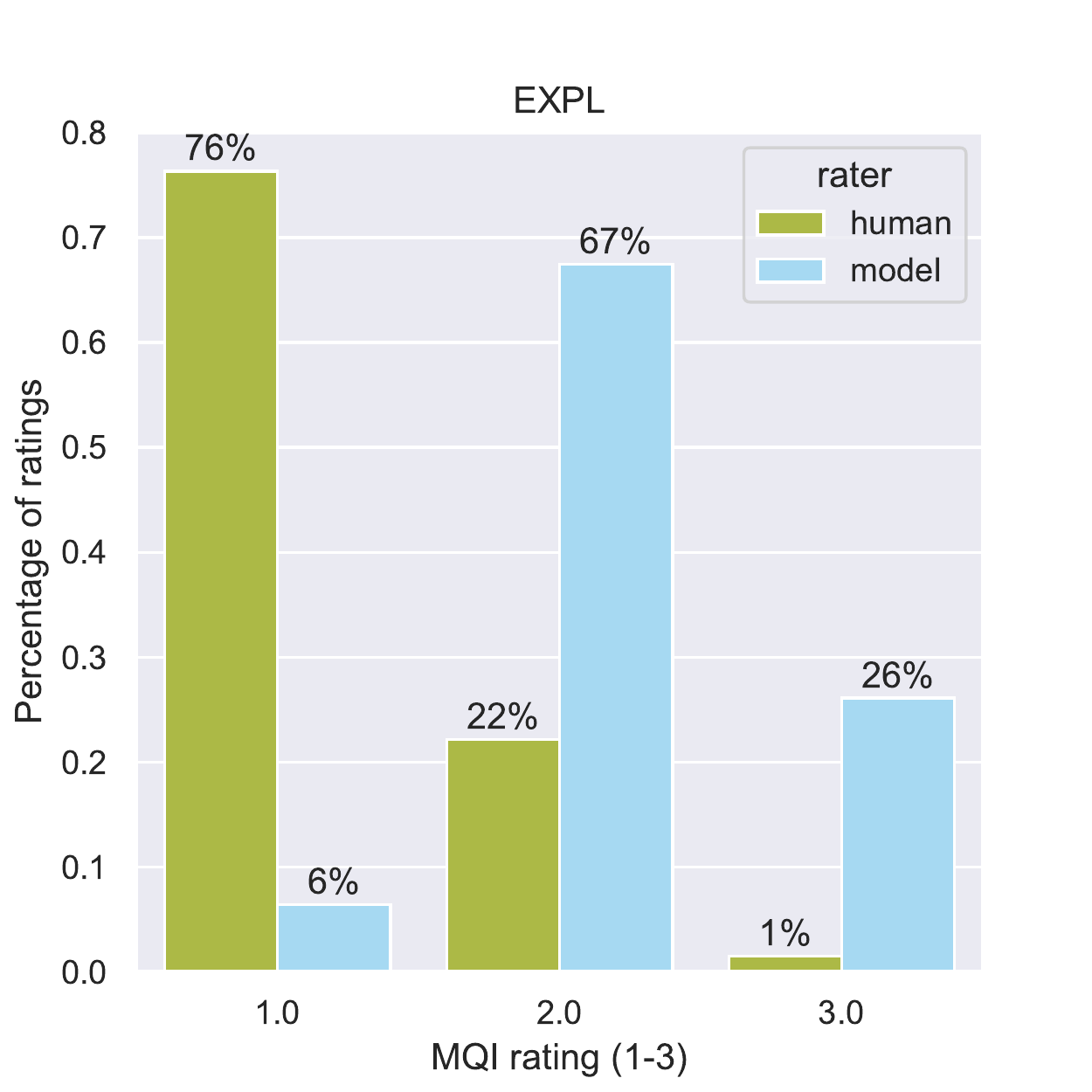}
              \end{subfigure}
            & \begin{subfigure}{\ratio\linewidth}
                \includegraphics[width=\linewidth]{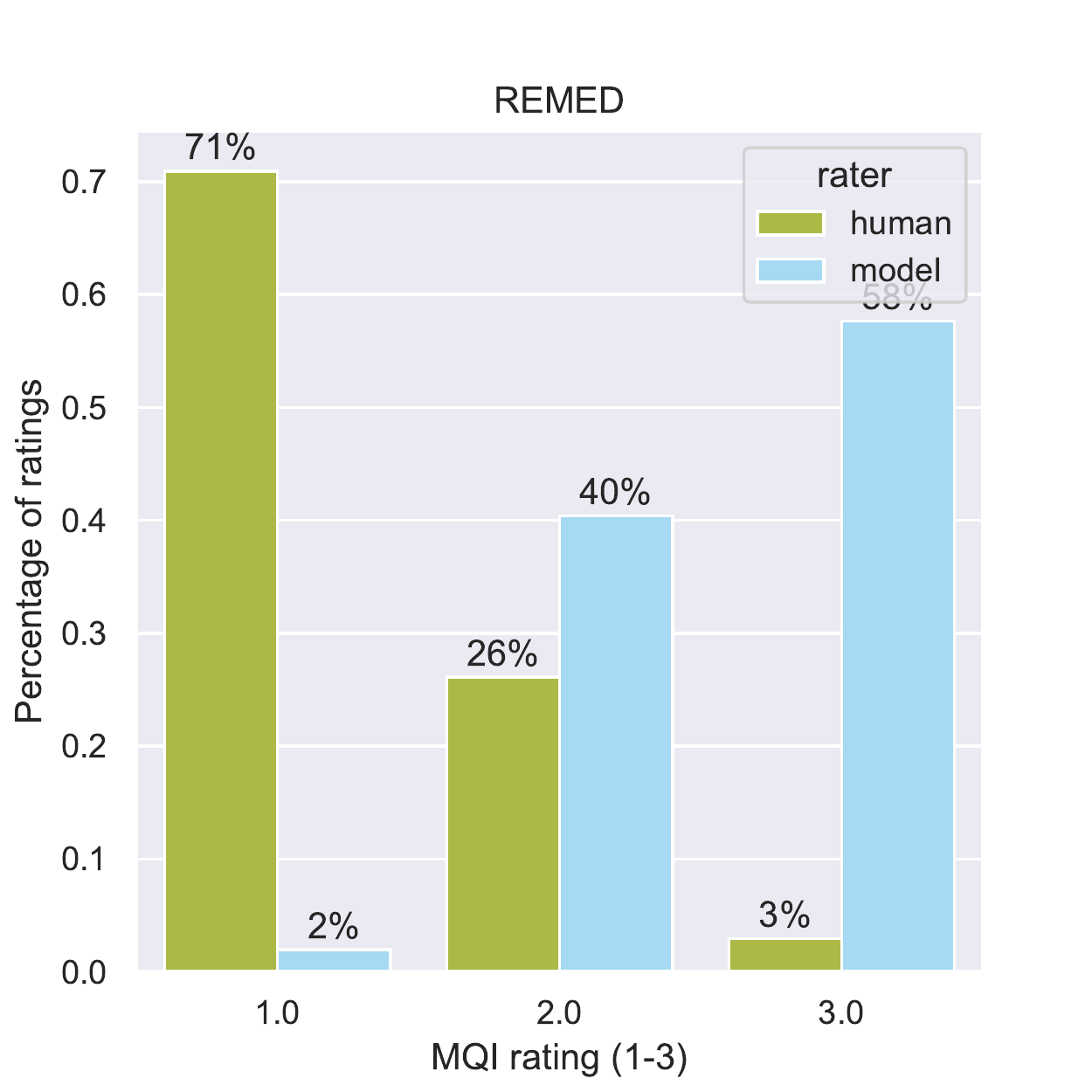}
              \end{subfigure}
            & \begin{subfigure}{\ratio\linewidth}
                \includegraphics[width=\linewidth]{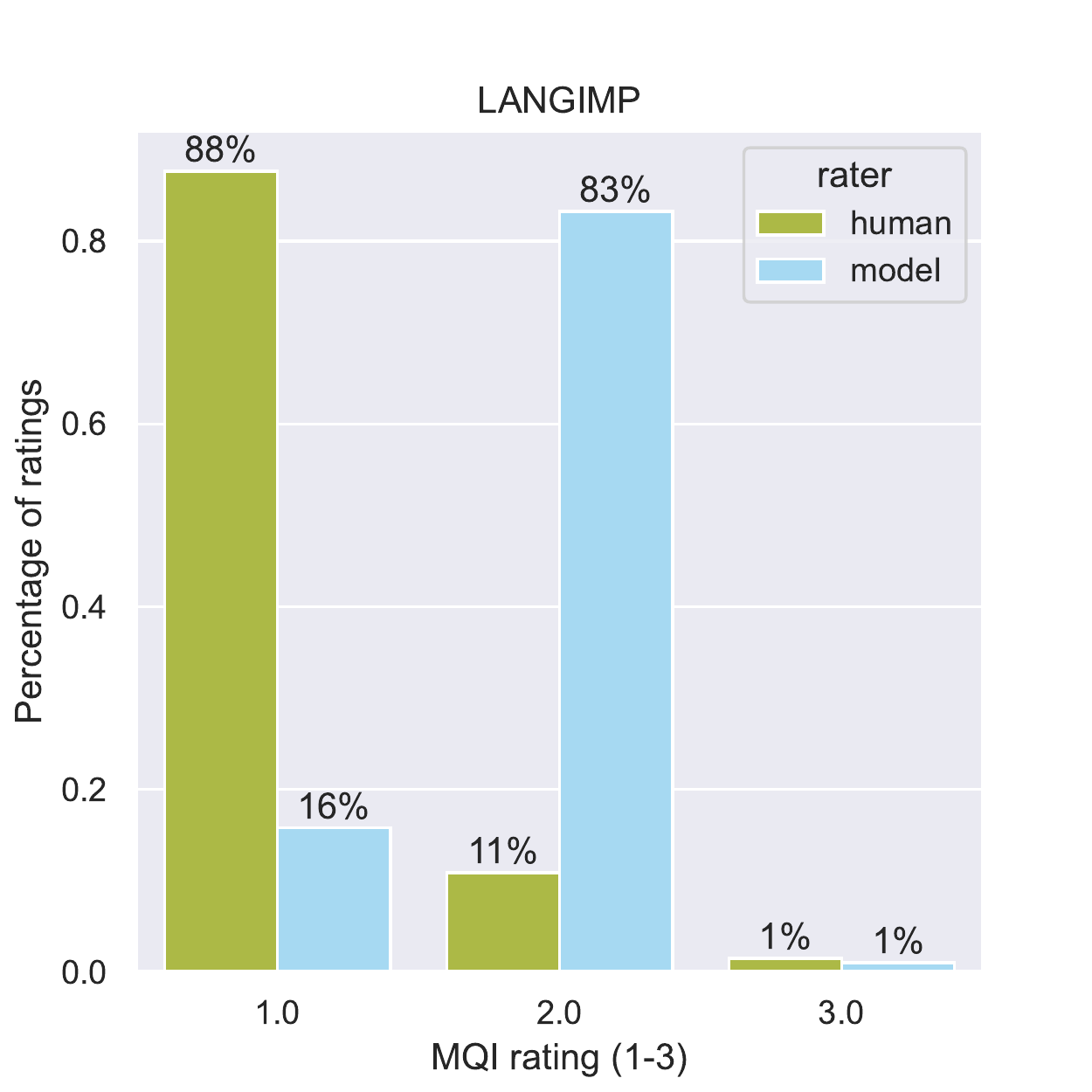}
              \end{subfigure}
            & \begin{subfigure}{\ratio\linewidth}
                \includegraphics[width=\linewidth]{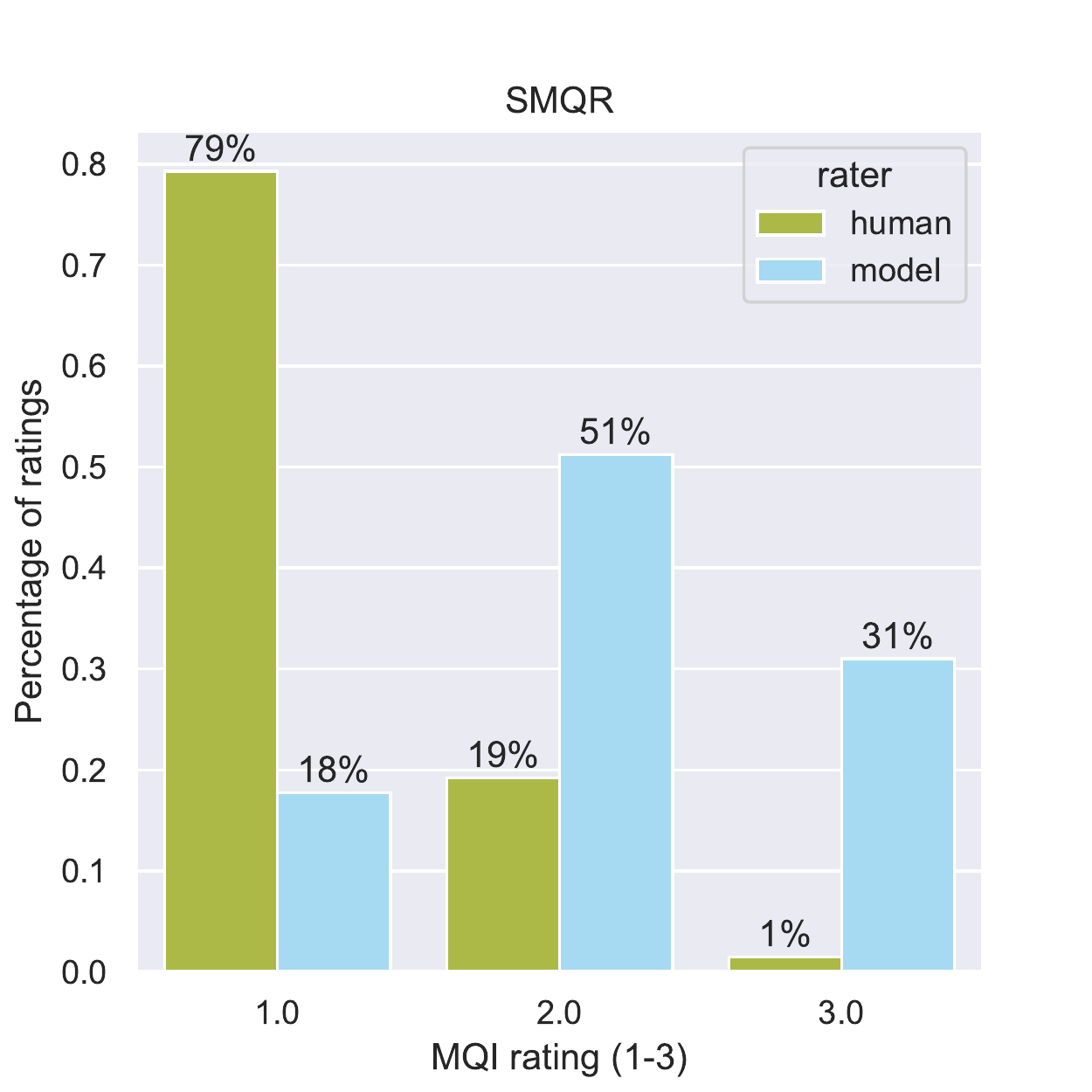}
              \end{subfigure} \\
            \multirow{-10}{*}{\rotatebox[origin=c]{90}{\reasoningAnswer}}& \begin{subfigure}{\ratio\linewidth}
                \includegraphics[width=\linewidth]{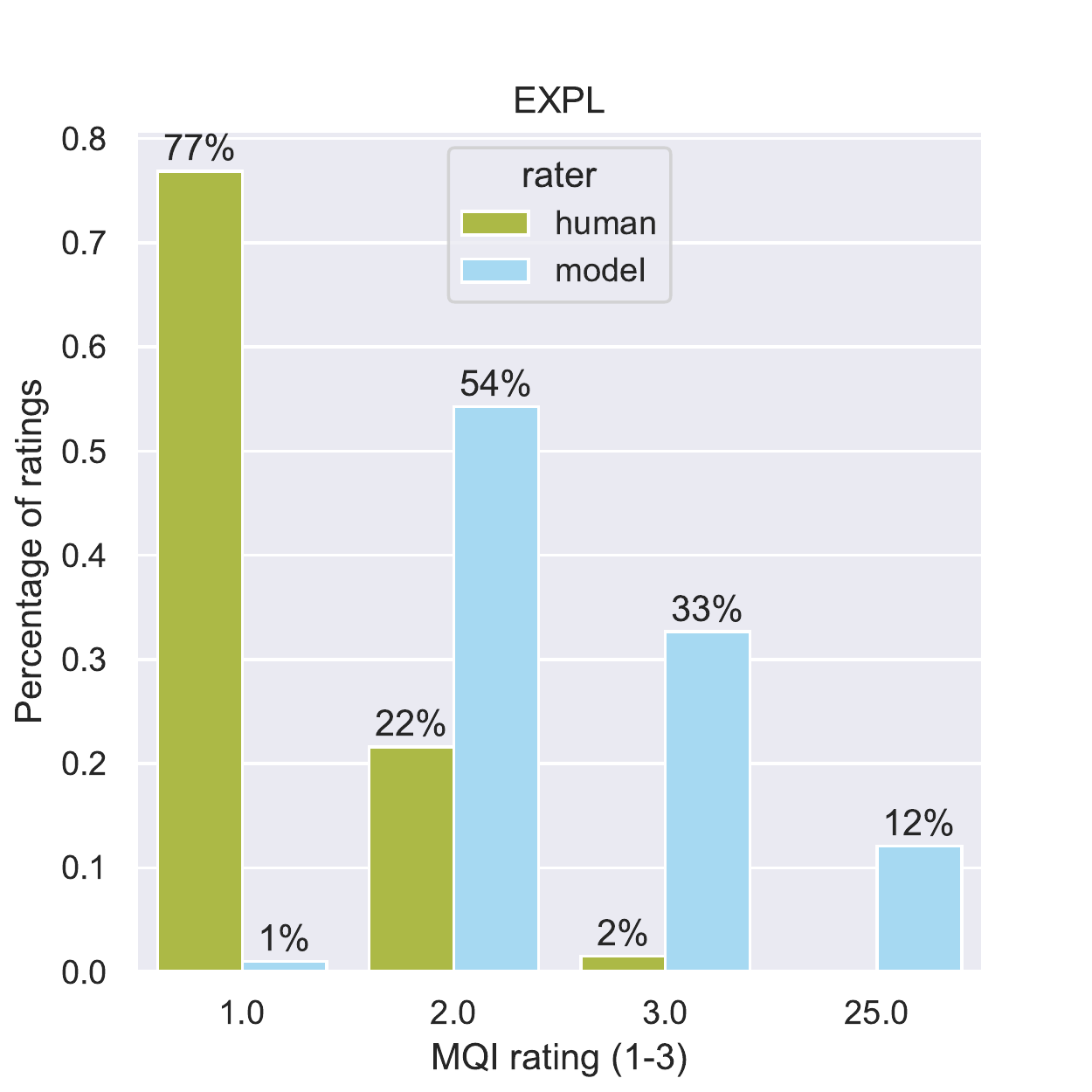}
              \end{subfigure}
            & \begin{subfigure}{\ratio\linewidth}
                \includegraphics[width=\linewidth]{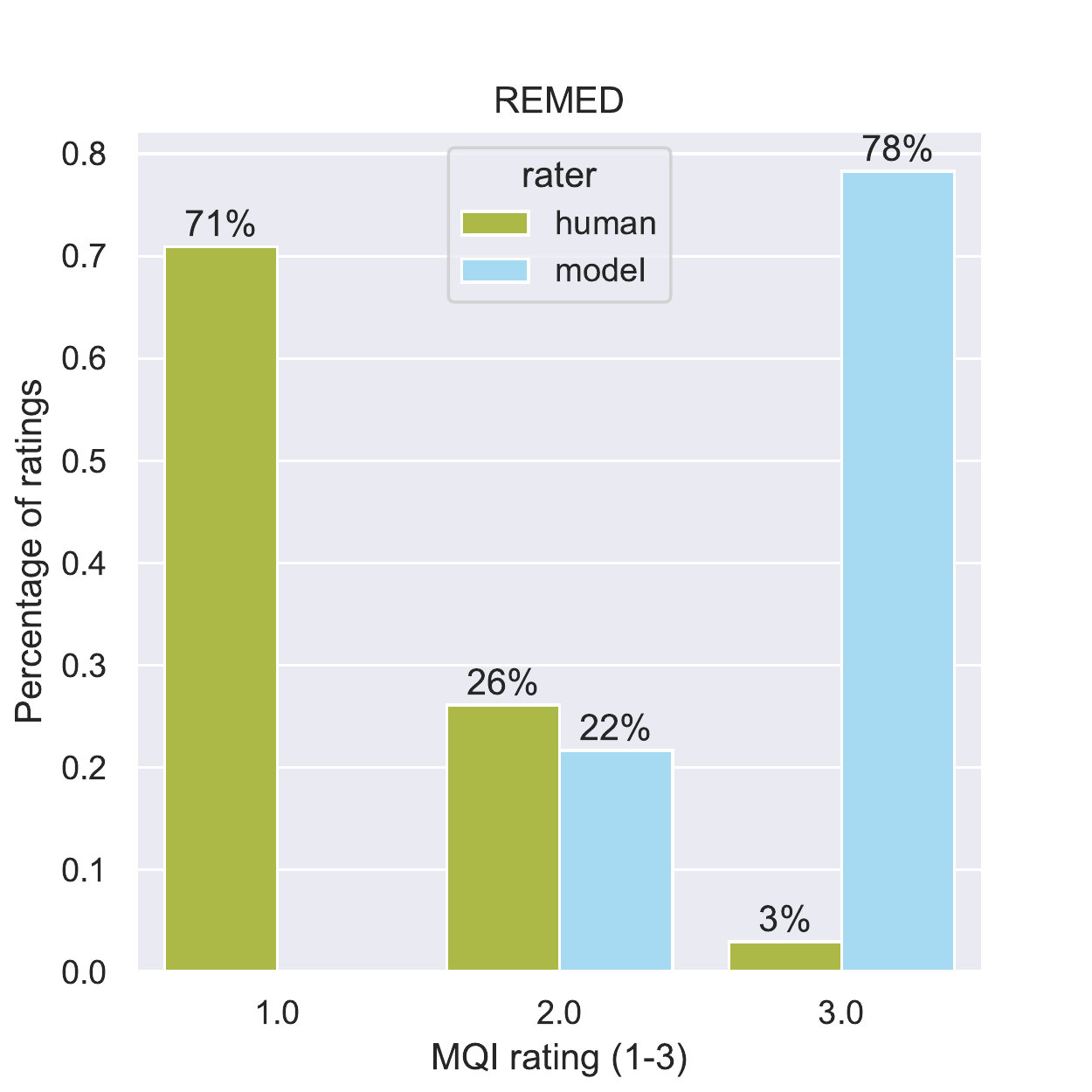}
              \end{subfigure}
            & \begin{subfigure}{\ratio\linewidth}
                \includegraphics[width=\linewidth]{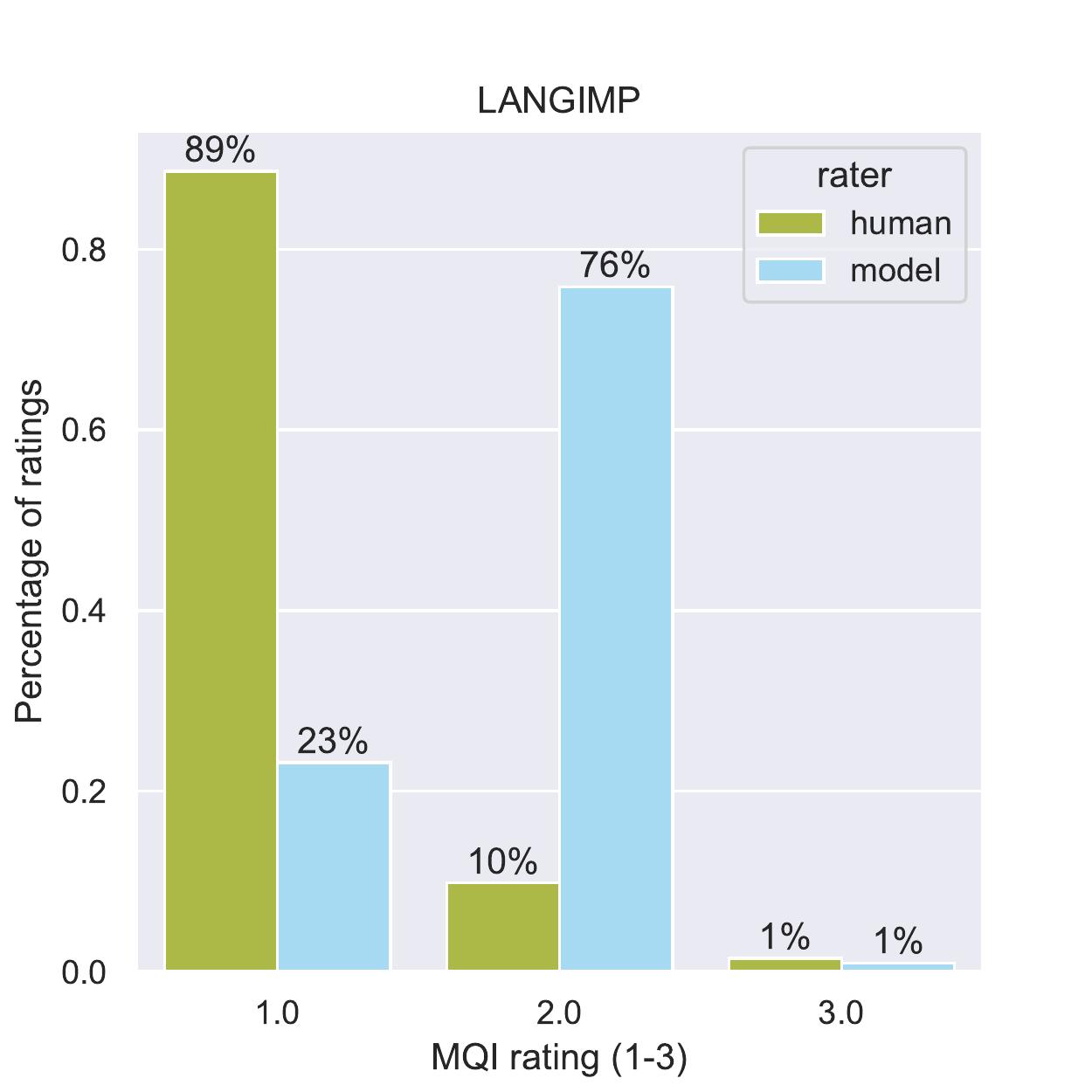}
              \end{subfigure}
            & \begin{subfigure}{\ratio\linewidth}
                \includegraphics[width=\linewidth]{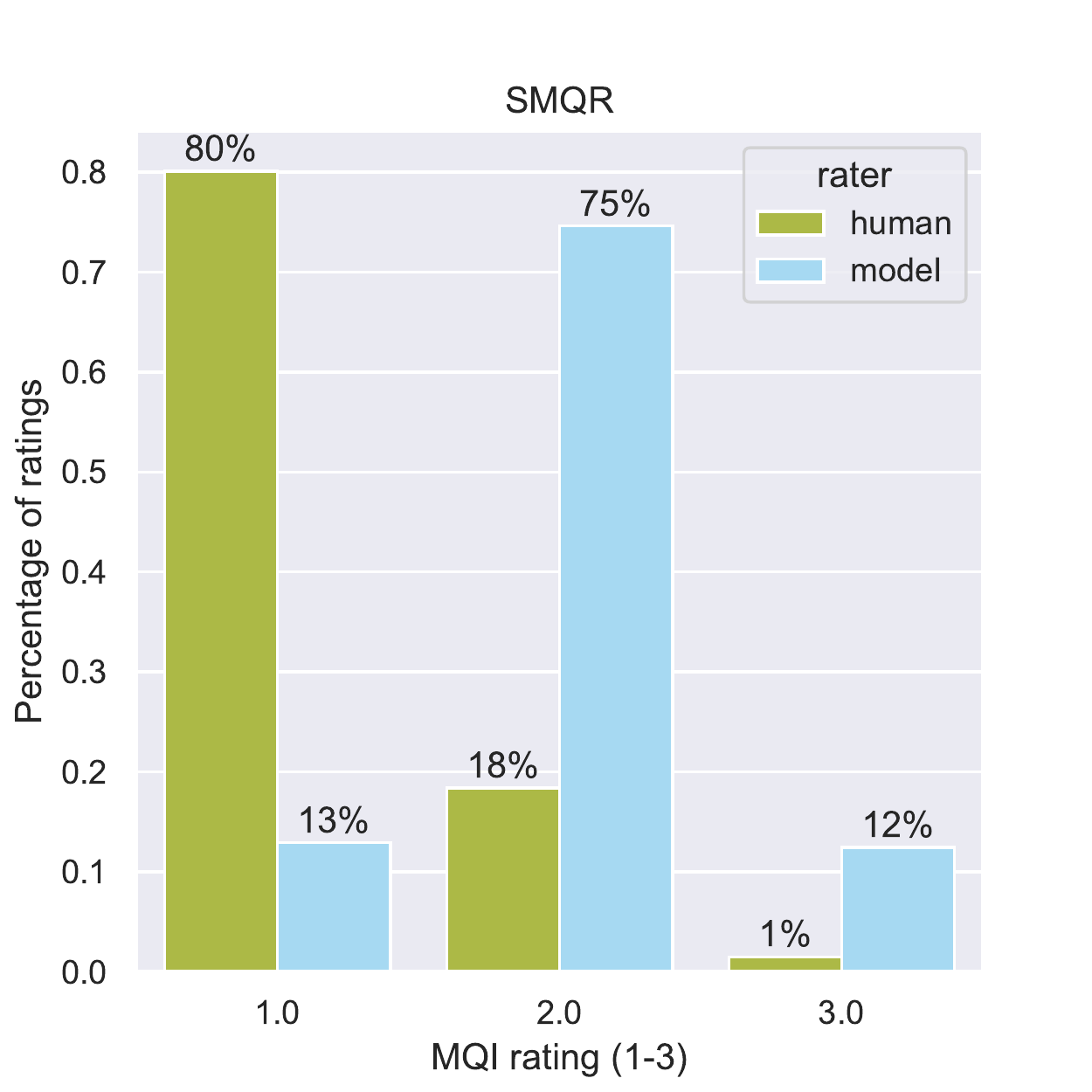}
              \end{subfigure} \\
        \end{tabular}
    \caption{Bar plots comparing MQI scores from humans vs. ChatGPT model.}
    \label{fig:mqi_prediction_barplots}
\end{figure*}

\paragraph{MQI}
Figure~\ref{fig:mqi_prediction_correlation} shows scatter plots of the model predicted scores vs. the human scores. 
It shows this across MQI observation items and prompting methods (\directAnswer, \directAnswerDescription, and \reasoningAnswer). 
Figure~\ref{fig:mqi_prediction_barplots} shows the same data, but compares the human and model predicted score distribution.

\subsection{Interrater Agreement}

We compute interrater agreement on the examples that both teachers rated (20\%). Since our goal was to collect teachers' unbiased perceptions, we did not conduct any calibration for this task; we leave this for future work. 
For task B, we measure a Cohen's kappa with linear weighting of $0.16$ for \textit{relevance}, $0.23$ for \textit{faithfulness}, and $0.32$ for \textit{insightfulness}. 
Figure~\ref{fig:confusion_matrix_examples} illustrates why there is particularly low agreement on relevance: 
One rater tends to select more extreme values for relevance, whereas the other rater selects more uniformly across the values. This results in low agreement for relevance.
The Cohen's kappas with quadratic weighting are $0.23$ for \textit{relevance}, $0.36$ for \textit{faithfulness}, and $0.37$ for \textit{insightfulness}.
The Cohen's kappas with quadratic weighting is slightly higher as it adjusts the penalty between scores 1 and 3 to be different from the penalty between scores 1 and 2 for instance.
For Task C, we only have 2 examples per criterion, which is too sparse for computing Cohen's kappa.

\section{Examples of Transcripts, Model Responses, and Human Evaluations \label{app:examples}}

Figure~\ref{fig:suggestion_prompt_example} shows a concrete example of the suggestions prompt given to the model.
Figure~\ref{fig:suggestion_prompt_example_model_response} then shows one of the suggestions that the model generates. 
Figure~\ref{fig:suggestion_prompt_example_human_response} then shows the ratings provided from one of the human annotators on that suggestion.

\begin{figure*}[t]
    \centering 
    \begin{tcolorbox}[
    suggestionprompt,
    title={\textbf{Model prompt}},
    ]
    \small
    Consider the following classroom transcript. \\

Transcript: \\
    1. teacher: Well, it is division.  Take my word for it.  I'll write them bigger next time.  Raise your hand to tell me, what should I do first?  Student H, what are you going to do first? \\
2. student: What's in the parenthesis. \\
3. teacher: So you're going to do 30 minus 6 first?  And what did you get? \\
4. student: 23. \\
5. teacher: Check your subtraction. \\
6. student: I got 24. \\
7. teacher: You still got that?  What's 10 minus 6? \\
8. student: 4. \\
9. teacher: So 30 minus 6 can't be 23.  It has to be— \\
10. multiple students: 24. \\
11. teacher: Now look.  When I saw we do it like this – this is what we did last week.  24 goes right in the middle of those parenthesis.  Next I have to bring down what I didn't use. What is 24 divided by 3? \\
12. student: 8. \\
13. teacher: Your final answer should have been 8.  Does that make sense? \\
14. multiple students: Yes. \\
15. teacher: Do you want to do one more example? \\
16. multiple students: Yes. \\
17. multiple students: No. \\
18. teacher: We're going to do one more.  I'm going to make it a hard one. You have to follow the steps in the order of operations to get it right.  If I do what I think is easiest first, the answer is going to end up different.  Here's a hard one. Ready?  Copy and complete this one.  This has more than three steps.  It is 5 times 23 minus 18 plus 7.  It looks really hard, but if I follow my steps, it won't be so bad. You have to make sure the parenthesis are in the right spot.  For those who are stuck, raise your hand and tell me what should they do first and why.  Student M?  What are you doing first? \\
19. student: 5 times— \\
20. teacher: Is that your order of operations?  Is that what you should do first?  What's in parenthesis? \\
21. student: 23 minus 18. \\
22. teacher: Do 23 minus 18 first.  You'll find out.  Remember to allow everyone to work.  You are quiet.  If you're wondering whether you're right or wrong, you'll see in just a minute when we do it together.  Where did you write it?  Look at the problem that's on the board.  It's a different one now. Watch how it looks when you solve it.  Student H, here's how it looks when you solve it.  The first step you said was to do 23 minus 18, which was what? \\

The transcript contains many short student responses. Please provide 5 suggestions for the teacher on how the teacher could elicit more student reasoning in the classroom. Student reasoning is counted broadly as
students asking questions, engaging in mathematical discourse with their
teacher or peers, and providing explanations such as justifying their
answers. \\

Format your answer as: \\
Advice to the teacher: \\
1. Line number: <specify line number>, Segment: "<copied from transcript>", Suggestion: <specify advice to the teacher> \\
2. ... \\

Advice to the teacher:
    \end{tcolorbox}
    \begin{tcolorbox}[
    modelresponse,
    title={\textbf{One of the generated model suggestions}},
    ]
    \small
    Line number: 18, Segment: "What are you doing first?", Suggestion: Encourage the student to explain their thought process and reasoning for choosing a particular step to start with.
    \end{tcolorbox}
    \caption{
    Top box: An example prompt for suggestions on eliciting more student mathematical reasoning in the classroom. Bottom box: one of the suggestions generated by the model.
    \label{fig:suggestion_prompt_example}}
\end{figure*}


\begin{figure*}[t]
    \centering 
    \begin{tcolorbox}[
    humanresponse,
    title={\textbf{Evaluation of one of the human annotators}},
    ]
    \small
    Rating RELEVANCE \\
Level (low, mid, high): high  \\
Comment: By having the student explain his thought process, the student may have been able to catch his own error rather than having the teacher fix it for him.  \\

Rating FAITHFULNESS  \\
Level (low, mid, high)): high  \\
Comment:  \\

Rating ACTIONABILITY  \\
Level (low, mid, high)): high \\
Comment:  \\

Rating NOVELTY  \\
Level (low, mid, high)): low \\
Comment: 
    \end{tcolorbox}
    \caption{
    One of the human annotator's ratings to the model's response in Figure~\ref{fig:suggestion_prompt_example}. 
    The human annotators are also shown the transcript the model saw.
    \label{fig:suggestion_prompt_example_human_response}}
\end{figure*}

\end{document}